NANYANG TECHNOLOGICAL UNIVERSITY

SCHOOL OF COMPUTER ENGINEERING

PhD Confirmation Report

on

# Object Detection in Real Images

Submitted by: Dilip Kumar Prasad
Research Student (PhD)
School of Computer Engineering
E-mail: s080047@ntu.edu.sg

Supervisor: Dr. Maylor K. H. Leung
Associate Professor,
School of Computer Engineering
E-mail: asmkleung@ntu.edu.sg

# ABSTRACT

**Object detection and recognition are important problems in computer vision. Since these problems are meta-heuristic, despite a lot of research, practically usable, intelligent, real-time, and dynamic object detection/recognition methods are still unavailable. We propose a new object detection/recognition method, which improves over the existing methods in every stage of the object detection/recognition process. In addition to the usual features, we propose to use geometric shapes, like linear cues, ellipses and quadrangles, as additional features. The full potential of geometric cues is exploited by using them to extract other features in a robust, computationally efficient, and less meta-heuristic manner. We also propose a new hierarchical codebook, which provides good generalization and discriminative properties. The codebook enables fast multi-path inference mechanisms based on propagation of conditional likelihoods, that make it robust to occlusion and noise. It has the capability of dynamic learning. We also propose a new learning method that has generative and discriminative learning capabilities, does not need large and fully supervised training dataset, and is capable of online learning. The preliminary work of detecting geometric shapes in real images has been completed. This preliminary work is the focus of this report. Future path for realizing the proposed object detection/recognition method is also discussed in brief.**

# ACKNOWLEDGEMENTS

I am indebted to Dr. M. K. H. Leung for providing me with the opportunity to work on this project. I am also sincerely thankful for his patience and enthusiastic guidance. Without him, the project would not be in the state it is today. I also thank the Nanyang Technological University for supporting this research.

My sincere thanks are due to my seniors, colleagues, and friends for their constant encouragement. I also thank Raj, Chaoying, Hengyi, Le, Deepak, Ha Tanh and Atiqur for being good colleagues, as well as friends. My friends are too many to count here. But, I indeed count everyone of them in my heart.

I specially thank all the volunteers who are currently putting their valuable time for generating 'Human Ground Truth' for the extensive image datasets. I also thank researchers from other universities for providing me their executables and theses. These include Alex Y. S. Chia ,Dr. F. Mai, Dr. R. A. McLaughlin, Dr. M. Alder, Dr. A. Opelt, Dr. J. Shotton, Dr. L. Fei-fei, and Dr. P. D. Kovesi. I also thank others who responded to my queries and tried to help. These include Dr. E. Kim, Dr. J. L. Tan, Dr. K. Hahn, and Dr. S. H. Ong.

I also thank my family for their love and support.

# Contents

# List of Figures

# 1 Introduction

## 1.1 Background and motivation

Imparting intelligence to machines and making robots more and more autonomous and independent has been a sustaining technological dream for the mankind. It is our dream to let the robots take on tedious, boring, or dangerous work so that we can commit our time to more creative tasks. Inspired by this, many movies show fictitious robots that can do domestic work, repair machines, and fight for human. Two of the famous robots are the C-3PO and R2D2 machines in Star Wars made in 1977. Nowadays, the advances of hardware have created many good robots. One good example is the ASIMO from Honda. Another example is iCub from the RoboCub Consortium. Unfortunately, the intelligent part seems to be still lagging behind. In real life, to achieve this goal, besides hardware development, we need the software that can enable robot the intelligence to do the work and act independently. One of the crucial components regarding this is vision, apart from other types of intelligences such as learning and cognitive thinking. A robot cannot be too intelligent if it cannot see and adapt to a dynamic environment.

For a few decades, computer scientists and engineers have attached cameras and simplistic image interpretation methods to a computer (robot) in order to impart vision to the machine. A lot of interest has been shown towards object recognition, object detection, object categorization etc. Simply speaking, object recognition deals with training the computer to identify a particular object from various perspectives, in various lighting conditions, and with various backgrounds; object detection deals with identifying the presence of various individual objects in an image; and object categorization deals with recognizing objects belonging to various categories. For example, a domestic help robot can be trained to recognize if an object is a coffee machine(object recognition), it may be trained to detect a coffee machine in the kitchen (object detection), and it may be trained to identify cups of various types and forms into a common category called cups. Despite the simplistic definition mentioned above, the lines separating the three skills above are very blur and the problems often intermingle in terms of the challenges as well as solution approaches. Further, it is evident that for practical purposes, a good combination of all the three skills is essential.

Great success has been achieved in controlled environment for object detection/recognition problem but the problem remains unsolved in uncontrolled places, in particular, when objects are placed in arbitrary poses in cluttered and occluded environment. As an example, it might be easy to train a domestic help robot to recognize the presence of coffee machine with nothing else in the image. On the other hand imagine the difficulty of such robot in detecting the machine on a kitchen slab that is cluttered by other utensils, gadgets, tools, etc. The searching or recognition process in such scenario is very difficult. So far, no effective solution has been found for this problem. Despite a lot of research in this area, the methods developed so far are not efficient, require long training time, are not suitable for real time application, and are not scalable to large number of classes.

Object detection is relatively simpler if the machine is looking for detecting one particular object (say coffee machine). However, recognizing all the objects inherently requires the skill to differentiate one object from the other, though they may be of same type. Such problem is very difficult for machines, if they do not know about the various possibilities of objects.

Categorization, in our opinion is the most challenging problems of the three because of various reasons. First, the objects belonging to the same category may vary greatly in various respects. For example, in the category cups, the shapes may vary from circular cylindrical to polygonal prisms, to conical surface, to spherical surface and so on. Cups may have single handle, no handles, or two handles. They may vary in colors and may have pictures or patterns on them. The cups have to be recognized as belonging to the same category despite such diversity. The challenge is not only due to the large variations in shapes and color appearances of different objects belonging to the same category but also due to distortion by background clutter, illumination and viewpoint changes, partial occlusion and geometrical transformations (scale change, rotation, skew, etc.). Moreover, for articulated (like horse) and flexible/polymorphic (like car) categories, object instances are often presented in a diverse set of poses.

On the other hand, there might be other objects that are similar to cups in some respect but belonging to a different category. For example, tumblers, bowls, jars, certain types of bottles, vases, etc. may resemble certain cups. In an image, two distinct objects may be placed such that their overlap may be confused as a cup. For example, if a toilet paper roll is kept in front of a tape roll on a table, it may be interpreted as a cup by the machine. Moreover, some objects may belong to two categories. A cup may be used as a cup or a pen stand.

Besides intra-class and inter-class separation problem, another big problem is how many categories should be considered and what should be the basis for recognizing a category, forming a new category, etc. This also involves the problem of enabling the machine to classify objects in existing categories, if possible, recognizing the objects that do not belong to any existing categories, and learning new categories. For object categorization, there is no exhaustive and exclusive set of objects, which can be used to train the machine.

Any development of such capability "to detect and recognize objects in terms of different but known categories and then followed by 3D hand-eye coordination to pick up such an object" would bring us closer to realize our dream of employing a domestic help robot. The aim of this research is to develop a novel vision system for domestic help robot that can do category object detection and recognition in cluttered and occluded environment. This report is revised and updated version of previous report.

## 1.2 Literature review

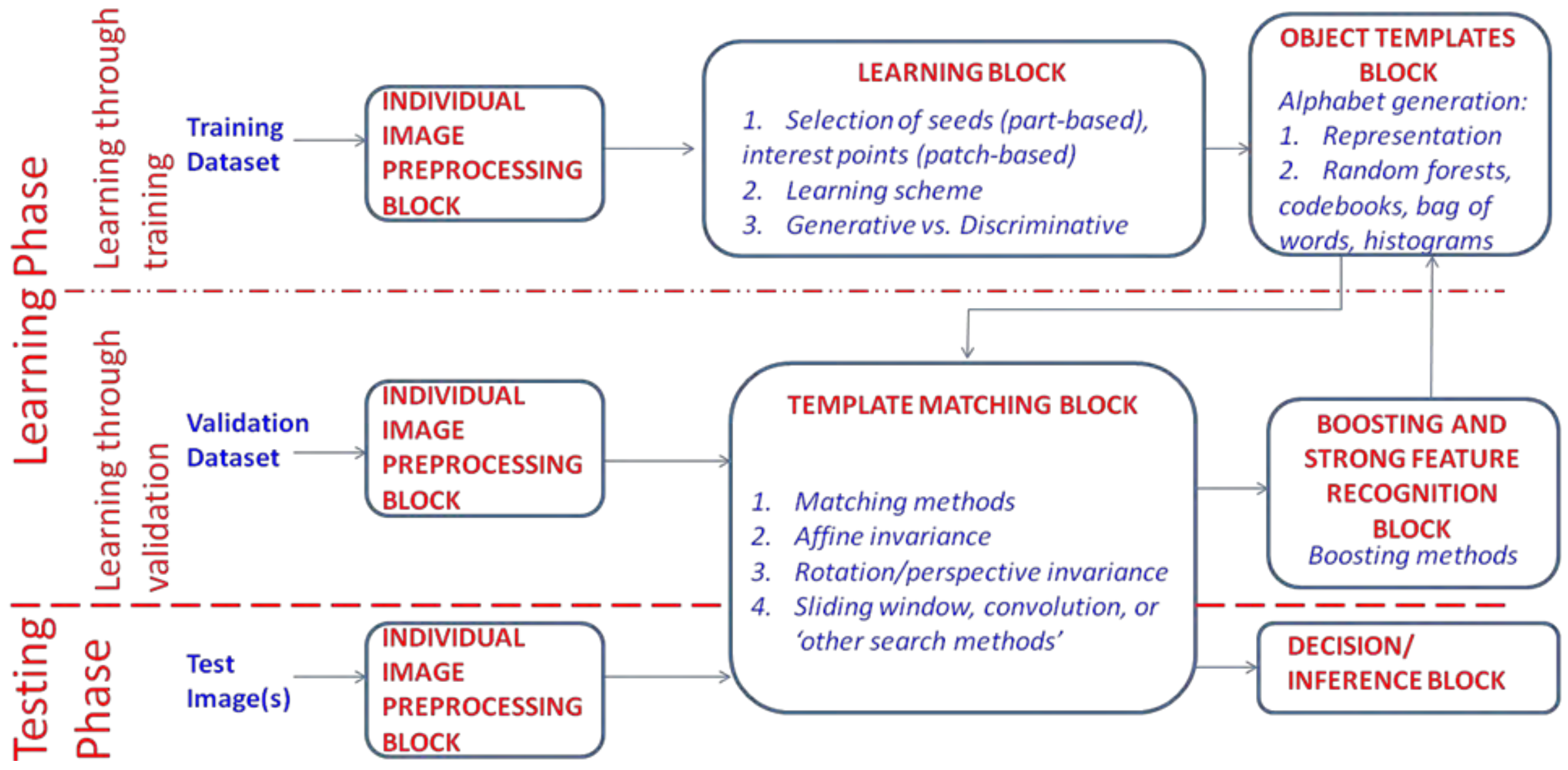

Fig. 1: Basic block diagram of a typical object detection/recognition system. The capitalized words/phrases show various blocks while the italicized words/phrases represent the various major research topics that shall be discussed in the literature review (section 1.2).

A lot of research is being done in the area of object recognition and detection. In order to facilitate the discussion about the methods and ideas of various research works, we first present a general block diagram applicable to any object detection/recognition method in Fig. 1. Specific methods proposed by various researchers may vary slightly from this generic block diagram [1].

Any such algorithm can be divided into two different phases, viz. learning phase and testing phase. In the learning phases, the machine uses a set of images which contains objects belonging to specific pre-determined class (es) in order to learn to identify the objects belonging to those classes. Once the algorithm has been trained for identifying the objects belonging to the specified classes, in the testing phase, the algorithm uses its knowledge to identify the specified class objects from the test image(s).

The algorithm for learning phase can be further subdivided into two parts, viz. learning through training and learning through validation. A set of images containing objects of the specified classes, called the training dataset, is used to learn the basic object templates for the specified classes. Depending upon the type of features (edge based features or patch based features), the training images are pre-processed and passed into the learning block. The learning block then learns the features that characterize each class. The learnt object features are then stored as object templates. This phase is referred to as 'learning through training'. The object templates learnt in this stage are termed as weak classifiers. The learnt object templates are tested against the validation dataset in order to evaluate the existing object templates. By using boosting techniques, the learnt object templates are refined in order to achieve greater accuracy while testing. This phase is referred to as 'learning through validation' and the classifiers obtained after this stage are called strong classifiers.

The researchers have worked upon many specific aspects of the above mentioned system. Some examples include the choice of feature type (edge based or patch based features), the method of generating the features, the method of learning the consistent features of an object class, the specificity of the learning scheme (does it concentrate on inter-class variability or intra-class variability), the representation of the templates, the schemes to find a match between a test/validation image and an object template (even though the size and orientation of an object in the test image may be different from the learnt template), and so on. The following discussion considers one aspect at a time and provides details upon the work done in that aspect.

### *1.2.1 Feature types*

Most object detection and recognition methods can be classified into two categories based on the feature type they use in their methods. The two categories are edge-based feature type and patch based feature type. It is notable that some researchers have used a combination of both the edge-based and patch-based features for object detection [2-6]. In our opinion, using a combination of these two features shall become more and more prevalent in future because such scheme would yield a system that derives the advantages of both the feature types. A good scheme along with the advances in computational systems should make it feasible to use both feature types in efficient and semi-real time manner.

#### *1.2.1.1 Edge-based features*

The methods that use edge-based feature type extract the edge map of the image and identify the features of the object in terms of edges. Some examples include [2, 3, 7-22]. Using edges as features is advantageous over other features due to various reasons. As discussed in [7], they are largely invariant to illumination conditions and variations in objects' colors and textures. They also represent the object boundaries well and represent the data efficiently in the large spatial extent of the images.

In this category, there are two main variations: use of the complete contour (shape) of the object as the feature [8-13, 15, 18] and use of collection of contour fragments as the feature of the object [2, 3, 7, 14-21]. Fig. 2 shows an example of complete contour and collection of contours for an image.

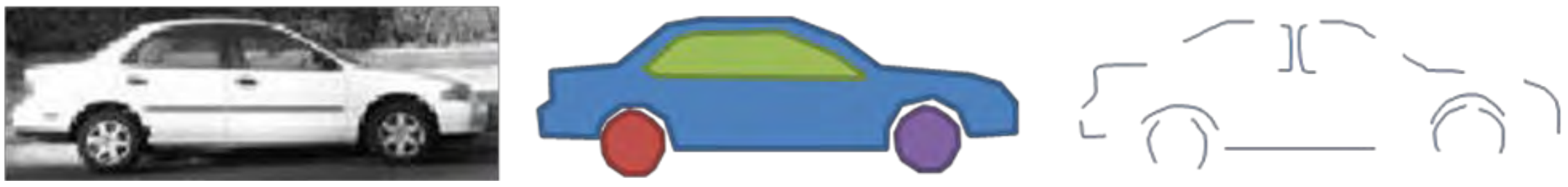

(a) Example image (b) Contour (shape) as feature (c) Contour fragments as feature

Fig. 2: Edge-based feature types for an example image

The main motivation of using the complete contours as features is the robustness of such features to the presence of clutter [7, 12, 18, 23]. One of the major concerns regarding such feature type is the method of obtaining the complete contours (especially for training images). In real images, typically incomplete contours are inevitable due to occlusion and noise. Various researchers have tried to solve this problem to some extent [8, 12, 13, 15, 18, 21]. Hamsici [8] identified a set of landmark points from the edges and connected them to obtain a complete shape contour. Schindler [12] used segmenting approaches [24, 25] to obtain closed contours from the very beginning (he called the areas enclosed by such closed contours as super pixels). Ferrari [18, 21] used a sophisticated edge detection method that provides better edges than contemporary methods for object detection. These edges were then connected across the small gaps between them to form a network of closed contours. Ren [15] used a triangulation to complete the contours of the objects in natural images, which are significantly difficult due to the presence of background clutter. Hidden state shape model was used by Wang [26] in order to detect the contours of articulate and flexible/polymorphic objects. It is noticeable that all of these methods require additional computation intensive processing and are typically sensitive to the choice of various empirical contour parameters. The other problem involving such feature is that in the test and validation images, the available contours are also incomplete and therefore the degree of match with the complete contour is typically low [12]. Though some measures, like kernel based [8, 27] and histogram based methods [9, 10], can be taken to alleviate this problem, the detection of the severely occluded objects is still very difficult and unguaranteed. Further, such features are less capable of incorporating the pose or viewpoint changes, large intra-class variability, articulate objects (like horses) and flexible/polymorphic objects (like cars) [12, 18, 21]. This can be explained as follows. Since this feature type deals with complete contours, even though the actual impact of these situations is only on some portions of the contour, the complete contour has to be trained.

On the other hand, the contour fragment features are substantially robust to occlusion if the learnt features are good in characterizing the object [2, 7, 9, 17, 18, 21, 28]. They are less demanding in computation as well as memory as the contour completion methods need not be applied and relatively less data needs to be stored for the features. The matching is also expected to be less sensitive to occlusion [7, 29]. Further, special cases like viewpoint changes, large intra-class variability, articulate objects and flexible/polymorphic objects can be handled efficiently by training the fragments (instead of the complete contour) [3, 7, 9, 18, 21, 29].

However, the performance of the methods based on contour fragment features significantly depends upon the learning techniques. While using these features, it is important to derive good feature templates that represent the object categories well (in terms of both inter-class and intra-class variations) [2, 30]. Learning methods like boosting [28, 30-51] become very important for such feature types, and shall be discussed further in section 1.2.5.

The selection of the contour fragments for characterizing the objects is an important factor and can affect the performance of the object detection/recognition method. While all the contour fragments in an image cannot be chosen for this purpose, it has to be ensured that the most representative edge fragments are indeed present and sufficient local variation is considered for each representative fragment. In order to look for such fragments, Opelt [2] used large number of random seeds that are used to find the candidate fragments and finally derives only two most representative fragments as features. Shotton [7] on the other hand generated up to 100 randomly sized rectangular units in the bounding box of the object to look for the candidate fragments. It is worth noting that the method proposed in [2] becomes computationally very expensive if more than two edge fragments are used as features for an object category. While the method proposed by Shotton [7] is computationally efficient and expected to be more reliable as it used numerous small fragments (as compared to two most representative fragments), it is still limited by the randomness of choosing the rectangular units.

On the other hand, Chia [16] used some geometrical shape support (ellipses and quadrangles) in addition to the fragment features for obtaining more reliable features. Use of geometrical structure, relationship between arcs and lines, and study of structural properties like symmetry, similarity and continuity for object retrieval were proposed in [52]. Though the use of geometrical shape (or structure) for estimating the structure of the object is a good idea, there are two major problems with the methods in [16, 52]. First problem is that some object categories may not have strong geometrical (elliptic and quadrangle) structure (example horses) and the use of weak geometrical structure may not lead to robust descriptors of such objects. Though [16] demonstrates the applicability for animals, the geometrical structure derived for animals is very generic and applicable to many classes. Thus, the inter-class variance is poor. The classes considered in [16], viz., cars, bikes and four-legged animals (four-legged animals is considered a single class) are very different from each other. Similarly, [52] concentrates on logos and the images considered in [52] have white background, with no natural background clutter and noise. Its performance may degrade significantly in the presence of noise and natural clutter. The second problem is that sometimes occlusion or flexibility of the object may result in complete absence of the components of geometrical structure. For example, if the structural features learnt in [52] are occluded, the probability of detecting the object is very low. Similarly, if the line features learnt in [16], used for forming the quadrangle are absent, the detection capability may reduce significantly.

Though we strongly endorse the idea of using geometric shapes for object detection [53], we suggest that such information should not be used as the only features for object detection. In addition, they can be used to derive good fragment features and reduce the randomness of selection of the fragments. Our ideas regarding this are discussed in more detail in section 1.3 and chapter 4.

*1.2.1.2 Patch-based features*

The other prevalent feature type is the patch based feature type, which uses appearance as cues. This feature has been in use since more than two decades [54], and edge-based features are relatively new in comparison to it. Moravec [54] looked for local maxima of minimum intensity gradients, which he called corners and selected a patch around these corners. His work was improved by Harris [55], which made the new detector less sensitive to noise, edges, and anisotropic nature of the corners proposed in [54].

In this feature type, there are two main variations:

1) patches of rectangular shapes that contain the characteristic boundaries describing the features of the objects [2, 56-61]. Usually, these features are referred to as the local features.

2) irregular patches in which, each patch is homogeneous in terms of intensity or texture and the change in these features are characterized by the boundary of the patches. These features are commonly called the region-based features.

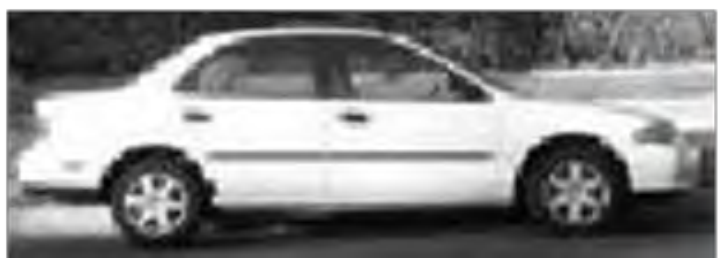

(a) Example image

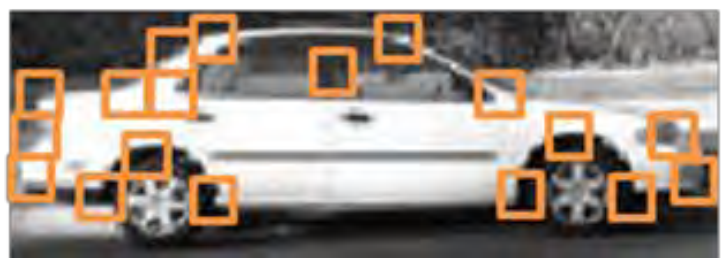

(b) Regular patches

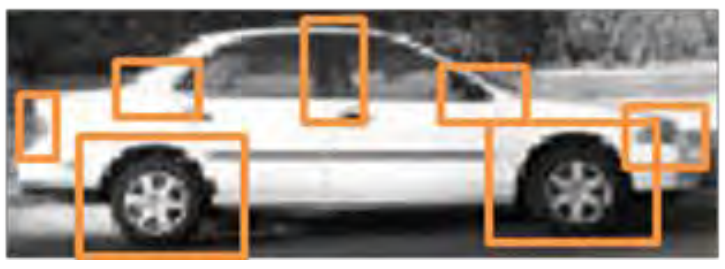

(c) Regular patches of various sizes

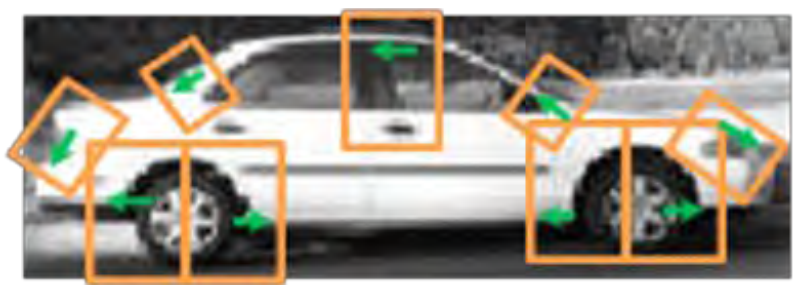

(d) Oriented regular patches

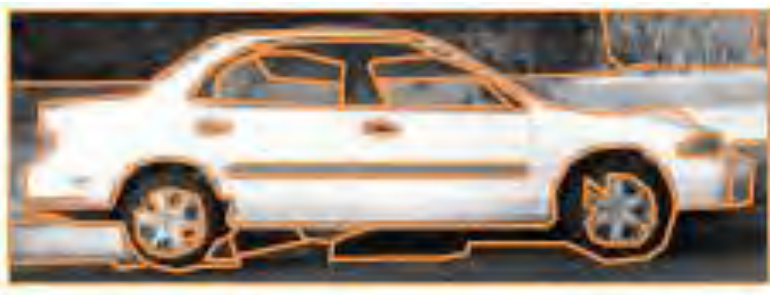

(e) Irregular region patches

Fig. 3: Patch-based feature types for an example image. Feature types shown in (b)-(d) are called local features, while the feature type shown in (e) is called region-based features.

Fig. 3 shows these features for an example image. Subfigures (b)-(d) show local features while subfigure (e) shows region based features (intensity is used here for extracting the region features). As shown in Fig. 3(b)-(d), the local features may be of various kinds. The simplest form of such features use various rectangular or square local regions of the same size in order to derive the object templates [62]. Such features cannot deal with multi-scaling (appearance of the object in various sizes) effectively. A fixed patch size may not be suitable because of the following reason. If the patch size is small, it may not cover a large but important local feature. Information of such feature may be lost in the smaller patch. On the other hand, if the patch size is large, it may cover more than one independent feature, which may or may not be present simultaneously in other images. Further, there is no way to determine the size that is optimal for all the images and various classes. Another shortcoming is that many small rectangular patches need to be learnt as features and stored in order to represent the object well. This is both computationally expensive and memory intensive.

A better scheme is to use features that may be small or big in order to appropriately cover the size of the local feature such that the features are more robust across various images, learning is better and faster, and less storage is required [63].

A pioneering work was done by Lowe [59], which enabled the use of appropriately oriented variable sized features for describing the object. He proposed a scale invariant feature transformation (SIFT) method. Lowe describes his method of feature extraction in three stages. He first identified potential corners (key points) using difference of Gaussian function, such that these feature points were invariant to scale and rotation. Next, he identified and selected the corners that are most stable and determined their scale (size of rectangular feature). Finally, he computed the local image gradients at the feature points and used them to assign orientations to the patches. The use of oriented features also enhanced the features' robustness to small rotations. With the use of orientation and scale, the features were transformed (rotated along the suitable orientation and scaled to a fixed size) in order to achieve scale and rotational invariance. In order to incorporate the robustness to illumination and pose/perspective changes, the features were additionally described using the Gaussian weighing function along various orientations.

One of the major concerns in all the above schemes is the identification of good corner points (or key-points) that are indeed representative of the data. This issue has been studied by many researchers [5, 59, 64-66]. Lowe [59] studied the stability of the feature points. However, his proposal would apply to his schema of features only. Carneiro [65] and Comer [67] proposed stability measures that could be applied to wide range and varieties of algorithms.

Another major concern is to describe these local features. Though the features can be directly described and stored by saving the pixel data of the local features, such method is naive and inefficient. Researchers have used many efficient methods for describing these local features. These include PCA vectors of the local feature (like PCA-SIFT) [22, 68], Fischer components [69, 70], wavelets and Gabor filters [14], eigen spaces [71], kernels [8, 22, 27, 72, 73], etc. It is important to note that though these methods use different tools for describing the features, the main mathematical concept behind all of them is the same. The concept is to choose sufficient (and yet not many) linearly independent vectors to represent the data in a compressed and efficient manner [14]. Another advantage of using such methods is that each linearly independent vector describes a certain property of the local feature (depending on the mathematical tool used). For example, a Gabor wavelet effectively describes an oriented stroke in the image region [14]. Yet another advantage of such features is that while matching the features in the test images, properties of linear algebra (like linear dependence, orthogonality, null spaces, rank, etc.) can be used to design efficient matching techniques [14].

The region-based features are inspired by segmentation approaches and are mostly used in algorithms whose goal is to combine localization, segmentation, and/or categorization. While intensity is the most commonly used cue for generating region based features [48, 64, 74], texture [3, 74-77], color [76-78], and minimum energy/entropy [79, 80] have also been used for generating these features. It is notable that conceptually these are similar to the complete contours discussed in edge-based features. Such features are very sensitive to lighting conditions and are generally difficult from the perspective of scale and rotation invariance. However, when edge and region based features are combined efficiently, in order to represent the outer boundary and inner common features of the objects respectively, they can serve as powerful tools [3]. Some good reviews of feature types can also be found in [56, 81, 82].

In our opinion, SIFT features provide a very strong scheme for generating robust object templates [59]. It is worth mentioning that though SIFT and its variants were proposed for patch-based features, they can be adapted to edge-fragments based features too. Such adaptation can use the orientation of edges to make the matching more efficient and less sensitive to rotational changes. Further, such scheme can be used to incorporate articulate and flexible/polymorphic objects in a robust manner. These are discussed in more detail in chapter 4.

It has been argued correctly by many researchers that a robust object detection and characterization scheme shall typically require more than one feature types to obtain good performance over large number of classes [2, 3, 6, 18, 19, 21, 47, 83-88]. Thus, we shall use region features along with contour fragments. As compared to [2], which has used only one kind of object template for making the final decision, we shall use a combined object template that stores edge, shape, and region features and assigns a strength value to each feature so that combined probabilistic decision can be made while testing. Such scheme shall ensure that potential objects are identified more often, though the trust (likelihood) may vary and the decision can be made by choosing appropriate threshold. This shall be especially useful in severely occluded or noisy images.

### *1.2.2 Generative model vs. discriminative model*

The relationship (mapping) between the images and the object classes is typically non-linear and non-analytic (no definite mathematical model applicable for all the images and all the object classes is available). Thus, typically this relationship is modeled using probabilistic models [89]. The images are considered as the observable variables, the object classes are considered as the state variables, and the features are considered as intermediate (sometimes hidden) variables. Such modeling has various advantages. First, it provides a generic framework which is useful for both the problems of object detection and recognition (and many other problems in machine vision and outside it). Second, such framework can be useful in evaluating the nature and extent of information available while training, which subsequently helps us to design suitable training strategies.

The probabilistic models for our problems can be generally classified into two categories, viz. discriminative models and generative models [90, 91]. It shall be helpful to develop a basic mathematical framework for understanding and comparing the two models. Let the observable variables (images) be denoted by $\mathbf{x}_i$, $i = 1$ to $N$, where $N$ is the number of training images. Let the corresponding state variables (class labels) be denoted as $c_i$ and the intermediate variables (features/ feature descriptors) be denoted as $\boldsymbol{\theta}_i$. Accordingly, a simplistic graphical representation [91] of the discriminative and generative models is presented in Fig. 4.

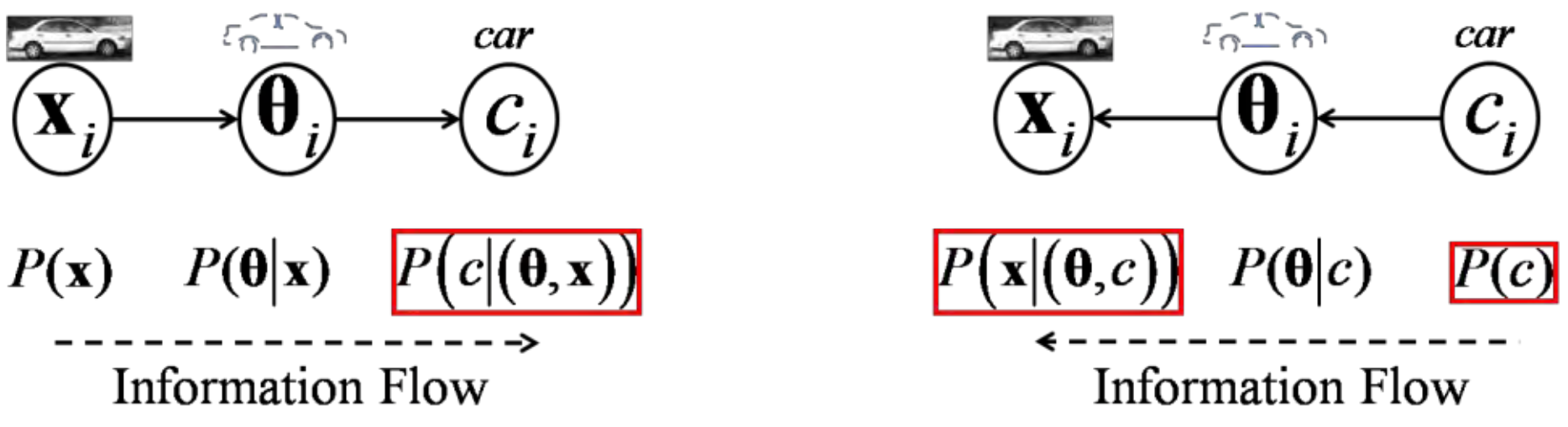

(a) Discriminative model

(b) Generative model

Fig. 4: Graphical illustration of the discriminative and generative models. The probabilities in boxes are the model defining probabilities for the respective models.

As seen in the figure, the discriminative model uses a map from the images to the class labels, and thus the flow of information is from the observables (images) to the state variables (class labels) [91]. Considering the joint probability $P(c,\boldsymbol{\theta},\mathbf{x})$, discriminative models expand $P(c,\boldsymbol{\theta},\mathbf{x})$ as $P(c,\boldsymbol{\theta},\mathbf{x}) = P\left(c|(\boldsymbol{\theta},\mathbf{x})\right)P(\boldsymbol{\theta}|\mathbf{x})P(\mathbf{x})$. Thus, $P\left(c|(\boldsymbol{\theta},\mathbf{x})\right)$ is the model defining probability [90] and the training goal is:

$$P\left(c|(\boldsymbol{\theta},\mathbf{x})\right) = \begin{cases} \alpha & \text{if } \mathbf{x} \text{ contains object of class } c \\ \beta & \text{otherwise} \end{cases} \tag{1}$$

Ideally, $\alpha = 1$ and $\beta = 0$. Indeed, practically this is almost impossible to achieve, and values between [0,1] are chosen for $\alpha$ and $\beta$.

In contrast, the generative model uses a map from the class labels to the images, and thus the flow of information is from the state variables (class labels) to the observables (images) [91]. Generative models use the expansion of the joint probability $P(c,\boldsymbol{\theta},\mathbf{x}) = P\left(\mathbf{x}|(\boldsymbol{\theta},c)\right)P(\boldsymbol{\theta}|c)P(c)$. Thus, $P\left(\mathbf{x}|(\boldsymbol{\theta},c)\right)$ and $P(c)$ are the model defining probabilities [90] and the training goal is:

$$P\left(\mathbf{x}|(\boldsymbol{\theta},c)\right)P(c) = \begin{cases} \alpha & \text{if } \mathbf{x} \text{ contains object of class } c \\ \beta & \text{otherwise} \end{cases} \tag{2}$$

Ideally, $\alpha = 1$ and $\beta = 0$. Indeed, practically this is almost impossible to achieve, and some realistic values are chosen for $\alpha$ and $\beta$. It is important to note that in unsupervised methods, the prior probability of classes, $P(c)$ is also unknown.

Further mathematical details can be found in [90, 91]. The other popular model is the descriptive model, in which every node is observable and is interconnected to every other node. It is obvious that the applicability of this model to the considered problem is limited. Therefore, we do not discuss this model any further. It shall suffice to make a note that such models are sometimes used in the form of conditional random fields/forests [13, 48, 75].

With the above mentioned mathematical structure as a reference, we can now compare the discriminative and generative models from various aspects, in the following sub-sections.

*1.2.2.1 Comparison of their functions*

As the name indicates, the main function of the discriminative models is that for a given image, it should be able to discriminate the possibility of occurrence of one class from the rest. This is evident by considering the fact that the probability $P\left(c|\left(\boldsymbol{\theta},\mathbf{x}\right)\right)$ is the probability of discriminating the class labels $c$ for a given instance of image $\mathbf{x}$. On the other hand, the main function of generative models is to be able to predict the possibility of generating the object features $\boldsymbol{\theta}$ in an image $\mathbf{x}$ if the occurrence of the class $c$ is known. In other words, the probabilities $P\left(\mathbf{x}|\left(\boldsymbol{\theta},c\right)\right)P(c)$ together represent the probability of generating random instances of $\mathbf{x}$ conditioned to class $c$. In this context, it is evident that while discriminative models are expected to perform better for object detection purposes, generative models are expected to perform better for object recognition purposes [19]. This can alternatively be understood as the generative models are used to learn class models (and be useful even in large intra-class variation) [47, 60, 92, 93] while discriminative models are useful for providing maximum inter-class variability [93].

*1.2.2.2 Comparison of the conditional probabilities of the intermediate variables*

In the discriminative models, the intermediate conditional probability is $P(\boldsymbol{\theta}|\mathbf{x})$, while in the generative models, the intermediate conditional probability is $P(\boldsymbol{\theta}|c)$. Since we are interested in the joint probability $P(c,\boldsymbol{\theta},\mathbf{x})$, the probabilities $P(\boldsymbol{\theta}|\mathbf{x})$ and $P(\boldsymbol{\theta}|c)$ play an important role, though they do not appear in the training goals. In the discriminative models, $P(\boldsymbol{\theta}|\mathbf{x})$ represents the strength of the features $\boldsymbol{\theta}$ in representing the image well [18, 21], while in the generative models, $P(\boldsymbol{\theta}|c)$ represent the strength of features in representing the class well. Though ideally we would like to maximize both, depending upon the type of feature and the problem, the maximum value of these probabilities is typically less than one. Further, it is difficult to quantitatively measure these probabilities in practice. In our opinion, while the shape features (closed contours) and region features (irregular patches) are more representative of the class (the object's 3-dimensional or 2-dimensional model), the edge fragments and local features are more representative of the images [2, 47]. Thus, while shape and region features are widely used for segmentation and recognition, local features and edge fragments have been used more often for object detection [18, 19, 21, 47, 85]. Considering this argument, though most methods that use multiple feature types choose these feature types randomly, we recommend to choose a combination of two feature types where one feature is robust for characterizing the image, while the other is good in characterizing the class. In this regard, combining edge fragments and region features is the combination that is easiest to handle practically. Due to this many new methods have used a combination of these features [3, 6, 86-88].

#### *1.2.2.3 Training data size and supervision*

Mathematically, the training data size required for generative model is very large (at least more than the maximum dimension of the observation vector $\mathbf{x}$ ). On the other hand, discriminative models perform well even if the training dataset is very small (more than a few images for each class type). This is expected because the discriminative models invariably use supervised training dataset (the class label is specifically mentioned for each image). On the other hand, generative models are unsupervised (semi-supervised, at best) [94]. Not only the posterior probability $P\left(\mathbf{x}\middle|\left(\boldsymbol{\theta},c\right)\right)$ is unknown, the prior probability of the classes $P(c)$ is also unknown for the generative models [90]. Another point in this regard is that since generative models do not require supervision and the training dataset can be appended incrementally [19, 90, 92] as vision system encounters more and more scenarios, generative models are an important tool for expanding the knowledge base, learning new classes, and keeping the overall system scalable in its capabilities.

#### *1.2.2.4 Comparison of accuracy and convergence*

The discriminative models usually converge fast and correctly (explained by supervised dataset). If the size of training dataset is asymptotically large, the convergence is guaranteed for the generative models as well. However, such convergence may be correct convergence or misconvergence. If the generative models converge correctly, then the accuracy of generative models is comparable to the accuracy of the discriminative models. But, if there has been a misconvergence, then the accuracy of the generative models is typically poorer than the discriminative models [95]. Since the dataset is typically finite, and in most cases small, it is important to compare the accuracy of these models when the dataset is finite. Mathematical analysis has shown that in such cases, the accuracy of the generative models is always lower than the discriminative methods [95]. It is notable that due to their basic nature (as described by information flow and discussed in section 1.2.2.1), generative models provide good recall but poor precision, while discriminative models provide poorer recall but good precision. The restrictive nature of generative models has prompted more and more researchers to consider discriminative models [2, 18, 21, 78, 96-102]. On the other hand, considering the scalability, generalization properties, and non-supervised nature of generative models, other researchers are trying to improve the performance of generative models by using partial supervision or coupling the generative models and discriminative models in various forms [5, 19, 28, 60, 78, 92, 94, 103].

#### *1.2.2.5 Learning methods*

Generative models use methods like Bayesian classifiers/networks [19, 28, 60, 92], likelihood maximization [92, 103], and expectation maximization [5, 78, 94, 103]. Discriminative models typically use methods like logistic regression, support vector machines [18, 21, 78, 96-100], and k-nearest neighbors [78, 101, 102]. The k-nearest neighbors scheme can also be used for multi-class problems directly, as demonstrated in [101]. Boosting schemes are also examples of methods for learning discriminative models [2], though they are typically applied on already learnt weak features (they shall be discussed later in greater detail). In the schemes where generative and discriminative models are combined [78, 104], there are two main variations: generative models with discriminative learning [5, 86, 90], and discriminative models with generative learning [91]. In the former, typically maximum likelihood or Bayesian approaches are combined with boosting schemes or incremental learning schemes [5, 47, 86, 90, 92], while in the latter, usual discriminative schemes are augmented by 'generate and test' schemes in the feedback loop [91, 105].

#### *1.2.2.6 Our preference*

In the context of the above discussion, we consider generative models with incremental/discriminative learning [5, 90] more preferable than other schemes. The first reason for our choice is that discriminative learning is indeed being done in learning through validation stage. Thus, instead of making the method highly discriminative (by choosing discriminative model in the 'learning through training' stage as well), it shall be better to incorporate the generalization and scalable capabilities of generative models. Further, as demonstrated by a few previous works, using incremental learning, generative models can be made more accurate (in both recall and precision) and real-time online learning capable [5, 47, 92]. The learning of generative models can be enhanced by selecting suitable training images (for example, uncluttered images), thus effectively using semi-supervised approach.

### *1.2.3 Object templates and their representation*

The learning method has to learn a mapping between the features and the classes. Typically, the features are extracted first, which is followed by either the formation of class models (in generative models) or the most discriminative features for each class (in discriminative models) or random fields of features in which a cluster represents an object class (descriptive models, histogram based schemes, Hough transform based methods, etc). Based on them, the object templates suitable for each class are learnt and stored for the future use (testing). This section will discuss various forms of object templates used by researchers in computer vision.

While deciding on an object template, we need to consider factors like:

- Is the template most representative form of the class (in terms of the aimed specificity, flexibility of the object, intra-class variation, etc)? For example, does it give the required intra-class and inter-class variability features? Does it need to consider some common features among various classes or instances of hierarchical class structure? Does it need to consider various poses and/or perspectives? Does it need to prioritize certain features (or kind of features)?

- Is the model representation an efficient way of storing and using the template? Here, memory and computations are not the only important factors. We need to also consider if the representation enables good decision mechanisms.

The above factors will be the central theme in discussing the specific merits and demerits of the various existing object templates. We begin with the object templates that use the spatial location of the features. Such templates specifically represent the relative position of the features (edge fragments, patches, regions) in the image space. For this, researchers typically represent each feature using a single representative point (called the centroid) and specify a small region in which the location of the centroid may vary in various objects belonging to the same class [2, 7]. Then all the centroids are collected together using a graph topology. For example some researchers have used a cyclic/chain topology [12]. This simplistic topology is good to represent only the external continuous boundary of the object. Due to this, it is also used for complete contour representation, where the contour is defined using particular pivot points which are joined to form the contour [12]. Such a topology may fail if the object is occluded at one of the centroid locations, as the link between the chain is not found in such case and the remaining centroids are also not detected as a consequence. Further, if some of the characteristic features are inside the object boundary, deciding the most appropriate connecting link between the centroids of the external and internal boundaries may be an issue and may impact the performance of the overall algorithm. Other topology in use is the constellation topology [92, 106, 107], in which a connected graph is used to link all the centroids. A similar representation is being called multi-parts-tree model in [79], though the essentials are same. However, such topology requires extra computation in order to find an optimal (neither very deep nor very wide) representation. Again, if the centroids that are linked to more than one centroid are occluded, the performance degrades (though not as strongly as the chain topology). The most efficient method in this category is the star topology, in which a central (root) node is connected to all the centroids [2, 7, 9, 61]. The root node does not correspond to any feature or centroid and is just a virtual node (representing the virtual centroid of the complete object). Thus, this topology is able to deal with occlusion better than the other two topologies and does not need any extra computation for making the topology.

Other methods in which the features are described using transformation methods (like the kernel based methods, PCA, wavelets, etc., discussed in section 1.2.1.2), the independent features can be used to form the object templates. The object templates could be binary vectors that specify if a particular feature is present in an object or not. Such object templates are called bag-of-words, bag of visual words, or bag of features [2, 80, 96, 97, 100, 108-110]. All the possible features are analogous to visual words, and specific combinations of words (in no particular order) together represent the object classes. Such bag of words can also be used for features like colors, textures, intensity, shapes [80], physical features (like eyes, lips, nose for faces, and wheels, headlights, mirrors for cars) etc. [78, 109, 111]. As evident, such bag of words is a simple yet powerful technique for object recognition and detection but may perform poorly for object localization and segmentation. As opposed to them, spatial object templates are more powerful for image localization and segmentation.

In either of the above cases, the object templates can also be in the form of codebooks [2, 7, 18, 21, 60, 61, 110, 112]. A codebook contains a specific code of features for each object class. The code contains the various features that are present in the corresponding class, where the sequence of features may follow a specific order or not. An unordered codebook is in essence similar to the concept of bag of words, where the bag of words may have greater advantage in storing and recalling the features and the object templates. However, codebooks become more powerful if the features in the code are ordered. A code in the order of appearance of spatial templates can help in segmentation [7], while a code in the order of reliability or strength of a feature for a class shall make the object detection and recognition more robust.

Other hierarchical (tree like) object templates may be used to combine the strengths of both the codebooks and bag of words, and to efficiently combine various feature types [5, 19, 60, 69, 74, 77, 86, 94, 103, 107, 110, 113].

Another important method of representing the object templates is based on random forests/fields [75, 105, 114]. In such methods, no explicit object template is defined. Instead, in the feature space (where each feature represents one dimension), clusters of images belonging to same object class are identified [59, 60, 114]. These clusters in the feature space are used as the probabilistic object templates [69]. For every test image, its location in feature space and distance from these clusters determine the decision.

We prefer a hierarchical codebook, similar to the multi-parts-tree model [79, 94], which combines at least two feature types. We intend to place the strongest (most consistent and generic) features at the highest level and weaker features in subsequent nodes. Any single path in the hierarchy shall serve as a weak but sufficient object template and typically the hope is that more than one paths are traversed if object of the class is present in an image. If all the paths are traversed, the image has a strong presence of the object class. The final inference will be based on the number and depth of the paths traversed. It is worth mentioning that while [79] used a minimization of the energy and Mahalanobis distance of the parts for generating the tree, we shall use the likelihood of each feature independently, and likelihood of each feature conditioned to the presence of higher level features in the tree. We might have considered another hierarchical structure where the strongest (but few) descriptors appear at the leaf nodes and the path towards the root incrementally confirms the presence of the object. But that would either require multiple bottom-up traversal (in order to reach the root) or a top-down traversal with very low initial confidence. On the other hand, the chosen top-down structure will ensure that we begin with a certain degree of confidence (due to the generic features with high likelihood at the highest level, details in section 4.1) in the presence of the object class and then tweak our confidence as we go further down the tree. If we cannot go further down the tree, we need not look for multiple other traversal paths beginning again from the top.

### *1.2.4 Matching schemes and decision making*

Once the object templates have been formed, the method should be capable of making decisions (like detecting or recognizing objects in images) for input images (validation and/or test images). We first discuss about the methods of finding a match between the object template and the input image and then discuss about the methods of making the final decision.

Discussion regarding matching schemes is important because of various reasons. While the training dataset can be chosen to meet certain requirements, it cannot be expected that the test images also adhere to those requirements. For example, we may choose that all the training images are of a particular size, illumination condition, contain only single object of interest viewed from a fixed perspective, in uncluttered (white background), etc., such restrictions cannot be imposed on the real test images, which may be of varying size, may contain many objects of interest and may be severely cluttered and occluded and may be taken from various viewpoints.

The problem of clutter and occlusion is largely a matter of feature selection and learning methods. Still, they may lead to wrong inferences if improper matching techniques are used. However, making the matching scheme scale invariant, rotation and pose invariant (at least to some degree), illumination independent, and capable of inferring multiple instances of multiple classes is important and has gained attention of many researchers [7, 53, 65, 67, 107, 115-144].

If the features in the object templates are pixel based (for example patches or edges), the Euclidean distance based measures like Hausdorff distance [133, 143, 145, 146] and Chamfer distance [2, 7, 18, 23, 79, 120, 129] provide quick and efficient matching tools. However, the original forms of both these distances were scale, rotation, and illumination dependent. Chamfer distance has become more popular in this field because of a lot of incremental improvement in Chamfer distance as a matching technique. These improvements include making it scale invariant, illumination independent, rotation invariant, and more robust to pose variations and occlusions [2, 7, 18, 23, 79, 120, 129]. Further, Chamfer distance has also been adapted for hierarchical codebooks [120]. In region based features, concepts like structure entropy [80, 147], mutual information [80, 113], and shape correlation have been used for matching and inference [116, 117]. Worth attention is the work by Wang [80] that proposed a combination of local and global matching scheme for region features. Such scheme can perform matching and similarity evaluation in an efficient manner (also capable of dealing with deformation or pose changes) by incorporating the spatial mutual information with the local entropy in the matching scheme.

Another method of matching/inferring is to use the probabilistic model in order to evaluate the likelihood ratio [3, 5, 60] or expectation in generative models [89, 94]. Otherwise, correlation between the object template and the input image can be computed or probabilistic Hough transform can be used [62, 77, 78, 99]. Each of these measures is linked directly or indirectly with the defining ratio of the generative model (see section 1.2.2), $P\left(\mathbf{x}|\left(\boldsymbol{\theta},c\right)\right)$, which can be computed for an input image and a given class through the learnt hidden variables $\boldsymbol{\theta}$ [14]. For example, in the case of wavelet form of features, $P\left(\mathbf{x}|\left(\boldsymbol{\theta},c\right)\right)$ will depend upon the wavelet kernel response to the input image for a particular class [14]. Similarly, the posterior probability can be used for inference in the discriminative models. Or else, in the case of classifiers like SVM, k-nearest neighbors based method, binary classifiers, etc, the features are extracted for the input image and the posterior probability (based on the number of features voted into each class) can be used for inference [18, 21, 69]. If two or more classes have the high posterior probability, multiple objects may be inferred [60, 79]. However, if it is known that only one object is present in an image, refined methods based on feature reliability can be used.

If the object class is represented using the feature spaces, the distance of the image from the clusters in feature space is used for inference. Other methods include histograms corresponding to the features (the number of features that were detected) to decide the object category [59, 69, 109, 114].

Since, we propose to use a hierarchical object template, we will use suitable matching schemes for each of the nodes in the tree [19, 69]. For example, the edge features could use Chamfer distance, while the region features could use correlation/entropy based matching schemes. Finally, matching at each node gives a probabilistic value, thus giving us a hierarchical tree of probabilities which will be easy to combine and infer. This may be similar to the belief propagation scheme used in [107].

### *1.2.5 Boosting methods - learning while validation*

The weak object templates learnt during training can be made more class specific by using boosting mechanisms in the validation phase [148-169]. Boosting mechanisms typically consider an ensemble of weak features (in the object templates) and gives a boost to the stronger features corresponding to the object class. Technically, boosting method can be explained as follows. Suppose validation images $\mathbf{x}_i$, $i = 1$ to $N$ contain the corresponding class labels $c_i = \pm 1$, where the value 1 indicates that the object of the considered class is present and $-1$ represents its absence. Let the weak classifier learnt while training be a combination of several individual classifiers $h_j(\cdot)$, $j = 1$ to $J$. Here, $h_j(\cdot)$ operates on the input image and gives an inference/decision regarding the presence/absence of class object. Evidently, $h_j(\cdot)$ is determined by the feature $\theta_j$ in the codebook and the inference mechanisms. Further, let us say that we want to extract maximum $T$ strong classifiers. Then most boosting methods can be generally explained using the algorithm below:

| | |
|---|---|
| Step 1: | Initialize the image weights $w_{i,t} = 1/N$; $\forall i$ |
| Step 2: | For $t = 1$ to $T$ |
| Step 2.1: | Find the strongest classifier, $h_t(\cdot)$, using the current image weights. For this, first compute the error function for each classifier: $\varepsilon_j = \sum_i \left( w_{i,t} I_{j,i} \right)$, where $I_{j,i} = 1$ if $c_i = h_j(\mathbf{x}_i)$, and 0 otherwise. Here, the index $j$ is used to denote the $j$ th classifier and the index $i$ is used to denote the $i$ th image. Find the classifier that resulted in minimum error (this is the strongest classifier for the weights $w_{i,t}$): $h_t(\cdot) = \arg\left(\min\left(\varepsilon_j\right)\right)$. |
| Step 2.2: | Update the classifier weight for the chosen classifier: $\alpha_t = f\left(\varepsilon_t\right)$, where the function $f(\cdot)$ depends upon the chosen boosting technique and $\varepsilon_t$ is the error corresponding to the current strongest classifier $h_t(\cdot)$. |
| Step 2.3: | If a termination condition is satisfied, then go to step 3. The termination condition depends upon the application or the boosting method used. |
| Step 2.4: | Update the weights $w_{i,t+1} = w_{i,t} g(\alpha_t I_i)$. Here, $g(\cdot)$ is the function that changes the weight distribution given to the validation images and is generally called the loss function. The general characteristic of $g(\cdot)$ is that it reduces the weight of the images that resulted in correct classification, so that in the next iteration, the method is less biased towards the current strong feature. Typically, $w_{i,t+1}$ is normalized after computation such that the sum of all the weights is 1. |
| Step 3: | The output of the boosting algorithm is typically specified as the strong classifier $h_{\text{strong}}(\cdot) = \sum_t \alpha_t h_t(\cdot)$. |

Fig. 5: Generic algorithm for boosting

It is notable that some features may be repeatedly selected in step 2 of Fig. 5, which indicates that though the method is getting lesser and lesser biased towards that feature, that feature is strong enough to be selected again and again.

There are many variations of boosting methods, which are typically differentiated based upon their loss function $g(\cdot)$ and the classifier update function $f(\cdot)$. We discuss some prominent methods used often in computer vision. The original boost used a constant value for the classifier update function $f(\cdot) = 1$ and an exponential loss function $g(\alpha_t I_i) = \exp\left(-\alpha_t c_i h_t(\mathbf{x}_i)\right)$ [170, 171]. It was shown that such technique performed marginally better than the random techniques used for selecting the features from a codebook. However, the performance of boosting method was greatly enhanced by the introduction of adaptive boosting (Adaboost) [2, 170-173]. Here, the main difference is the classifier update function $f(\varepsilon_t) = 0.5 \ln\left(\left(1 - \varepsilon_t\right)/\varepsilon_t\right)$. Since the value of $f(\cdot)=0$ implies no further optimization, the termination condition is set as $\varepsilon_t \geq 0.5$. This boosting method was adapted extensively in the object detection and recognition field. Though it is efficient in avoiding the problem of over-fitting, it is typically very sensitive to noise and clutter.

A variation on the Ada-boost, Logit-boost [170, 171, 174] used similar scheme but a logistic regression function based loss function, $g(\alpha_t I_i)=\ln\left(1+\exp\left(-\alpha_t c_i h_t(\mathbf{x}_i)\right)\right)$. As compared to the Ada-boost, it is more robust to the noisy and cluttered scenarios. This is because as compared to the Ada-boost, this loss function is flatter and provides a softer shift towards the noise images.

Another variation on the Ada-boost is the GentleAda-boost [7, 9, 170, 171], which is similar to Ada-boost but uses a linear classifier update function $f(\varepsilon_t)=\left(1-\varepsilon_t\right)$. The linear form of the classifier update function ensures that the overall update scheme is not severely prejudiced.

In order to understand and compare the four boosting schemes, we present the plots between the error $\varepsilon$ and the loss function (which also incorporates the classifier update function through $\alpha$) for the four boosting schemes in Fig. 6. Fig. 6(a) shows the value of loss function when the chosen classifier gives the correct inference for an image. If the classifier is weak (high error) and yet generates a correct inference for an image, that image is boosted so that the classifier gets boosted. Similarly, Fig. 6(b) shows the plot when the chosen classifier generates incorrect inference for an image. If the classifier is strong (low error $\varepsilon$) and still generates an incorrect inference for an image, the classifier can be suppressed or weakened by boosting such image.

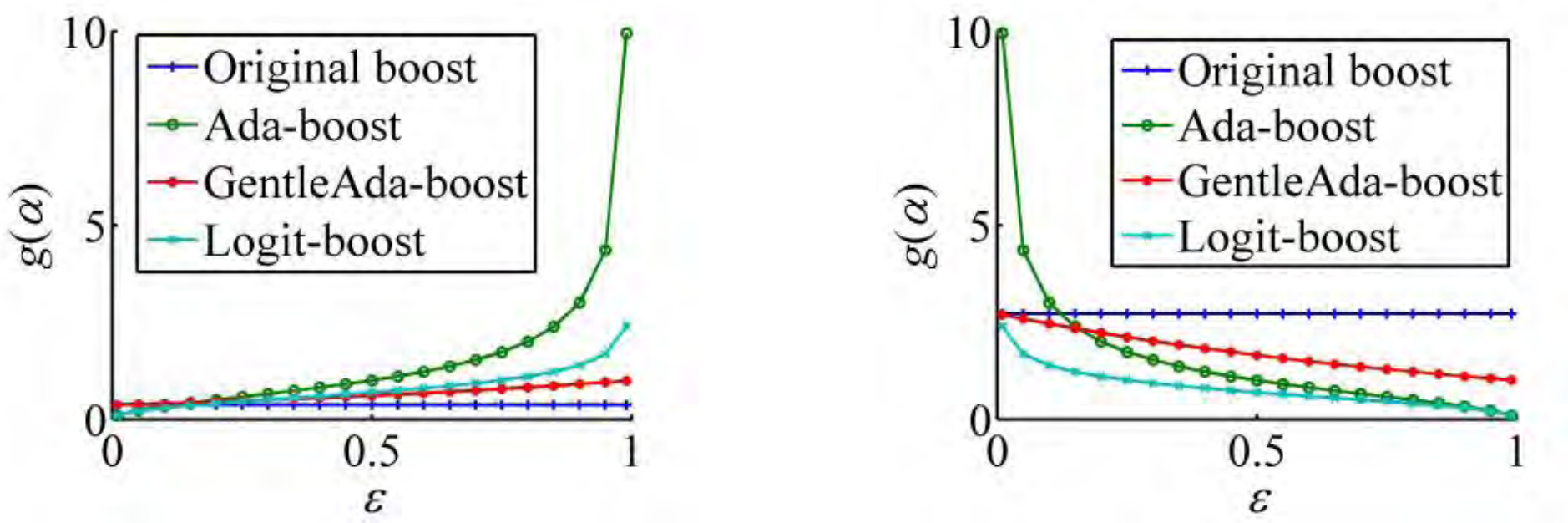

(a) Loss function when the inference is correct (b) Loss function when the inference is incorrect

Fig. 6: Comparison of boosting techniques. (a) Loss function when the inference is correct (b) loss function when the inference is incorrect.

It is evident that the desired property is not emulated well by the original boosting, which explains its slight (insignificant) improvement over the random selection of classifiers. On the other hand, Ada-boost is too strict in weakening or boosting the classifiers. Logit-boost and GentleAda-boost demonstrate a rather tempered performance, among whom, evidently Gentle-boost is the least non-linear and indeed the most gentle in weakening or boosting the classifiers. However, in our opinion, Logit-boost is the best among these methods precisely because of its combination of being gentle as well as non-linear. Due to the non-linearity, it is expected to converge faster than the GentleAda-boost and due to its gentle boosting characteristic, it is expected to be more robust than Ada-boost for noisy and cluttered images, where wrong inferences cannot be altogether eliminated.

The convergence of boosting techniques (except the original one) discussed above can be enhanced by using a gradient based approach for updating the weights of the images. Such approach is sometimes referred to as the Gradient-boost [171, 175, 176]. However, this concept can be used within the framework of most boosting approaches. Similar contribution comes from the LP-boost (linear programming boost) methods [33, 170], where concepts of linear programming are used for computing the weights of the images. In both the schemes, the iteration (step 2 of Fig. 5) is cast as an optimization problem in terms of the loss function, such that the convergence direction and rate can be controlled. Such schemes also reduce the number of control parameters and make boosting less sensitive to them.

A recent work by Mallapragada [177], Semi-boost, is a very important improvement over the existing boosting methods. While the existing boosting methods assume that every image in the validation dataset is labeled, [177] considers a validation dataset in which only a few images need to be labeled. In this sense, it provides a framework for incorporating semi-supervised boosting. In each iteration (step 2 of Fig. 5), two major steps are done in addition to and before the mentioned steps. First, each unlabelled image is pseudo-labeled by computing the similarity of the unlabelled images with the labeled images, and a confidence value is assigned to each pseudo-label. Second, the pseudo-labeled images with high confidence values are pooled with the labeled images as the validation set to be used in the remaining steps of the iteration. As the strong features are identified iteratively, the pseudo-labeling becomes more accurate and the confidence of the set of unlabelled data increases. It has been shown in [177] that Semi-boost can be easily incorporated in the existing framework of many algorithms. This method provides three important advantages over the existing methods. First, it can accommodate scalable validation sets (where images may be added at any stage with or without labeling). Second, since semi-boost learns to increase the confidence of labeling the unlabelled images, and not just fitting the features to the labeled data, it is more efficient in avoiding over-fitting and providing better test performances. Third, though not discussed in [177], in our opinion, the similarity and pseudo-labeling schemes should help in identifying the presence of new (unknown) classes, and thus provide class-scalability as well.

Though another recent work by Joshi [98] tries to attack the same problem as [177] by using a small seed training set that is completely labeled in order to learn from other unsupervised training dataset, his approach is mainly based on support vector machine (SVM) based learning. It may have its specific advantages, like suitability for multi-class data. However, semi-boost is an important improvement within the boosting algorithms, which have wider applicability than SVM based learning methods.

Another important method in the boosting techniques is the Joint-boost [2, 75], first proposed in [37, 178]. It can handle multi-class inferences directly (as opposed to other boosting techniques discussed above which use binary inference for one class at a time). The basis of joint boosting is that some features may be shared among more than one class [37, 178]. For this, the error metric is defined as $\varepsilon_j = \sum_{\kappa} \sum_{i} \left( {}^{\kappa}w_{i,t} \, {}^{\kappa}I_i \right)$, where $\kappa = 1$ to $K$ represents various classes, and the inference ${}^{\kappa}I_i$ is the binary inference for class $\kappa$. Thus, instead of learning the class-specific strong features, we can learn strong shared features. Such features are more generic over the classes and very few features are sufficient for representing the classes generically. Typically, the number of sufficient features is the logarithmic value of the number of classes [37, 178]. However, better inter-class distances can be achieved by increasing the number of features. Even then the number of features required for optimal generality and specificity is much lesser than boosting for one class at a time. Such scheme is indeed very beneficial if a bag of words is used for representing the object templates. Joint boost has also been combined with principal component analysis based system in [102] to further improve the speed of training.

## 1.3 Our approach

The ultimate objective of the current research is to develop a fast algorithm that can assist robots in detecting and recognizing the objects from real images. Our proposed approach is illustrated in Fig. 7. The key points of our approach are as follows:

1) We shall use a combination of edge fragment features, region features, and geometric shape features. We propose to use geometric shape cues as an important tool for identifying good features. Such cues shall be primarily used in identifying the features in edge map and textural region features in a non-stochastic manner. The shape cues that are strong and persistent can also be used as features themselves. In this context, we propose to use elliptic, linear and quadrangle/triangle cues for shapes, since these shapes are most widely present in both manmade and natural objects. We have completed the preliminary work of indentifying linear and elliptic cues, the details of which can be found in chapters 2 and 3.

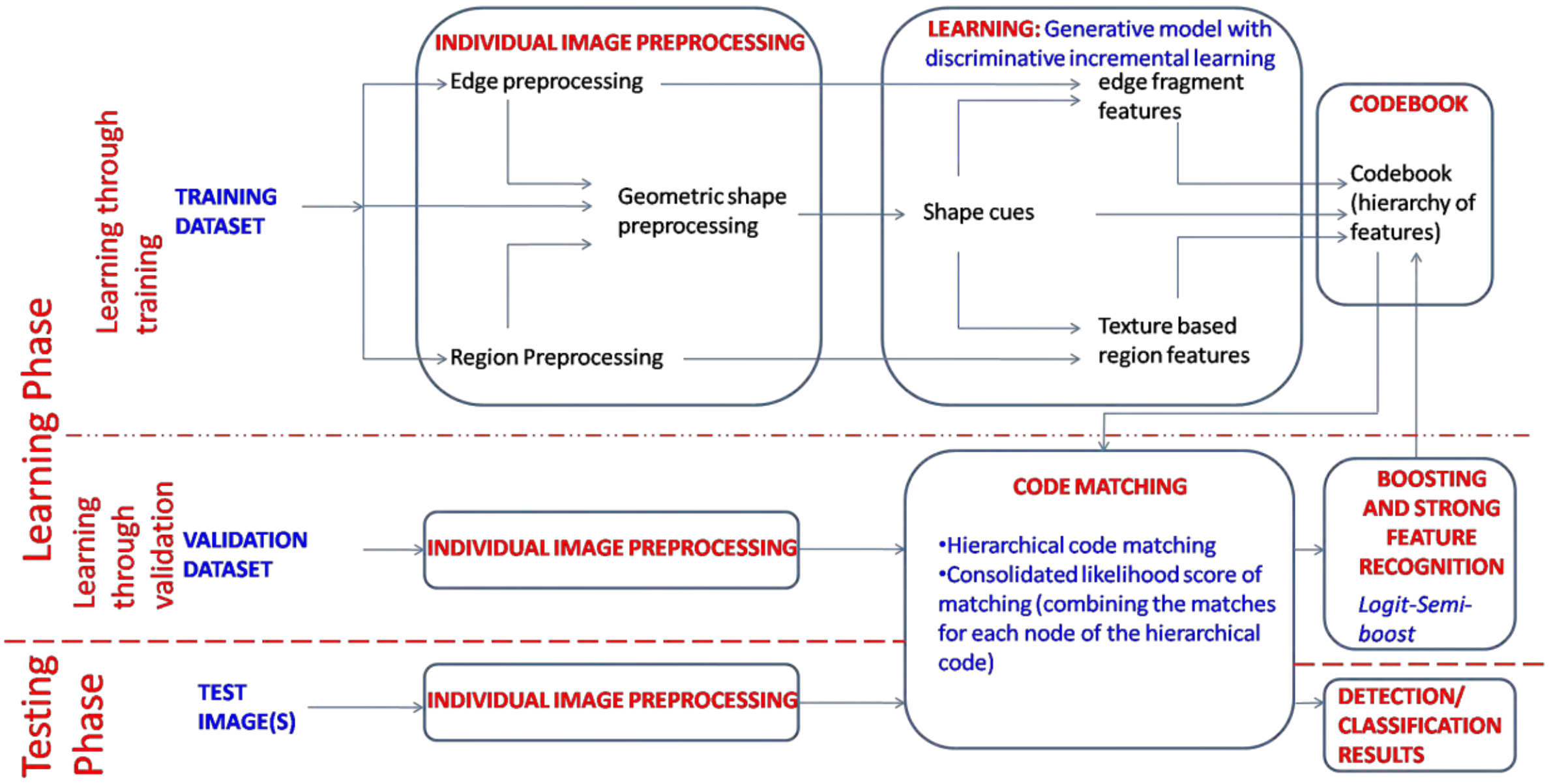

Fig. 7: Block diagram of the proposed object detection/recognition approach.

2) We shall use a generative model with discriminative and incremental learning. For this Bayesian classifier can be augmented with Semi-boosting (which is both discriminative and incremental). However, instead of using Semi-boost in its original Ada-boost framework, we propose to adapt Semi-boost for Logit-boost framework. The proposed semi-boost shall be used in the validation phase as well.

3) We propose to use hierarchical codebook, in which the higher nodes will be the strongest generic features for the object class, and as we move downwards, the features will become more specific and weaker. Thus, while matching, we first match the strong generic feature and then move to more and more specific features. In addition, since the generic strong features are on the top, they can be considered as guidance for building class-subclass hierarchies among the object classes.

4) Since the proposed hierarchical code shall require a top-down matching, the likelihood ratio will be a combination of the likelihood ratios and the matching scores (match values) at all the nodes.

The novelty of our approach is as follows:

1. While others have used at most two feature types, we propose to use three feature types: edge fragments, region features, and geometric cues.

2. While others have used geometric features [16], they have not utilized the full potential of the geometric cues. In addition to using the geometric features, we shall use them for finding more robust edge fragments and region features.

3. In the previous methods that use two feature types, the object template usually has a clear segregation between the feature types. In our case, the hierarchical object template does not necessarily contain a clear segregation of the feature types. Any path in the hierarchical tree or any level may contain all types of features.

4. The object template is the most powerful highlight of the proposed method. Unlike any other method, the hierarchy in our object template is designed on the basis of increasing likelihood of the object class. This gives us various novel advantages over other methods. First, a natural progression along the object template increases the confidence in detecting the object class. Second, there are multiple possible traversals, each of which may independently indicate the presence of class. Thus, the detection/recognition process is not dependent on the presence of one particular feature and is very robust to occlusion and noise. Third, this object template has good generalization as well as discriminative capabilities. We expect that the combined effect of the generalization and discrimination of the proposed object template will be much better than all the existing methods. Fourth, the proposed object template is amenable to dynamic learning, expansion, and pruning. Thus, the object template can be improved over the time and online learning will be possible.

5. We shall propose a new Semi-Boost scheme based upon Logit-Boost and Bayesian classification. Such method will have more generalization capabilities and exhibit more robustness to noise than the original Semi-Boost and Logit-Boost. It shall also be more discriminative than Bayesian classification and shall provide a direct method for calculating the conditional likelihoods of the various features. Thus, using Semi-Boost, the hierarchical object template can be formed in a direct manner, without extensive extra calculations. The use of Semi-Boost in training and validation phases shall also reduce the constraint on the training dataset's size and the amount of supervision.

6. The combination of the proposed object template and the learning scheme (Semi-Boost in Bayesian classification framework) makes the object recognition/detection method less dependent upon the choice of training dataset and does not require large supervised training dataset.

7. The unique but simple inference scheme ensures that the decision takes into account each path in the hierarchical object template as well as the overall object template. The scheme ensures that the potential of the proposed object template is indeed utilized.

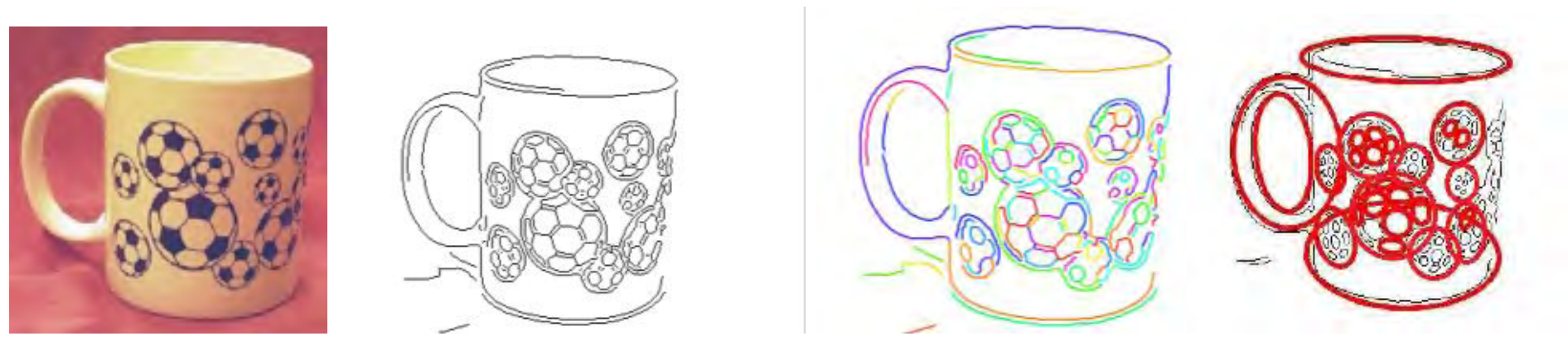

(a) Example 1: A cup with soccer ball motif

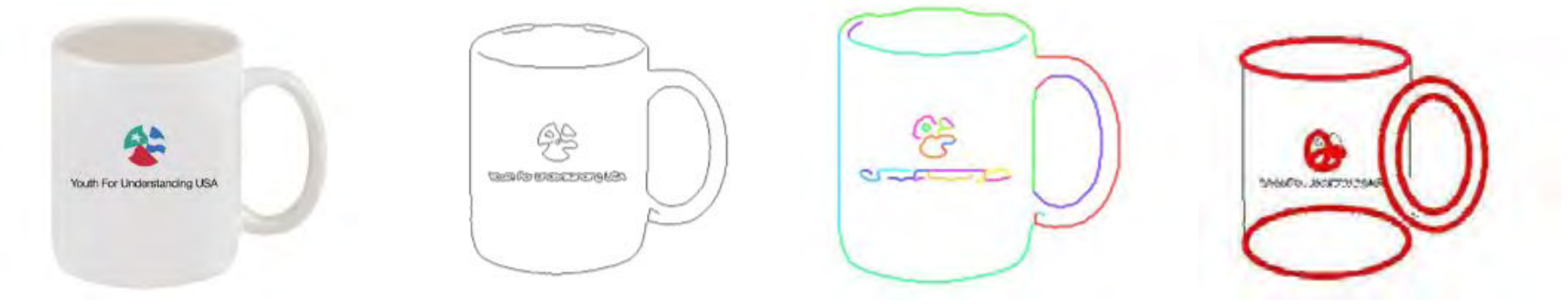

(b) Example 2: A cup with a logo

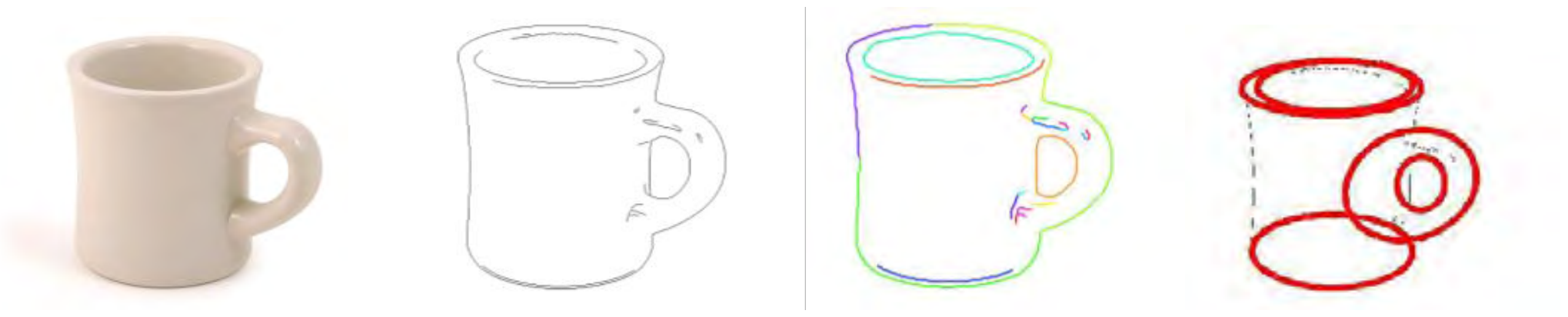

(c) Example 3: A simple cup.

Fig. 8: Using handles as cues for cups. Sub-figures (a)-(c) show three cups that are very different from each other. It is evident that though the edge map and edge contours are useful for detecting the cups, the elliptic shapes represented by the handles and mouths of the cups are consistent features that can be used as cues for recognizing the cups.

This report focuses on the geometric shape cues. With the help of the geometric cues, we intend to derive robust and reliable object features (applicable to vast variety within a class, as well as able to distinguish the objects from other classes). For example, in cups, the cues can be elliptic shapes corresponding to the mouth and handles as the most general occurrences. The cue for handle can be used to increase the inter-class separation between cups and tumblers. See Fig. 8 for example. The three cups shown in Fig. 8 are very different from each other. It is evident that though the edge map and edge contours are useful for detecting the cups, the elliptic shapes represented by the handles and mouths of the cups are consistent features that can be used as cues for recognizing the cups.

It is encouraging to notice that linear and elliptic shapes are the most common and widely present shapes in our real world, both in natural and manmade objects. Thus, the lines and elliptic shapes can be used as cues for various object categories. See Fig. 9 and Fig. 10 for a few examples.

Another focus of the current work is to extract the edge fragments that are continuous in curvature so that the efficiency of such edge cues can be increased. This work has also been completed and is reported in chapter 2. The remaining portion of the proposed method is discussed in detail in the future work (chapter 4).

(a) Original Image (b) Linear cues

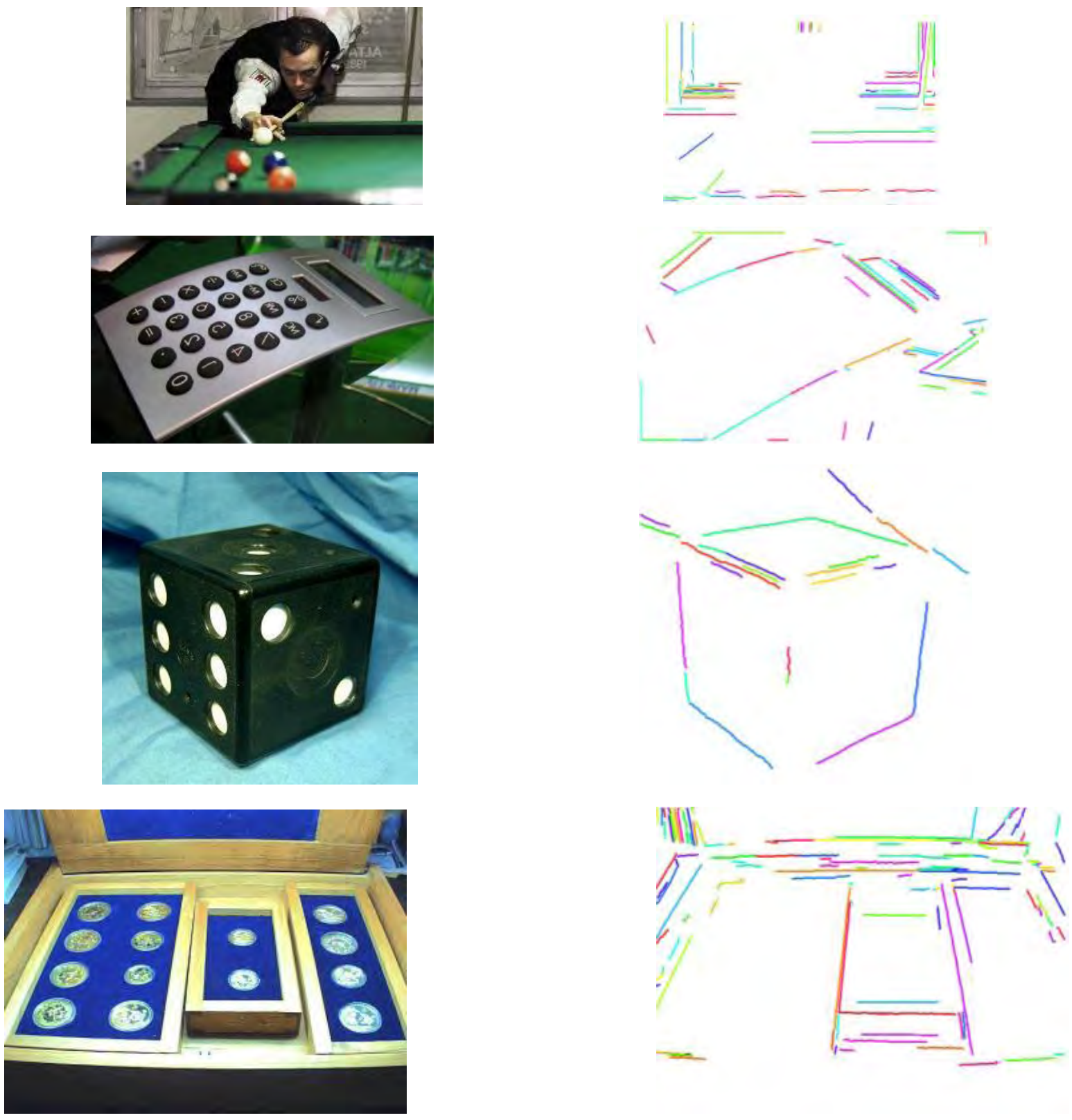

Fig. 9: Linear shapes as primary cues: a few examples (elliptic cues have not been shown). It is evident that the linear cues can be combined to detect quadrangles and other primitives.

(a) Original Image (b) Elliptic cues

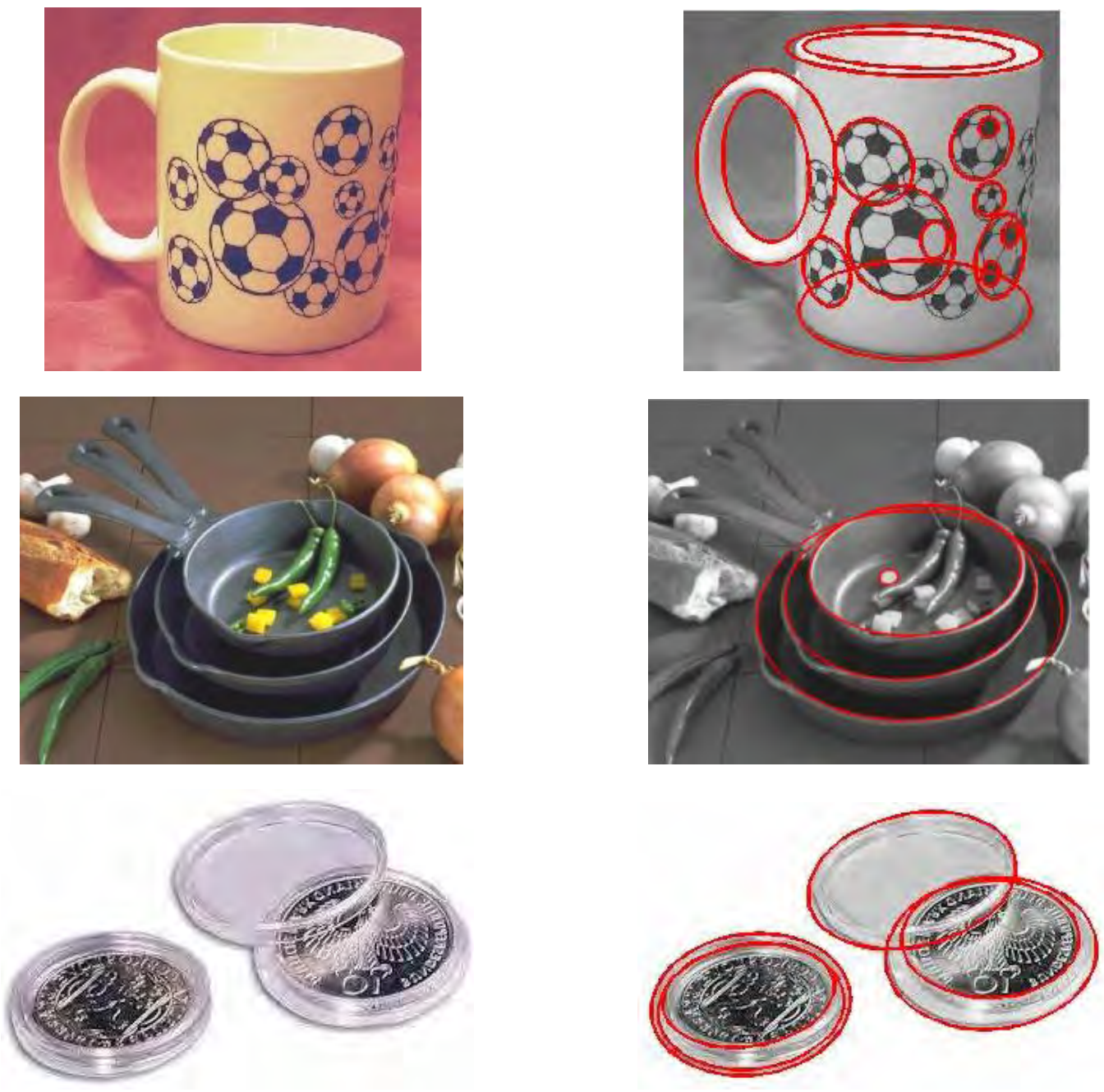

Fig. 10: Elliptic shapes as primary cues: a few examples. The examples show that the elliptic shapes can be used as cues for object recognition.

## 1.4 Outline of the report

The report is organized in the following manner. The preliminary work regarding the edge and line cues is presented in chapter 2. The preliminary work for elliptic shape based cues is presented in chapter 3. The future path and conclusion is presented in chapter 4.

# 2 Preliminary Work 1: Edge and Line Cues

As discussed earlier, edge based cues are widely used for object recognition. However, the edge map obtained by a typical edge detection algorithm is often too crude to be used directly for object recognition. Most methods may use edge pixels directly, as in most Hough transform based methods and distance based methods. However, using edge pixel points independently is restricted in its capabilities as it does not use the information regarding the connectivity of edge pixels in forming a continuous boundary, called edge contours here onwards. Using the information of the connected edge pixels or edge contours (i.e., edges representing boundaries or sections of boundaries) should improve the possibilities of generating stronger cues and help in subsequent shape based cues. This has motivated the current research to use edge contours, rather than edge pixels as the primary data.

The supporters of the edge pixels based methods might argue that various objects and their boundaries may intermingle due to overlap or occlusion, which may result into a situation that a continuous edge is a combination of two sections of boundaries that may actually belong to more than one object. However, it can also be counter argued that an edge pixel alone, without the information of its connectivity, cannot be associated with any shape or shapes and may be considered numerically part of any possible boundary formed by a combination of other pixels arbitrarily chosen from the image. Further, even though an edge may be a combination of two or more sections of boundaries, there is a possibility of determining such situations and breaking the edge at such predicted situations. This forms the next step in the current research work. The presence of two separate sections of boundaries in a single edge is often manifested in the form of sudden changes in the curvature of the edge. Since it is well known that the curvature of any elliptic shape changes continuously and smoothly, it is logical to break an edge where a sharp turn or an inflexion in the curvature is observed.

Another argument given in support of using the edge pixels directly is that the process of extracting the edges might be time consuming. However, there are methods to derive the connected edges in a computationally efficient manner. Further, various steps in the proposed method have the computations of the order of number of edge contours, which are far lesser than the number of edge pixels. Thus, a little computation invested in deriving the connected edges results into immense reduction in computational burden in the complete scenario.

The block diagram of the preliminary processing is shown in Fig. 11. Each of the blocks is explained in greater detail in the subsequent sections.

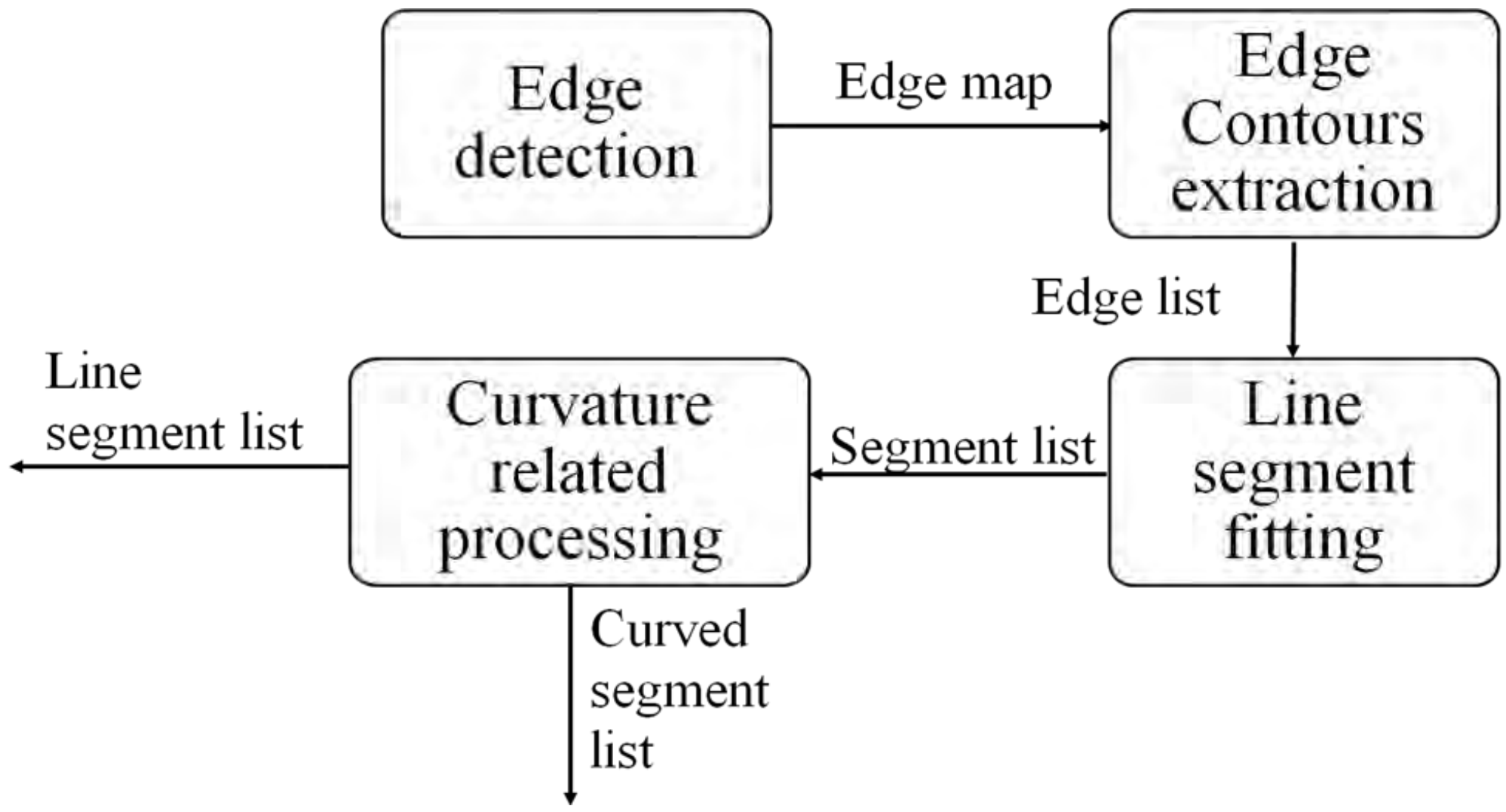

Fig. 11: Block diagram of the preliminary processing for line and curve segment cues

## 2.1 Edge detection

### *2.1.1 Process for obtaining edge map*

Before the edge may be extracted, the real image has to be converted to gray scale (if it is a color image). It is preferable in most cases to perform histogram equalization on the gray images. This is because performing histogram equalization distributes the image pixels over all the possible gray values, thus, effectively improving the contrast and enhancing the boundaries. Other image enhancements may also be applied in order to improve the quality of information in the contours. However, we refrain from discussing these issues as this is outside the scope of the current work. Further, we have not used any form of image enhancements except histogram equalization. After this, Canny edge detector is applied to the histogram equalized gray image. We have chosen the control parameters for Canny edge detector as follows: low hysteresis threshold $T_L = 0.1$ , high hysteresis threshold $T_H = 0.2$, and standard deviation for the Gaussian filter $\sigma = 1$. This choice of control parameters works satisfactorily for most of the images. After applying the Canny edge detector, the edge map obtained is a binary image, in which the edge pixels are given the value '1' and the non-edge pixels are assigned value '0'.

## 2.2 Edge contour extraction

To extract edge contours from the edge map, edge pixels have to be linked together to form lists of sequential edge points (with one list for each edge contour segment). In this section, we discuss the method to form edge contours from the edge pixels.

For our purpose, we are interested in continuous, non-branching edge contours. Thus, a contour segment should start (or stop) either at an open end of the contour or at a junction of two or more contours. If a junction is encountered, the current edge contour has to break there and new edge contours have to be initiated from that point. The idea is shown in Fig. 12. Below we discuss the process of linking the edge pixels and ensuring that contour does not have any junctions.

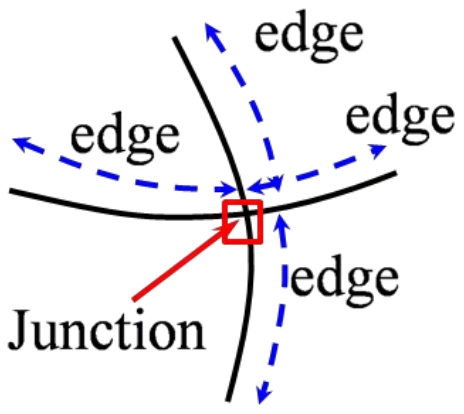

Fig. 12: Presence of junction in an edge map.

### 2.2.1 *Removing isolated pixels*

Removing isolated edge pixels (if any) serves two purposes. First, it removes spurious pixels that may have arisen due to noise. Second, removing such pixels in the first step reduces the memory and computational requirement for all the proceeding edges. Hence, the isolated pixel is seen as the edge contour with only one pixel point. The mask shown in Fig. 13 (a) of size $3 \times 3$ can be used to detect the isolated pixels.

| | | |
|---|---|---|
| 0 | 0 | 0 |
| 0 | 1 | 0 |
| 0 | 0 | 0 |

(a)

| | | |
|---|---|---|
| 0 | 0 | 0 |
| 0 | 1 | 0 |
| 0 | 1 | 0 |

(b)

Fig. 13: Masks: (a) Mask for detecting an isolated pixel (b) Scenario in which end of a contour may occur. Any possible rotations of this window represents end of edge contour.

### 2.2.2 *Finding ends and junctions*

Whether an edge pixel is an end pixel can be easily determined using a $3\times3$ pixel window. In a $3\times3$ pixel window centered at the considered edge pixel, if only one neighboring pixel is '1' and the rest neighboring pixels are '0', then the considered edge pixel is an end pixel (see Fig. 13(b)). It is evident that there are eight possible scenarios, obtained by the rotation of the mask shown in Fig. 13(b).

On the other hand, there are numerous scenarios in which junctions may occur. A few possible scenarios are shown in Fig. 14. Kovesi [179] has used a very simple and efficient technique for taking into account all such scenarios for junctions. His technique is also directly useful for finding the end pixels of the edge contours.

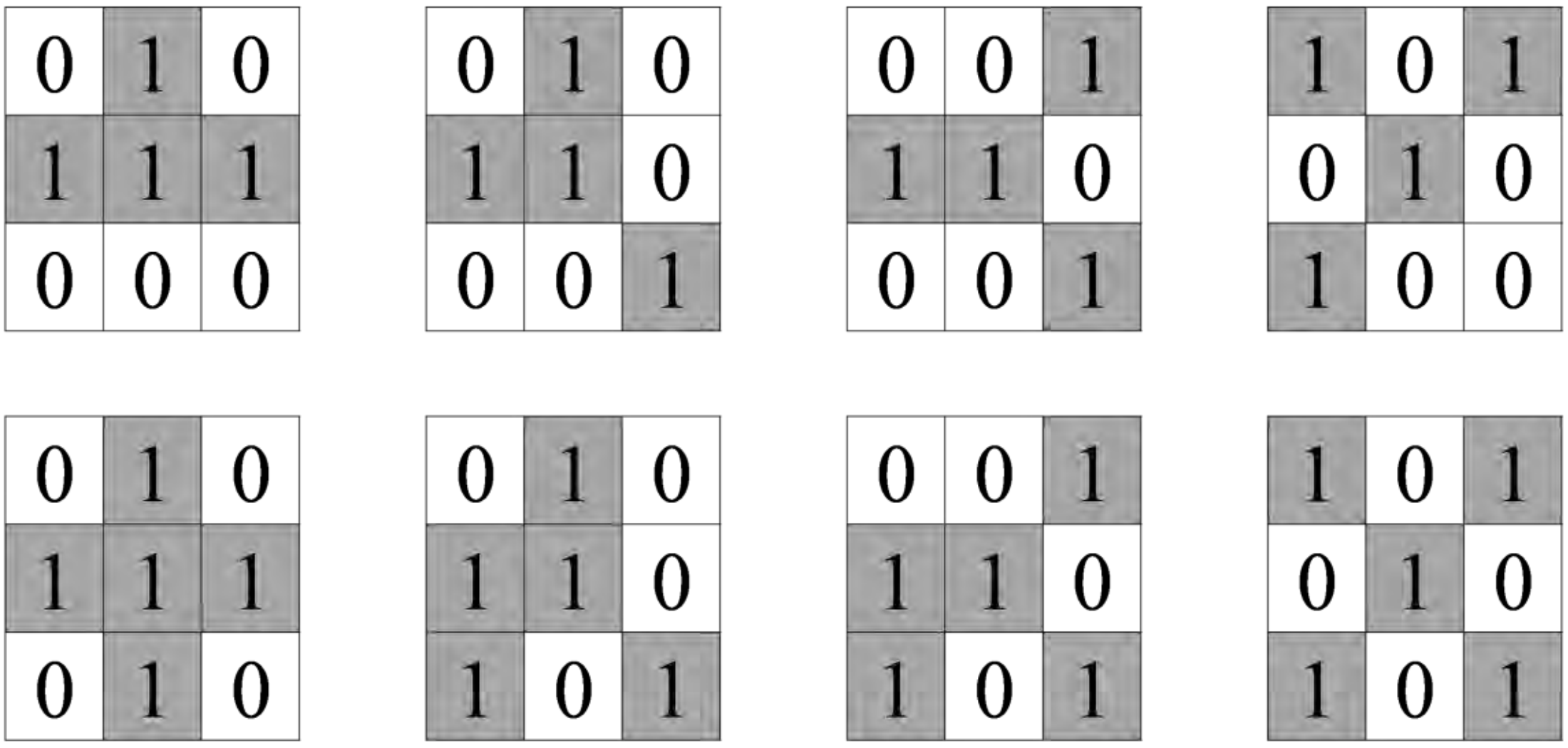

Fig. 14: Primary masks for junctions: All the above and every rotation of each of the above, will represent junctions. However, the above list is not exhaustive.

To explain his method, it shall be helpful to annotate the pixels in a $3\times3$ pixel window $w$ centered at an edge pixel. Let the annotations of the pixels in $w$ be as shown in Fig. 15(a). Then, we can define two vectors as below:

$$\bar{u} = \left[w(2) \quad w(3) \quad w(4) \quad w(5) \quad w(6) \quad w(7) \quad w(8) \quad w(1)\right], \tag{3}$$

$$\bar{v} = \left[w(1) \quad w(2) \quad w(3) \quad w(4) \quad w(5) \quad w(6) \quad w(7) \quad w(8)\right]. \tag{4}$$

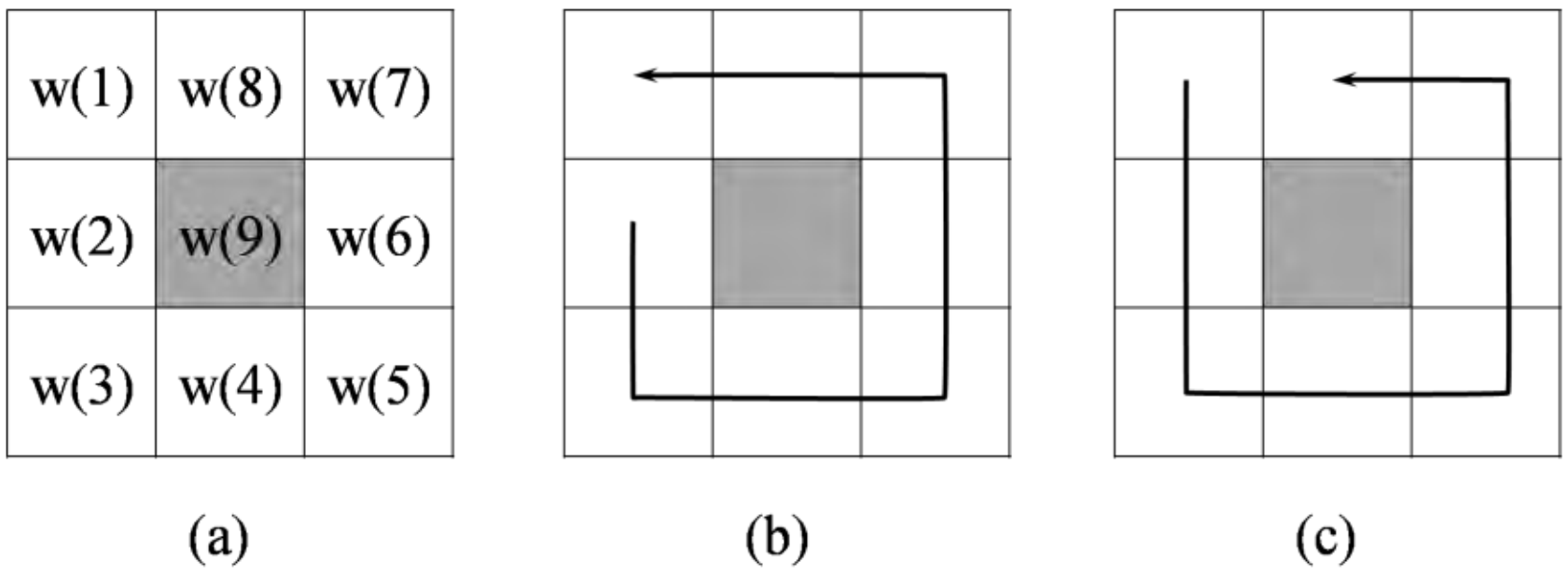

Fig. 15: Annotation of a $3\times3$ pixel window and representation of the vectors $\bar{u}$ and $\bar{v}$ in equations (3) and (4).

The representation of these vectors in the pixel window $w$ is shown in Fig. 15. A parameter $b$ can be defined as below:

$$b=\sum_{i=1}^{8}\left|u_i-v_i\right| . \tag{5}$$

The parameter $b$ has a value more than or equal to 6 if there is a junction present at the considered edge pixel. Further, the parameter $b$ has a value 2 if the considered edge pixel is the end of the edge contour. Further details can be found in [179]. The edge pixels that are detected as either junctions or end points can be collected in a look up table $T$ .

### *2.2.3 Extracting the edge contours*

The method to extract the edge contours can be succinctly explained as follows. Beginning with the first end pixel listed in table $T$ , the pixels in its continuity are collected in an edge list (that represents the current edge contour) till either an end point or a junction point is encountered. Once such situation is encountered, a new edge list is begun with the next end pixel in the list. The end pixels that are encountered in this process are removed from the table $T$ so that no edge contours are repeated. For a junction point, once all the edge contours meeting at the junction point are traversed, it is removed from the table $T$ .

## 2.3 Line segment fitting and extraction of linear cues

By representing the edge contours using piece-wise linear segments, we primarily get the following benefits:

1) This technique gives good representation of the contour using far lesser number of points.
2) The line segments can be very easily extracted and used as line based cues for object recognition and classification.

3) Line segments are less sensitive to problems related to image digitization. For example, since the contours are formed using pixels, there are often drastic changes in the slope and other curvature related aspects of contour over two or three adjacent pixels of a curved contour, whereas the slope and curvature of a line are computed from more pixels within a line and are less sensitive to noise.

Here we adopt the Ramer–Douglas–Peucker algorithm [180, 181] to approximate a curve into a set of line segments. Let us consider an edge contour $e = \{P_1 \quad P_2 \quad \dots \quad P_N\}$, where $e$ is the edge list formed using the method in section 2.2 and $P_i$ is the $i$ th edge pixel in the edge list $e$.

The line passing through a pair of points $P_1(x_1, y_1)$ and $P_N(x_N, y_N)$ is given by:

$$x(y_1 - y_N) + y(x_N - x_1) + y_N x_1 - y_1 x_N = 0 . \quad (6)$$

Then the deviation $d_i$ of a point $P_i(x_i, y_i) \in e$ from the line passing through the pair $\{P_1, P_N\}$ is given as:

$$d_i = \left| x_i (y_1 - y_N) + y_i (x_N - x_1) + y_N x_1 - y_1 x_N \right| . \quad (7)$$

Accordingly, the pixel with maximum deviation can be found. Let it be denoted as $P_{\max}$, as shown in Fig. 16(a). The line $P_1 P_N$ can then be replaced by line segments $P_1 P_{\max}$ and $P_{\max} P_N$, as shown in Fig. 16(b). The partition process goes as shown in Fig. 16(c,d) until $d_i$ reduces to below a chosen threshold. The threshold chosen determines how close the line segments fit on to the edge contour. For most purposes, choosing a threshold of 2 pixels is sufficient. The parameter independent modification of this method can also be used for better performance [182-186] or other dominant suitable dominant point detection method [187].

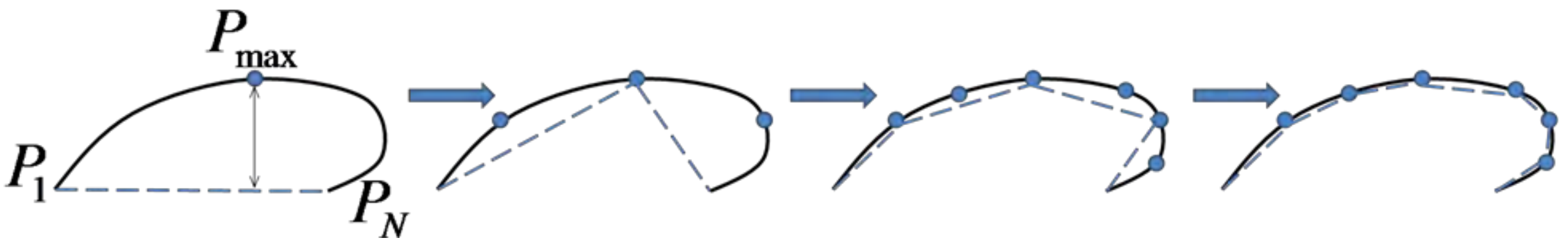

(a) (b) (c) (d)

Fig. 16: Illustration of line segment fitting on the edge contour.

The edges that are represented by a single line segment are identified as the linear edges that can be considered for linear cues. The use of these segments as cues for object detection will be in the future work.

## 2.4 Detecting edge portions with smooth curvatures

Since it is well known that the curvature of any elliptic shape changes continuously and smoothly, we intend to obtain edges with smooth curvature. The term smooth curvature is defined here as follows. A portion of an edge which does not have a sudden change in curvature, either in terms of amount of change or the direction of change, is called here as a smooth portion of edge. It should be noted that we are not performing any kind of smoothing operation. We are actually extracting curves from the existing data which are smooth as defined above.

Beginning from one end of the edge, we look for points where the curvature becomes irregular and break the edge at those points, such that every new edge formed out of this process is a smooth edge. Since we define the regularity in two terms – amount of change of curvature and direction of change of curvature, we deal with these two cases separately – and call them sharp turns and inflexion points respectively. In the following, we first develop a basic premise for dealing with these two cases and then deal with them individually.

Let us consider an edge $e$, on which line segments have been fitted using the technique presented in section 2.3. Let the collection of line segments that represent $e$ be $\{l_1, l_2, \ldots, l_N\}$. It should be noted that the index $N$ used here has nothing to do with previously defined index variable $N$ anywhere. Let the angles between all the pairs of consecutive line segments be denoted as $\{\theta_1, \theta_2, \ldots, \theta_{N-1}\}$, as shown in Fig. 17, where $\theta_i \in [-\pi, \pi]$ is the angle between $l_i$ and $l_{i+1}$, measured from $l_i$ to $l_{i+1}$.

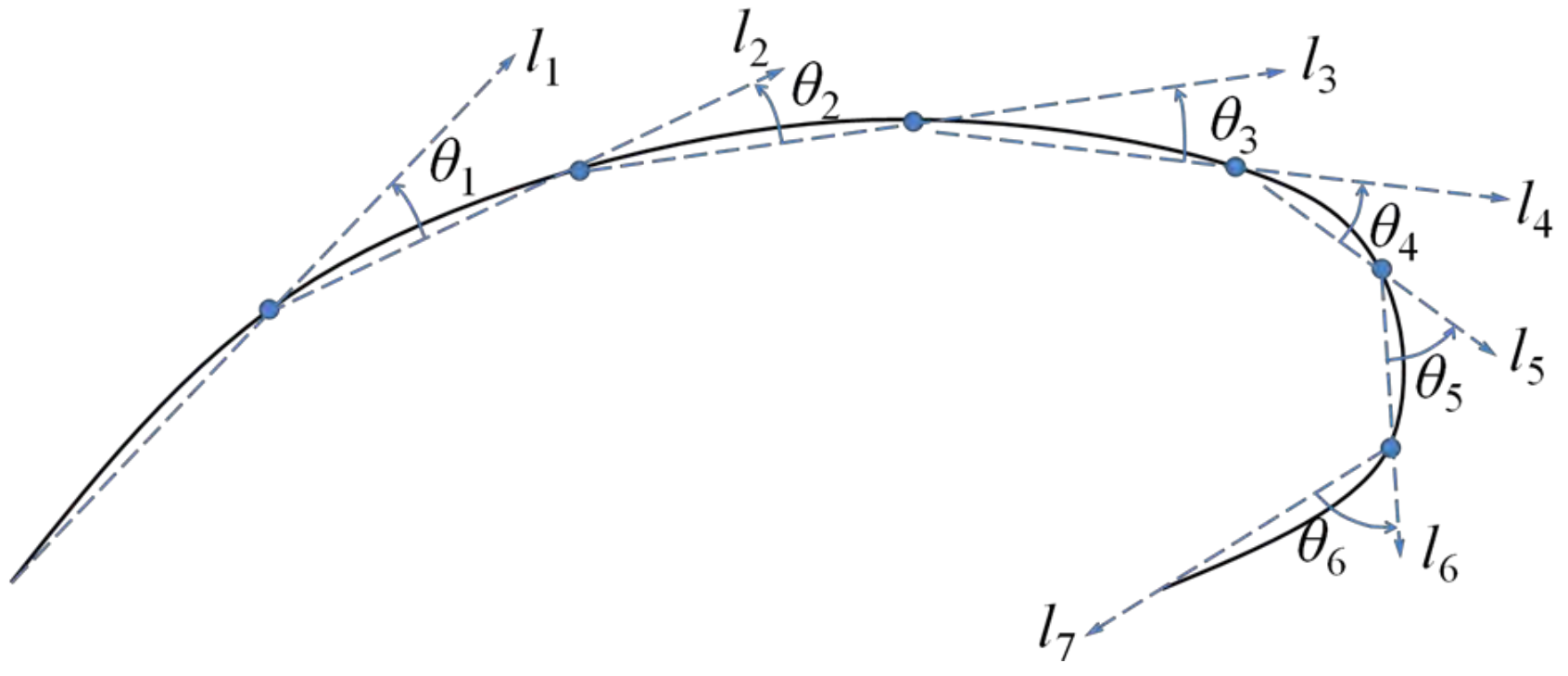

Fig. 17: Illustration of the calculation of angles for detecting edge portions with smooth curvatures

### 2.4.1 *Dealing with sharp turns*

In the sequence of the angles, $\{\theta_1, \theta_2, \ldots, \theta_{N-1}\}$, if any angle $\theta_i$ is very large, greater than a chosen threshold $\theta_0$ (say 90 degree empirically determined), then it is evident that the curvature of the edge changes sharply at such points $P_i$ (the intersection point of line segments $l_i$ and $l_{i+1}$), and the edge needs to be split at $P_i$ to ensure smooth curvature. An example is shown in Fig. 18.

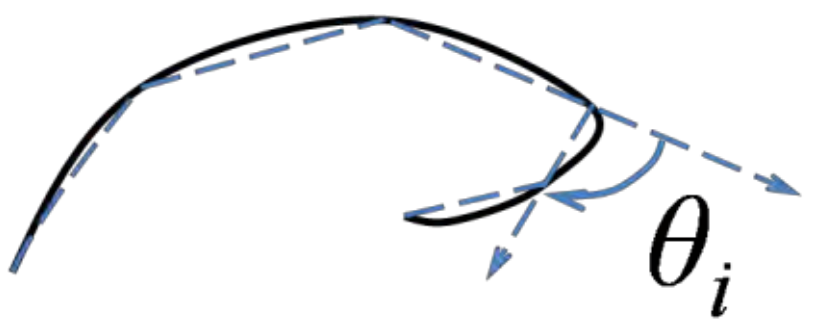

Fig. 18: Illustration of sharp turns.

### *2.4.2 Dealing with inflexion points*

From the above definition of the angles $\{\theta_1, \theta_2, \ldots, \theta_{N-1}\}$, the change in direction of curvature occurs in the change of the sign of the angles (negative or positive). Thus, we can create a Boolean sequence $\{b_1, b_2, \ldots, b_{N-1}\}$, where $b_i$ is '0' if the sign of $\theta_i$ and $\theta_1$ is the same. This Boolean sequence can be used to identify the inflexion points and decide the exact places where the edge contour should be split.

It is worth noticing that there is more than one possibility for inflexion points. These possibilities are graphically illustrated in Fig. 19. The points where the edge needs to be split are also shown in the same figure.

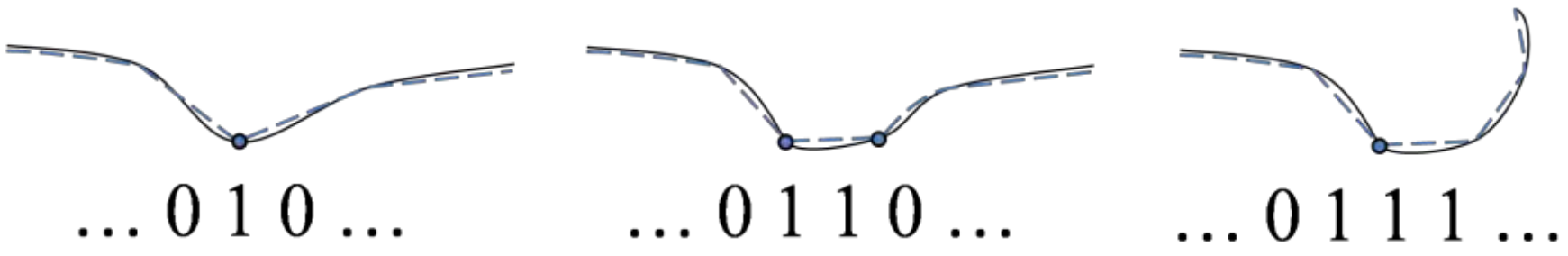

Fig. 19: Possibilities of inflexion points. In each case, the points where the contours are broken are also shown.

There may be none or many points at which an edge contour needs to be split in order to obtain smaller contours, each with smooth curvature. If there are $N'$ such points on an edge, the edge can be split at these points to form ( $N'+1$ ) smaller edges of smooth curvature.

## 2.5 Results: edge list with smooth curvature and linear cues

In this section, we present various numerical examples. The examples are presented in Fig. 20 and Fig. 21. Corresponding to an image (showed in first column), the second column shows the smooth edge list obtained by the proposed method. The linear cues obtained from the image are shown in the third column, while the last column shows the line segments corresponding to the linear cues in the fourth column.

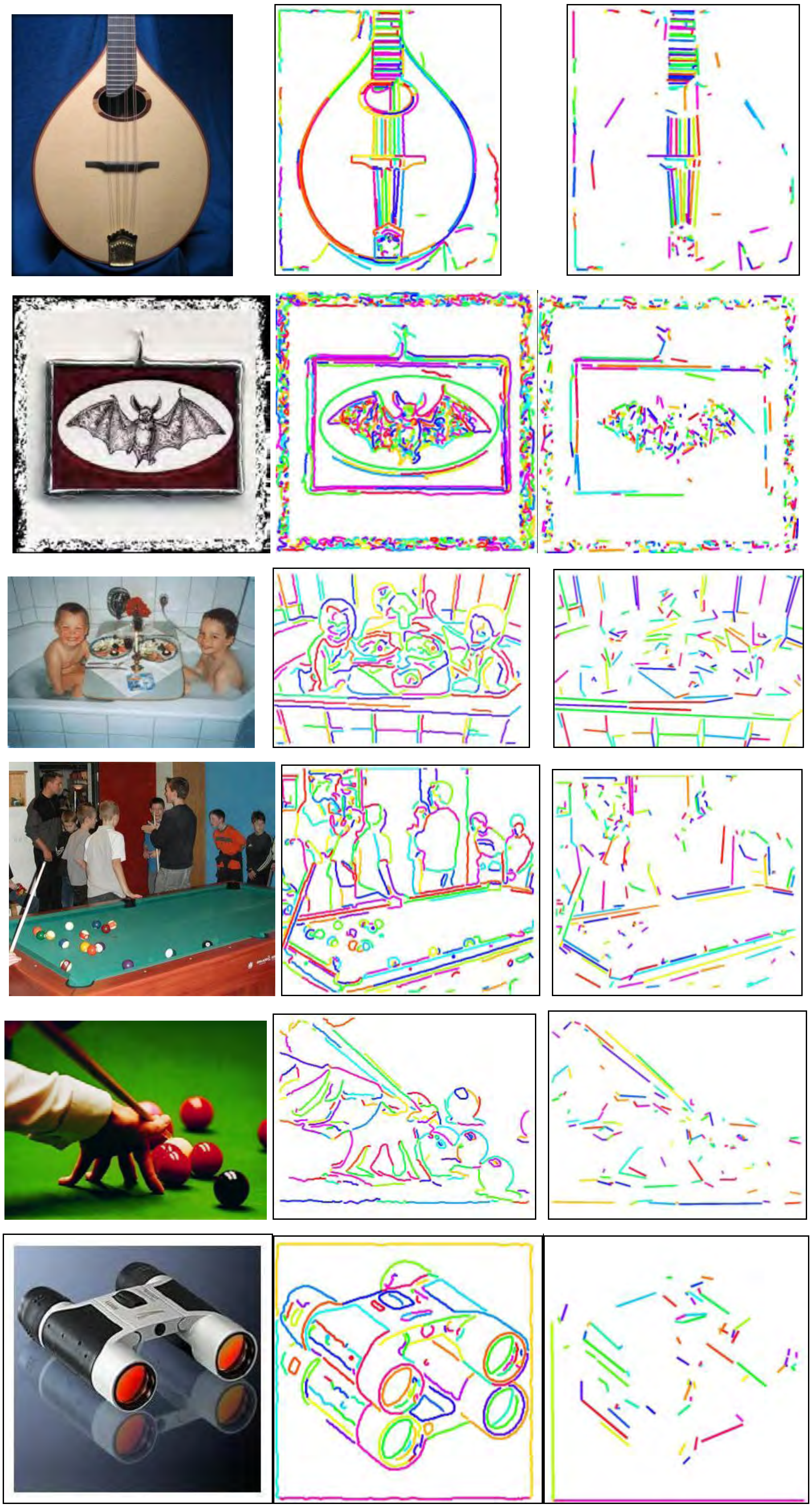

(a) (b) (c)

Fig. 20: Smooth edge list and linear cues obtained using the proposed procedure. (a) Original image (b) smooth edge list (c) linear cues.

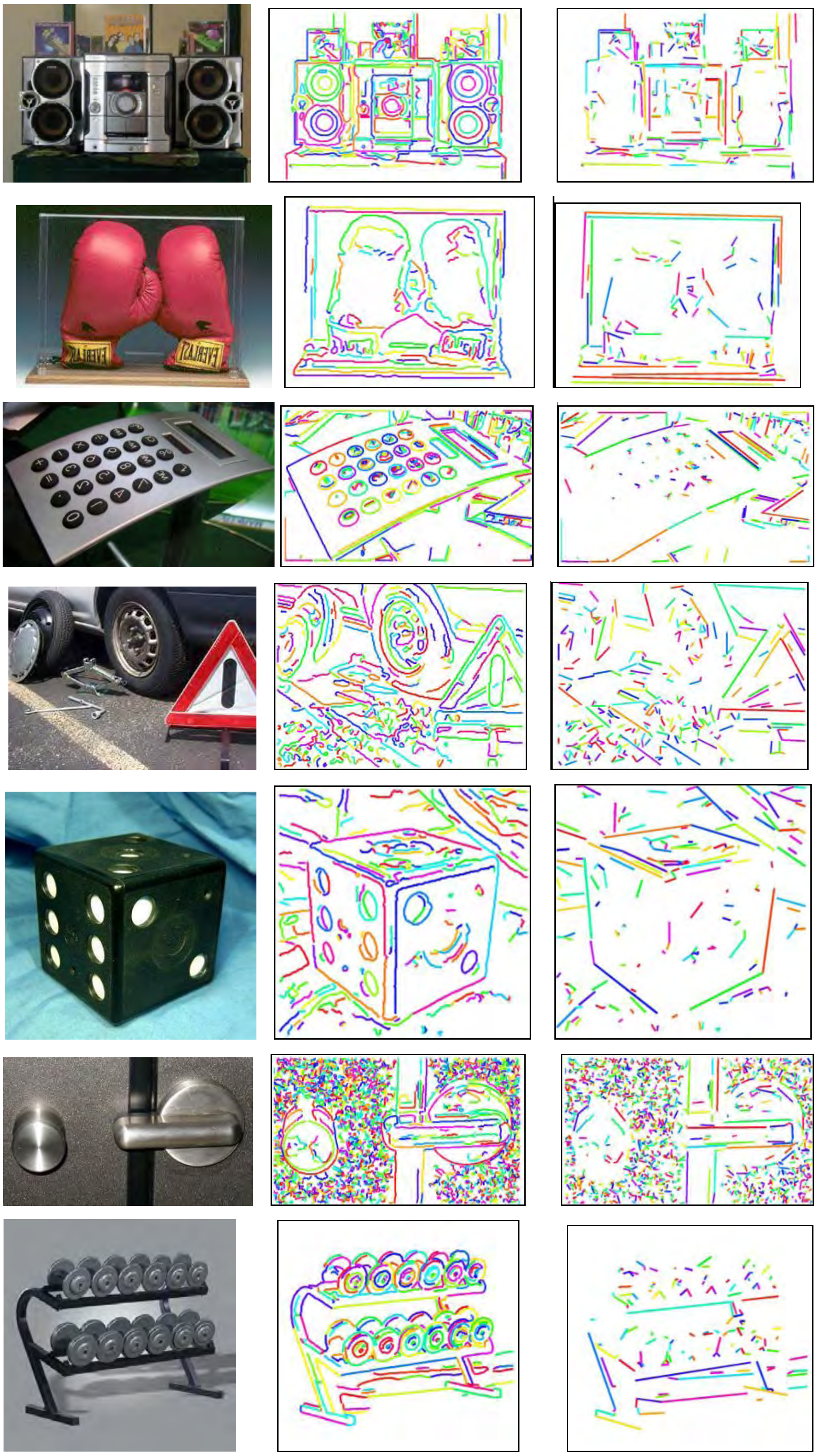

(a) (b) (c)

Fig. 21: Smooth edge list and linear cues obtained using the proposed procedure (continued). (a) Original image (b) smooth edge list (c) linear cues.

# 3 Preliminary Work 2: Ellipse Detection

## 3.1 Introduction

It has been mentioned earlier that one motivation for using elliptic cues is the common occurrence of elliptic shapes in natural and manmade objects. Another motivation towards using elliptic cues is that the mathematical structure and geometry of ellipses are well defined and easy to understand and implement. The ease of handling elliptic shapes arises from the fact that the equations governing ellipses are quadratic equations, with at most five coefficients. Further, various concepts like transformations, rotation, etc. have already been developed by mathematicians. Elliptic curves are probably the most extensively studied curves in geometry.

## 3.2 Challenges in ellipse detection in real images

One issue that can be easily envisaged in real images is that objects are usually present in overlap with each other. If the overlapping object is transparent or translucent, the boundaries of overlapped objects might still be available in the edge map (obtained after edge detection). However, if the overlapping object is opaque, the overlapped object is occluded and its incomplete boundary will appear in the edge map. Even in case of translucent overlapping object, the boundaries of the overlapping and overlapped objects will intermingle and consequently the edge map will contain incomplete parts of the complete object. Examples of overlapping and occluded elliptic objects are shown in Fig. 22. If an image is cluttered by various objects of such nature, the problem gets very complicated to handle as such scenario results in various incomplete small edges in the image.

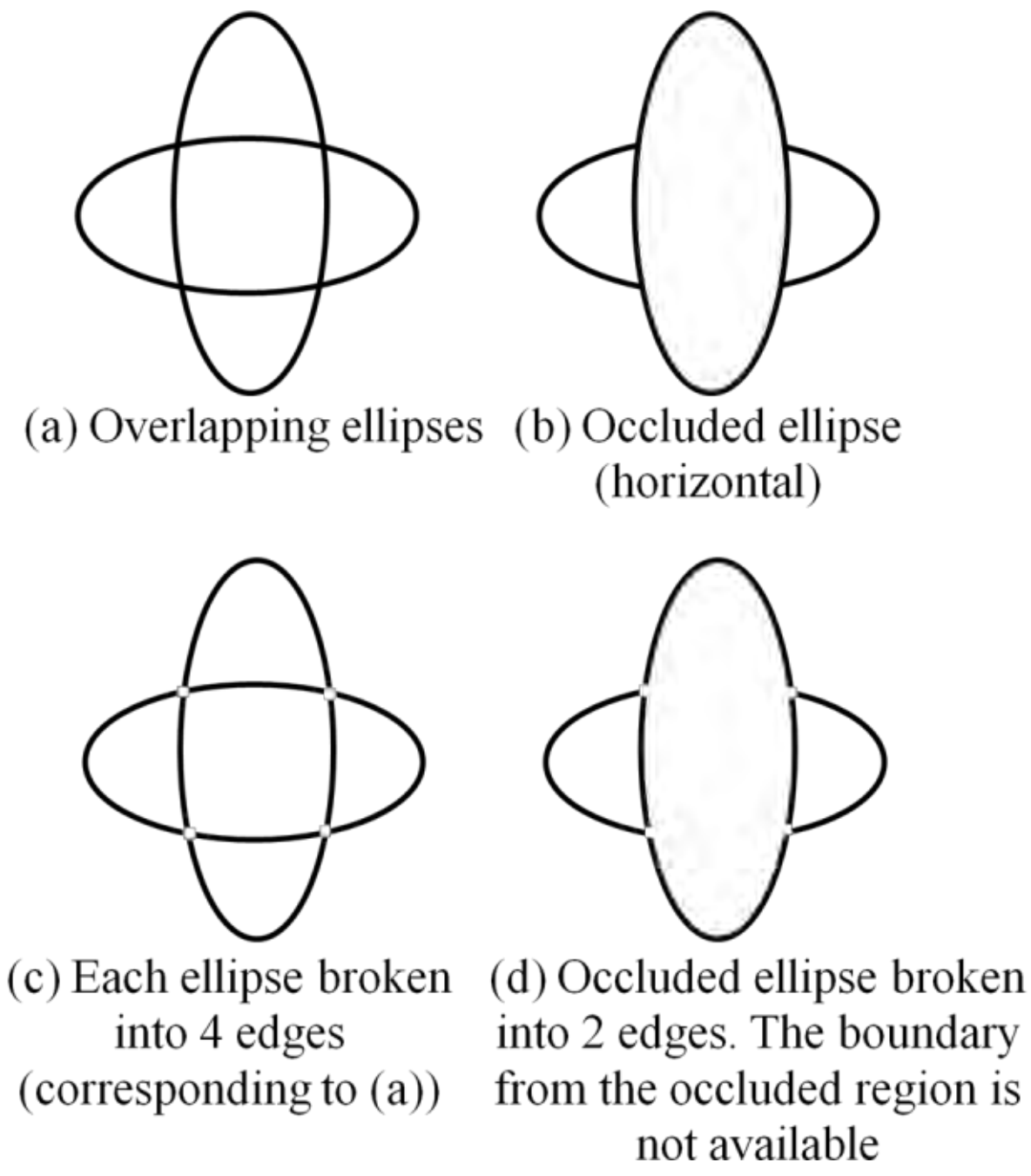

Fig. 22: Illustration of the presence of overlapping and occluded ellipses

Another problem that is encountered in real images is the deterioration of the boundary of the edge map due to the light and shadow conditions and the perspective of the object. Under different lighting conditions, boundaries in some region may be emphasized and appear sharp and clear, while boundaries in other regions might blur and deteriorate the edge in that region.

Shadow effect may blur the region in which the boundaries of two objects overlap. Due to this, the boundaries of two objects may merge and appear to be smooth.

Further, noise can appear in image due to imperfect imaging instruments and other external conditions like fog, glare, etc. Noise corrupts the quality of edge, rendering it to be non-smooth over small sections of edge, and abrupt breaks in the boundaries. One such example can be found in Fig. 23.

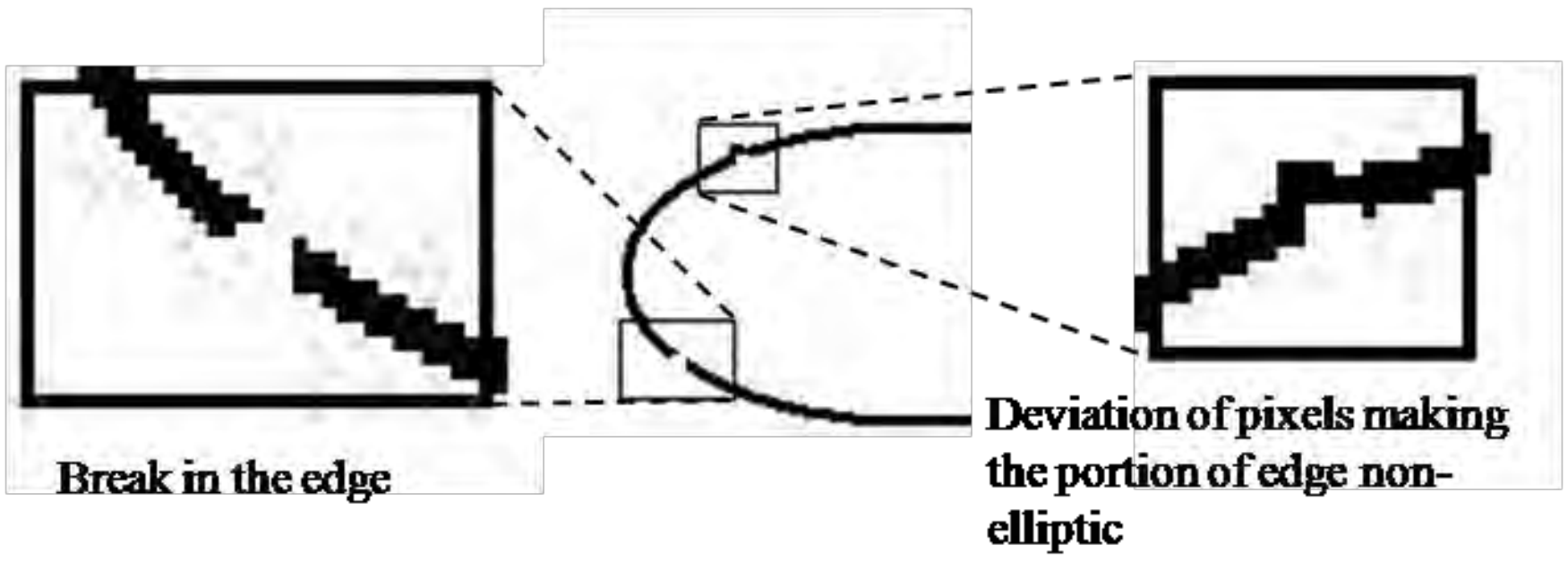

Fig. 23: Illustration of the effect of noise

In the above paragraphs, various challenges and their individual impacts are discussed. It is easily apprehensible that simultaneous presence of all these factors greatly compounds the challenges in the detection of elliptic objects. Another aspect of the considered problem is that no a priori information is available. It might have helped if the expected number of elliptic shapes, expected size of ellipses, or expected regions of focus were known a priory. However, real images may vary greatly in the scale and content, and in general such a priori information cannot be generated reliably.

From the above discussion, the main technical challenges in the detection of elliptic shapes in real images are:

1. Presence of incomplete elliptic shapes.
2. Presence of outliers (non-elliptic shapes that may be misinterpreted as being elliptic).
3. Corruption in the quality of edge
4. Lack of a priori information

### *3.2.1 Tackling incomplete elliptic shapes*

This problem has two scenarios. The first scenario is that there might be only one incomplete elliptic edge in the image. The other scenario is that there might be multiple incomplete edges belonging to one ellipse. The scenarios are illustrated in Fig. 24. In both the scenarios, every such edge should have sufficient number of edge pixels needed to generate the geometric information about the ellipse to which it belongs. However, the problem of detecting ellipses is simplified, if the relationship of different edges belonging to a single ellipse can be identified and grouped together.

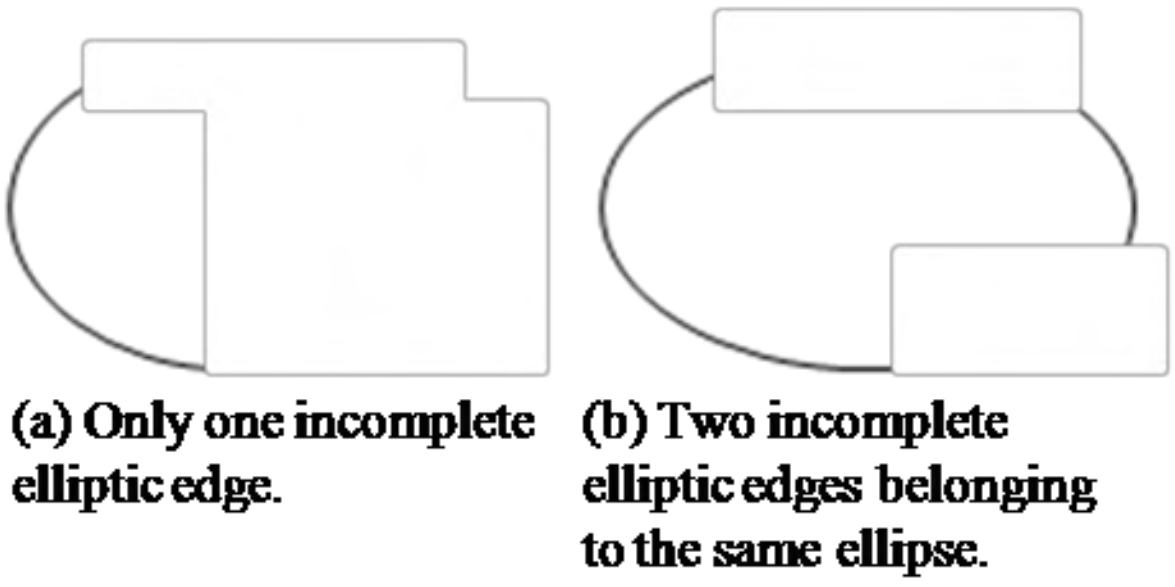

Fig. 24: Illustration of the presence of incomplete elliptic edges

The grouping can be performed on the basis of various criteria depending on the method chosen for ellipse detection. For example, if least squares fitting method is used [188-193], we may group the elliptic edges after the fitting has been performed for each edge. The grouping results can then be verified by refitting ellipse on all the edges in a group. Thus, if there is an error in the fitting results, the grouping will also be erroneous. Another possible technique is to extend the edges virtually and look for other edges that may fall in the range of the virtually extended region of the edge [194-197]. Though this approach is rigorous, its efficiency would depend on the method and direction used for extension. Also, it may turn out to be computationally expensive if the edges are numerous and small in length.

In Genetic Algorithm based techniques [198-202], edges are grouped randomly and the grouping is evaluated based on a cost function. The reduction of cost function over various evolutions results in the selection of better grouping schemes. However, genetic algorithms take large computational resources and do not guarantee the convergence to the global cost function minima.

Another possible approach, as used in our method, is to group the edges based on a center finding technique discussed in section 3.5. This approach gives us various advantages. First, since ellipse detection can be performed in two stages, viz., retrieval of centers and retrieval of the rest of the parameters, we can make our algorithm efficient in terms of computational resources. Second, as compared to an approach where every pair of edges is considered for grouping, such center finding technique makes the grouping more guided, thus more reliable and efficient.

### *3.2.2 Tackling outliers*

In cluttered background, some curved edges that are non-elliptic, may appear as if they are part of some elliptic edge. An example is shown in Fig. 25. The presence of such edges, referred to as the outliers, often results in false ellipse detections and degrades the performance of ellipse detection algorithms. Due to this reason, the incorrectly detected ellipses need to be filtered out at the end of the ellipse detection method. The saliency scheme presented in section 3.6 makes use of three parameters, out of which the circumference ratio and the angular continuity factor play an important role in reducing the saliency of outliers.

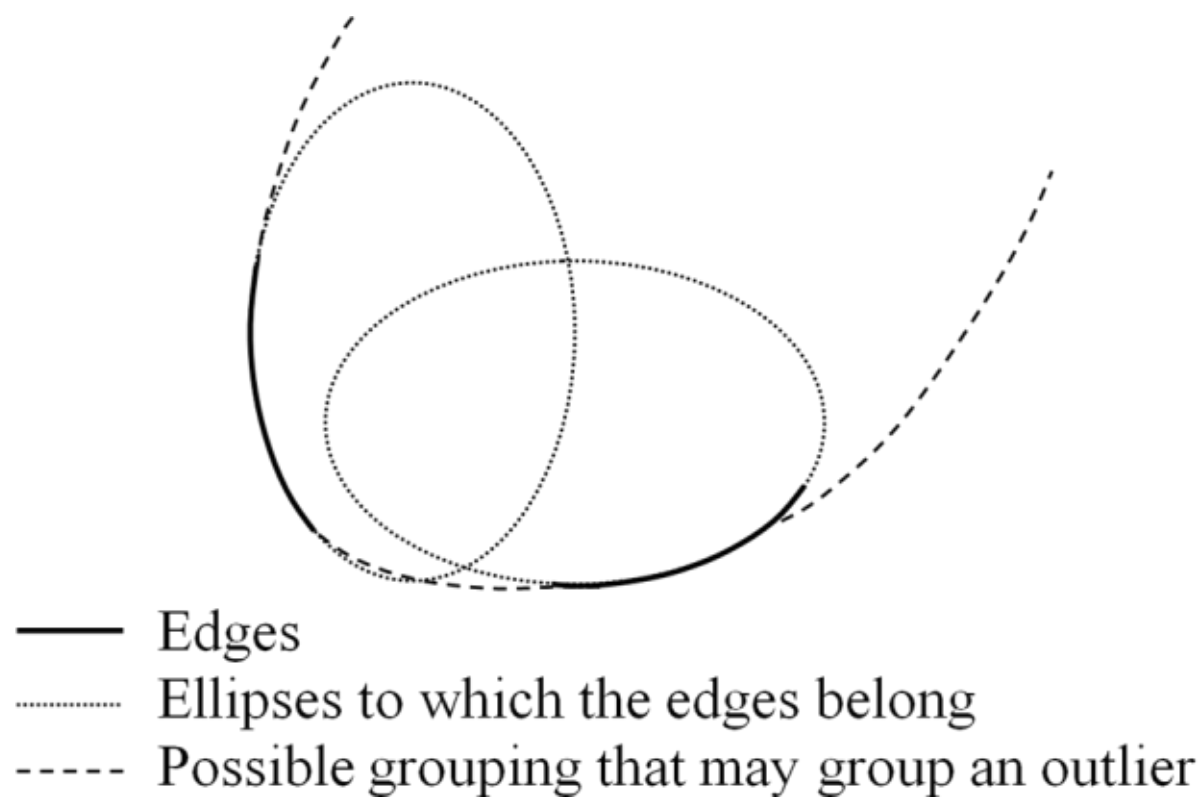

Fig. 25: Illustration of the presence of outliers. The solid lines are the actual ellipses and the dotted line is the non-elliptic edge created due to the cluttering of the two ellipses together.

### 3.2.3 *Tackling noisy edge*

The effect of noise, digitization, and light/shadow is often manifested in the form of corruption of the quality of edges. Two examples are presented in Fig. 26. Typically, researchers perform smoothing of the edge before further processing. As shown in Fig. 26(e), smoothing may result in loss of data and creation of outliers. The presented method does not use any form of smoothing. It instead fits a sequence of piece-wise linear segments on the edge [180]. This reduces the effect of local edge corruption due to noise while retaining the overall curvature information. The fitted sequence is not used to replace the edge. It is used only for performing curvature related analysis. Thus, there is effectively no loss of edge pixels.

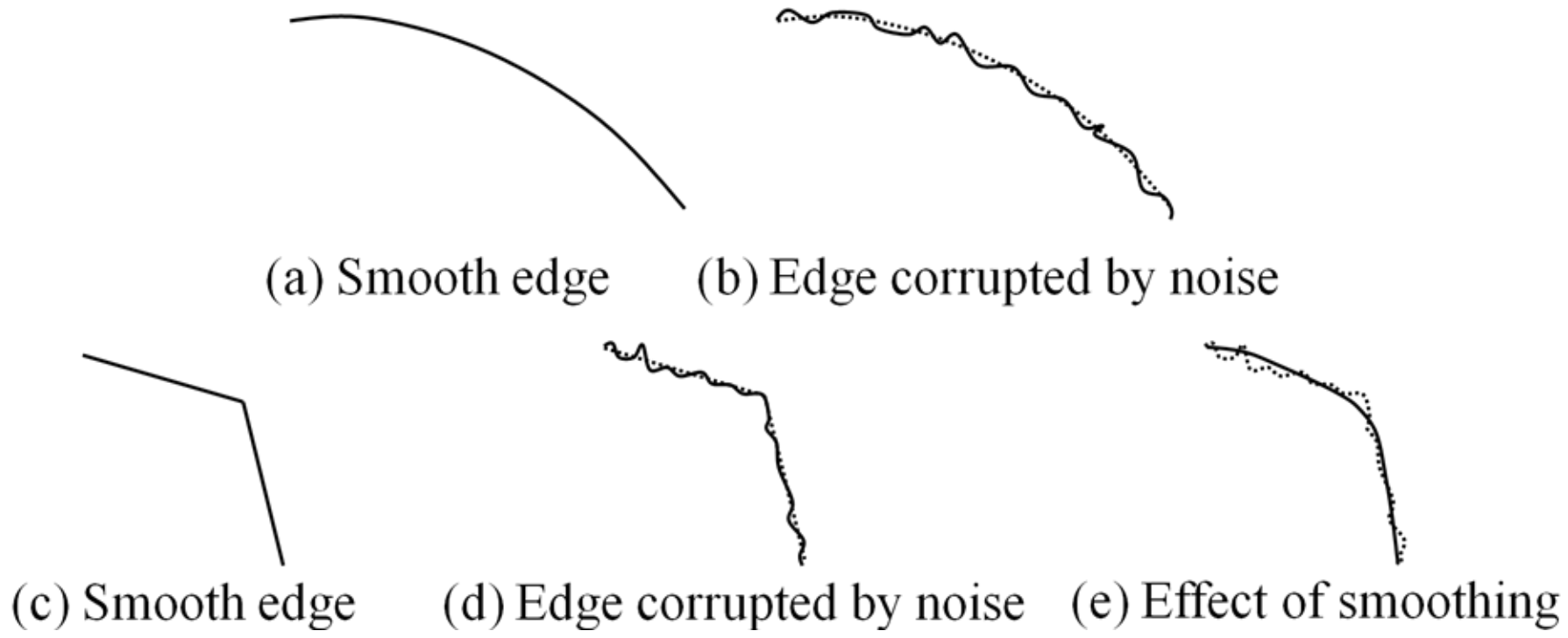

Fig. 26: Corruption of an edge due to noise. The dotted lines and solid lines in (b) and (d) represent two smooth edges and the noise-corrupted edges respectively. In (e), the dotted line shows a noise-corrupted edge, while the solid line shows the effect of smoothing.

### 3.2.4 *Tackling the reliability/precision uncertainty*

The concept of reliability/precision uncertainty was introduced in [203, 204] in the context of Hough transform (HT). It was shown that due to the quantization in parameter space, reliability and precision cannot be ideally increased together. Though the context in [203, 204] was Hough transform, it can be shown that the digitization of the image can cause similar problem. To explain it succinctly, given a fixed quantization or digitization, we need to consider local fitting (small number of closely located pixels/bins) in order to increase precision. However, for reliable fitting over the complete edge, we need to consider the larger distribution of the edge pixels, such that local disturbances get ironed out and a more global and suitable fit can be found.

In the context of elliptic shape detection, while detecting the ellipses, if we try to increase the precision and find the ellipses that match very well to an edge or group of edges, due to the digitization and incomplete data, the reliability of detection shall be poor. We shall encounter this duality in various stages of ellipse detection. For each such situation, we propose methods to reduce the impact of such duality. In general, we achieve good results by:

1. trading off between the precision or reliability at the time of detecting ellipses
2. Applying saliency criteria for checking precision and reliability of the detected elliptic hypotheses
3. Using saliency criteria to make final decisions about selection of good elliptic hypotheses.

## 3.3 Review of existing ellipse detection methods

Extraction of elliptic shapes from images has captured the interest of researchers for a long time. For two decades, numerous researchers are working on this problem. Many methods have been proposed for ellipse extraction in real images. The methods used for ellipse detection can be primarily categorized into four categories, viz., Hough transform (HT) based methods, edge-following methods, least squares based methods and genetic algorithms based methods. Besides the basic methods used, a lot of work has also been done in aiding these methods with other powerful tools like mathematical models of ellipses and geometric theorems. Further, some work has also been done in evaluating the elliptic hypotheses in order to reduce the false positive rate. In this chapter, we first present a review of the previous work done in these areas and then place our work in their context.

### *3.3.1 Hough transform based methods*

Hough Transform (HT) was first introduced in a patent filed in 1962. In 1972, Duda and Hart [205], adapted the basic notion of the Hough transform for the detection of linear and curved patterns in a picture. Their adaptation is called the Simplified HT (SHT). Since then it has been adapted, improved and applied for many applications in the field of image processing and computer vision. The literature documenting the evolution of HT and its various applications is quite vast. A good review and survey is presented in [206].

A key advantage of HT is that it does not require perfect connectivity of all edge pixels belonging to an ellipse and each edge pixel is used to vote independently on the five parameters of an ellipse. In real-life, where the pictures are noisy and the patterns are occluded, the segmentation techniques result in poor edge segmentation. HT, being a point based detection algorithm, is more robust and performs better than other edge based detection methods in such scenarios [207].

However, since the ellipse detection problem involves 5-dimensional parametric space in HT [206]. HT turns out to be a computation intensive method, requiring huge computation time and memory. In the last two decades, the researchers using HT to detect elliptical and circular segments have focused on providing computationally more efficient and faster HT adaptations.

One approach towards this problem is to reduce the dimensionality of the parametric space. This approach has been explored by [208-212]. While [210, 211] use the basic idea of the piece-wise linear approximation of the curved segments, [209] used the idea of converting any set of three points into 2-D Hough planes and then finding the overlapping planes. Similarly, [208, 213] use the information of the direction of the tangents drawn at the edge pixels to reduce the parametric space.

One interesting approach, similar to [207], is proposed in [214], where the Hough transform is performed in several stages, each stage using only 1-D parametric space. This approach is shown to reduce the memory requirements significantly and make the algorithm computationally efficient. The reference [215] follows a similar approach by finding the centers of the ellipses in the first stage and retrieving the other parameters in the second stage. However, the innovation in this paper is the use of a focusing approach rather than a typical polling approach, which makes the computationally inefficient job of finding the centers more efficient.

All the above mentioned research works use some or other form of the geometrical knowledge of the problem. They have only modified the problem set-up and approach, and not touched the HT algorithm in essence. However, a parallel stream of work has also been carried out in order to modify the HT algorithm itself and thus reduce the computational burden, retain the robustness, and in the meanwhile make HT more tractable to non-linear problem sets like the one being discussed. The most significant contribution in this regard is the development of the randomized HT (RHT) and probabilistic HT (PHT) over the years by various research groups [216-219]. Here, the approach used is to subsample the edge pixels of an image and use only the sampled pixels to vote on the parameters of valid ellipses. RHT differs from generalized HT in terms of the basic mapping performed by the HT. While each pixel under consideration is mapped to a curve in the parametric space in HT, a group of pixels is mapped to a single point in the parametric space in RHT [206]. However, RHT cannot detect all the obvious ellipses of an image and as the number of ellipses increases, its accuracy reduces.

Other interesting works related to Hough transform are discussed here. Li [220] and Aguado [213] used the gradient information at various edge pixels to aid the Hough transform. Li [220] proposed a segmented HT for circle detection, where 8-neighbour angle chain code and a modified direction measurement scheme is used to segment the whole edge into several segment with different geometric properties. This is followed by the application of HT on the identified edge segments rather than on the edge pixels. Though this method shows improvement over the standard HT, its application is limited to ellipses with low eccentricity.

Aguado [208] shows how positional constraints like distance and angular relationships between sets of edge points can be used to decompose the parameter space. The local properties of ellipse have been used to avoid the constraint of relative position between set of edge points. However, there are two important points that restrict the applicability of this method. First is the restrictive assumption that the pair of points chosen for HT should be more than 25 pixels apart and angle between them should be greater than 10 degree. The second set-back is that the gradient direction information is difficult to estimate in the presence of noise and quantization errors.

Cheng and Liu [218] proposed a method to select three points from the image in a deterministic manner, such that number of false selections of McLaughlin's method [216] was reduced. They used [207] for centre finding, then shifted the centre to the origin, and calculated the remaining parameters of ellipse.

Bennett [221] used projective geometry to reduce the computation time and increase the efficiency of HT based ellipse detection method. It used prior information like maximum area, maximum eccentricity or minimum axis length to extract feature or geometric parameters from images. Due to these assumptions of the availability of prior information, the applicability of this method is limited.

Lu [222] proposed a method for detection of incomplete ellipses under strong noise conditions by applying RHT to a region of interest in the image space. It used iterative parameter adjustment and reciprocating use of image space and parameter space. This method improves the performance of RHT in terms of robustness and efficiency while retaining the advantages of RHT. This method still suffers with the drawback of traditional RHT like performance degrades with the increase of number of ellipses and presence of overlapping and occluded ellipses.

Here we present a general note of HT-based methods. In a typical HT method, after finding parameters of most voted ellipse, pixels in the image which are in the vicinity of the contour of this ellipse are found and removed. This procedure is repeated until the method cannot find any suitable ellipses in the image. In such method, if the pixels belonging to a shape are few in number, the chances of detecting the shape accurately decrease. Further, the detection of pixels in the vicinity of the ellipse, and repeating the whole procedure many times while obtaining only one ellipse in each iteration is computation intensive. However, if edges (pixels) belonging to an object can be identified and grouped, the accuracy of parameters finding of ellipses using HT increases. This idea shall be used later in the proposed work.

### *3.3.2 Edge following methods*

Mai [194] proposed a modified RANSAC (RANdom SAmple Consensus) [223] based ellipse detection method, which first extracts the fitted line segments from the edge data of the image. It follows an edge in terms of its continuity to group the edge with other edges. Finally, RANSAC based ellipse fitting is performed on these grouped arc segments. This method shows good performance in terms of accuracy and computational efficiency over many existing methods. However, the performance of this method is highly dependent upon the choice of the two thresholds – proximity distance and angular curvature. Split and merge detector proposed by Chia [195] in essence the same. The performance of both these methods deteriorates in the presence of occluded ellipses.

### *3.3.3 Least squares based methods*

Least squares based methods usually cast the ellipse fitting problem into a constrained matrix equation in which the solution should give least squares error. From the mathematical perspective, important work for ellipse detection has been done by [188, 189, 193, 224]. In terms of application, some interesting works include [190, 191, 225].

Rosin [193, 224] proposed direct methods to detect lines, elliptical arcs and ellipses by segmenting digital arcs into combinations of straight lines and elliptical arc segments. With a given a list of connected edge pixels, the technique first recursively produces line approximation of the digital arc based on significance rating measure. Then in a similar manner, line segments are segmented into elliptical arc segments by fitting an ellipse to the end points of the line segments. A set of line segments is replaced with an elliptical arc segment if the replacement yields an improved significance rating. To overcome the faulty segmentation with the recursive splitting procedure, an additional stage was proposed to combine adjacent arc segments to yield a better arc approximation. This leads to an appreciable reduction in the number of line and elliptical segments. This method merges only the adjacent arc segments, but ellipse detected by them tends to be made of shorter elliptical arcs.

Cabrera [188] proposed an unbiased estimation of ellipses by using least squares method with median bias and bias reduction using bootstrap technique. Fitzgibbon [189] proposed a very robust and efficient least squares method for ellipse detection. This method is invariant to affine transformation and is computationally efficient.

Ellis [225] proposed a model based ellipse detection method. First, using the least mean squares (LMS) error fitting method, ellipses are fitted to a general conic function. LMS used covariance matrix to achieve this and keep the uncertainty intact. The initial fits are improved by extending the elliptic arcs. For this, it used the Mahalanobis distance measure to select line segment candidates with suitable orientations and location with respect to the current ellipse. The detected ellipses and their uncertainties are used to find object models, estimate view point, and infer structures in the scene using Kalman filters.

Kim [190] proposed a least-squares based ellipse detection algorithm which first fits short line segments to extract the elliptic arcs, and then determines the parameters of ellipse using least squares fit.

Meer [191] reviewed regression analysis methods used for fitting a model to the noisy data. In this paper, the comparison between the least-median-squares (LMedS) method with the RANSAC algorithm in the presence of noise has been discussed. It has been shown that, using the probabilistic speed-up techniques, the computation of LMedS estimates is feasible, although more demanding than that of M-estimates. On the other hand, RANSAC requires reliable initial estimates which can be obtained from LMedS algorithm.

Due to the matrix formulation, least squares methods are very fast and analytic solution approaches can be found for such problems. However, least squares based methods are very poor in handling outliers. Further, typically the value of least squares error (or residue) cannot be trusted to give a desirable fit. This is because generally the number of edge pixels are much larger than the number of unknowns (the five parameters of the ellipse), thus the problem is over-determined. The presence of constraints (for choosing elliptic solutions) makes the problem even more over-determined. In short, the solutions obtained by least squares method may not be reliable in all the scenarios. Their solutions should be reaffirmed using other techniques for increasing the reliability of the solutions. Based on this, we use least squares method as one of the judging criteria, while we use other geometrical methods to find the possibilities of existence of ellipses. Gestalt philosophy based ellipse detection method is also interesting due to the use of novel metric based on image composition [226].

### 3.3.4 Genetic algorithms

Kasemir [200] proposed an ellipse detection algorithm based on Hough transform using differential evolution to optimize the parameters of ellipses. The major drawback of this method is its applicability to very limited eccentricity ellipses, usually circles.

Kawaguchi [198, 199] proposed an algorithm to detect ellipses based on genetic algorithm. The algorithm groups adjacent edge pixels with similar gradient orientations into regions called as line-support regions. Next, candidates for elliptic arcs are selected from these line-support regions. In this algorithm each ellipse is defined by a triplet of line-support regions and using the genetic algorithm it searches for triplets of line-support regions that have the highest fitness to the image.

Procter [201] showed the comparison of RHT and GA based ellipse detection method and observed that with less noisy data, RHT performs better than GA while performance of GA increases with the increase of noise in the image.

Genetic algorithms are good at dealing with non-linear optimization problems in which there are many local minima. However, these algorithms are generally computation intensive and require a lot of time for convergence. The stochastic nature of such algorithms make them time consuming. Further, the possibility of premature saturation cannot be fully excluded in most cases and the algorithms have to be carefully designed for each problem.

### 3.3.5 Mathematical models

Hinton [227] proposed an integral transform based mathematical model for circle, ellipses and parabola parameterization. Its circular disc detection algorithm used two transforms to find the three parameters needed to define a circle. The first two-dimensional integral transforms the edge pixels into a parameter space locating disc centers. The second one-dimensional integral determines the disc radius for each of the previously determined disc centers. This algorithm has been extensively tested with extremely good results. Then, it is demonstrated that the circular disc algorithms can act as a precursor giving very accurate values for ellipse location and orientation, along with estimates of the semi-major and minor axis.

Though mathematical models provide good analytical framework and (in most cases) computation efficient framework, they are often restrictive in terms of application to real images. Often, they are good in handling outliers. However, they are generally very sensitive to digitization errors and noise.

### *3.3.6 Geometric center finding*

In 1989, Yuen [207] provided a very important contribution to the elliptic shape recognition problem. His work suggested two improvements over the application of conventional Hough transform for ellipse fitting problem. It is well known that the parametric space for the ellipse fitting problem is five-dimensional. Accordingly, at least five points are required to generate an elliptic hypothesis. Yuen’s method suggested that the five-dimensional parametric space be split into two subspaces, one space containing the centers of the ellipses, and the other space containing the remaining parameters. The first subspace is two-dimensional while the second subspace is three-dimensional. Since there are various geometric theorems that may be used directly to find the centers of the ellipses using less than five points, the parametric treatment typical of Hough transform needs to be done only for the second subspace, which requires three points. Further, instead of the usual parametric treatment, we can use other methods (like least squares, etc) to generate the parameters in the second subspace, if needed. All the remaining features, like binning of the parametric space, histogram count, voting in the five dimensional parametric space, etc., were kept in the same form as conventional Hough transform. Thus, the two main contributions were that less than five (three) points were sufficient to generate an elliptic hypothesis and that by splitting the parametric space into two subspaces, individually efficient methods can be used in each parametric subspace.

Following Yuen [207], geometry based center finding was used by many researchers. The geometric method proposed by Yuen used the tangent and the chords for the chosen three points to generate the information of center of the elliptic hypothesis. Elmowafy [228], Mc Laughlin [216, 217] used the same geometric method as Yuen [207]. Others used chord and chord bisectors as a method to generate the centers of the elliptic hypotheses [229, 230].

Ho [231] used a global geometric symmetry to locate all the possible symmetric centers of ellipses and circles in an image. Using these, the feature points are classified into several subimages. On each subimage, geometric symmetry is applied to find all possible sets of three parameters (major axis, minor axis, and orientation) for ellipses and radius for the circles. Finally, it used the accumulative concept of HT to extract all the ellipses and circles of the input image.

While Guil [215] used just two points for estimating the centers, Zhang [232] proposed an improvement over it using the concepts of convexity and associated convexity. A parallelizable version of 1-D HT has been proposed in [233]. It is worth noticing that instead of the centers of the ellipses, here, the foci of the ellipses are the pivot of dimensionality and computational reduction.

### 3.3.7 Grouping strategies

As discussed earlier, an edge usually represents a shape only partially. Typically a shape may appear in an edge map in the form of one or more broken edges, which may sometimes be so far apart that it is not easy to say if they belong to the same elliptic object. Due to this fact, the contemporary edge based grouping methods [194, 195] that use either a single edge or edges that are close to each other for predicting the ellipses may often fail. In order to solve this problem, it is essential to group edges that may be far apart and check if they might belong to a common ellipse. A straight forward approach would be to consider every pair of edges for grouping [195]. But, it is obvious that such approach shall greatly increase the computational burden and time. It shall be rather useful to group the edges more deterministically, taking cue from some information.

Mai [194] proposed a RANSAC based ellipse detection method, which first extracts the fitted line segments from the edge data of the image. Then the potential elliptic candidates belonging to same ellipse are identified and grouped. It used the proximity and angular curvature criteria to determine the merging decision between two arc segments. The performance of this method is highly dependent upon the choice of the two thresholds – proximity distance and angular curvature. Chia [195] also used a similar concept for grouping the edges.

Hahn [234] proposed an ellipse detection method based on grouping points on elliptic contour. Whether some curved segments belong to the same ellipse or not, are tested by comparing the parameters of candidate ellipses that are made by the curve segments. This method can reduce the total execution time because it estimates the ellipse parameters in the curve segment level not in the individual edge pixel level. However, the performance of this method deteriorates for complex real images.

Kawaguchi [198, 199] groups adjacent edge pixels with similar gradient orientations into the regions called as line-support regions, which are subsequently used for ellipse detection

Ji [235] proposed a grouping scheme to pair the arc segments belonging to the same ellipse as an improvement over [224]. While [224] groups the edges based on a scale invariant statistical geometric criterion which can be verified either in parametric space or in residual error space, Ji [235] takes proximity and direction of arc segments (clockwise and counter clockwise) into account.

Kim [190] proposed a grouping scheme based on three curvature and proximity based conditions as follows: an arc should be a neighbor in eight group classification [190], the arcs follow a convexity relationship as proposed in [215], and the inner angle between two edge pixel on an ellipse should not exceed 90 degree [190]. If the arcs satisfy these three constraints they represent circular arcs. In order to determine ellipses from the list of circular arc (obtained by merging three circular arc segments), it first finds the centre of ellipse by centre finding method [207]. Two arcs belong to one ellipse if they satisfy the parameters obtained by least squares method and the three constraints. These arcs are then merged and remaining elliptical parameters are extracted.

### *3.3.8 Hypotheses evaluation and reduction of false positives*

One of the problems faced by any ellipse detection algorithm is that in attempt to detect all the ellipses actually present in real images in the absence of prior knowledge of the number of ellipses, they have to compromise on the accuracy of the algorithm. Any ellipse detection method, including the proposed method, suffers from the problem of reliability/precision uncertainty. Due to this, the ellipse detection methods have to compromise on either the reliability or the precision. The popular choice is to compromise on the precision, as the quantization already limits the precision and due to the absence of a priori information on the elliptic objects present in the image, it is often important to be reliable in detecting the elliptic shapes. Due to this, often, the ellipse detection algorithms generate numerous elliptic hypotheses, not all of which correspond to actually present elliptic objects. Sometimes, many elliptic hypotheses are generated for a single elliptic object, and at other times, the hypotheses do not correspond to any actually present elliptic object in the image.

Thus, it becomes important to evaluate the possibility of an elliptic hypothesis actually corresponding to an elliptic object. In other words, a method is needed to identify the elliptic hypotheses which are more likely to correspond to an object in image. Though there are no direct methods to detect the false positive ellipses, some kind of filtering has to be performed on the elliptic hypotheses in order to determine the hypotheses that are more likely to belong to an actual elliptic object in the image. Researchers use the ‘saliency’ or ‘distinctiveness’ scores for quantifying the reliability of a hypothesis [196, 197, 206, 236-240] and select the more salient elliptic hypotheses.

‘Distinctiveness’ term is more often used to identify the elliptic hypotheses that may be similar to other hypotheses. We bring special attention to [239] that have recently proposed measures to cluster similar ellipses. Though [239] does present some important work in this regard, the measures suggested are neither simple nor computationally efficient. Further, in our opinion, the applicability of this method is restricted to Hough transform related ellipse detection methods and edge following methods. Basca [237] proposed a similarity measure that compares two elliptic hypotheses using the Euclidean distance between them in the Hough parametric space. Due to the absence of normalization, this method is sensitive to scale changes.

Various researchers have been working on this problem in order to devise indirect methods of evaluating the saliency of the elliptic hypotheses and this is still an open problem.

Qiao [197] proposed ellipse detection method based on saliency of an arc. This method explores the relationship between spatial connectivity and the incremental point angle of elliptic inliers this relationship can be used to detect elliptic arc end points. Then angle subtended by the elliptic arc can be used as ellipse validation criteria. This method depends on too many thresholds which are application dependent. This limits its use for wide range of application. The performance of this method depends on the proper choice of these threshold parameters.

Another popular measure is the percentage of circumference covered by the edges that generated an elliptic hypothesis. Typically, the number of pixels of the edges that were used to generate an elliptic hypothesis is divided by the perimeter of the ellipse. This has been called as the pixel count feature in [236]. Elmowafy [228] also verified the elliptic hypotheses by checking the ratio of count of pixel on elliptic curve and approximated circumference of the ellipse. There are two major concerns regarding such scheme [196]. First, the pixels are an approximate (quantized) representation of the elliptic hypothesis. Due to this, in case of a complete elliptic edge, the number of pixels is more than the actual perimeter of the ellipse. Even if there is an incomplete edge, the edge represents a fraction of perimeter of the ellipse using more pixels than the actual length of the fraction. Second, the determination of perimeter of an ellipse is a classical problem in mathematics for which no closed form analytic solutions are available. The perimeter used is typically one among the various numerical approximations provided by scientists [241]. It is notable that all these approximations are subject to some assumptions that may not be generally valid for all the elliptic hypotheses generated.

Another saliency criterion considers the distribution of the pixels around the elliptic hypothesis (alignment of the edge pixels along the elliptic hypothesis). This idea was first proposed by [238], though in the context of straight lines primarily. In the context of elliptic hypotheses, we present the following method in section 3.6.2.2.

## 3.4 Proposed approach for ellipse detection

Typically, the methods previously suggested by researchers begin with the extraction of the edge map. The edge map serves as a starting point for detecting the ellipses. As discussed earlier, the edge map obtained by a typical edge detection algorithm is too crude to be used directly. The edge map has to be made suitable for ellipse detection using some preliminary processing. This involves extraction of edges that are continuous, with smooth curvature, and contain no inflexion point. The work done in chapter 2 for edge and line cues sufficiently meets these requirements and is used as the input to the ellipse detection method.

The first stage of ellipse detection method is to generate the elliptic hypotheses. In the current work, we first use two conditions to find the edges that may be grouped together. First condition is the edges that can be grouped with an edge should lie inside the convex envelope of the edge. This is elaborated in section 3.5.1. Second condition is that the associated convexities of the edges should be suitable for them to be grouped. This is presented in section 3.5.2. After satisfying these conditions, we use the centers of possible ellipses as the guide for grouping edges. Finding the centers using three points is discussed in 3.5.3. After finding the centers, a relationship score is used to quantify the relationship between the edges and the centers computed by them. This is discussed in section 3.5.4 Then, a simple grouping scheme based on common centers is proposed in section 3.5.5. The algorithm for stage 1 and its flowchart are presented in section 3.5.6

The second stage of the ellipse detection method is to select the reliable elliptic hypotheses, and thus to reduce the false positive elliptic hypotheses. There are numerous parameters, direct or derived, that may be used for reducing the false positive elliptic hypotheses, and there are various ways of combining the various selected parameters. Though all these parameters and ways to combine them are good to some extent, and perform well in certain scenarios, a saliency scheme that is more widely applicable is very difficult to derive. The proposed saliency scheme performs well for a larger range of datasets [242, 243] is presented in section 3.6. The overall flowchart of the ellipse detection method is shown in Fig. 27

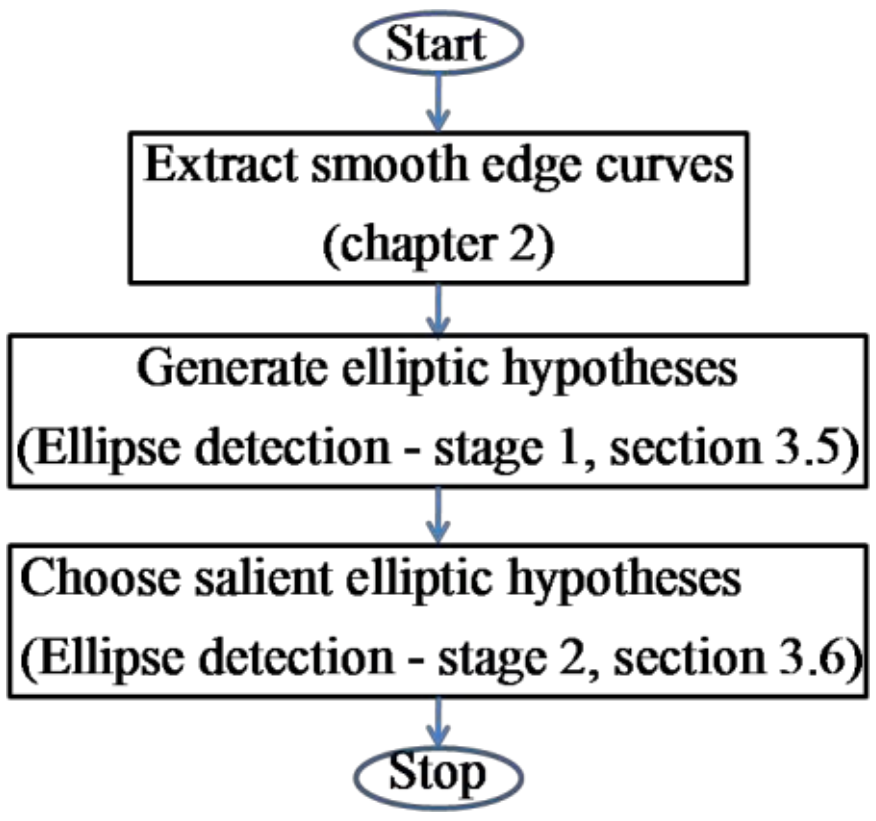

Fig. 27: Flowchart of the ellipse detection method

### *3.4.1 Comparison of the proposed ellipse detection method with the existing methods*

In the context of the literature review for ellipse detection methods presented above, we present below the important features of the proposed method that distinguishes it from the existing methods. The presented method is radically different from the previously reported method in many aspects, which are highlighted below (details can be found in sections 3.5 and 3.6):

1. The center finding method proposed in this method may be interpreted as a variation of the methods proposed in [207, 216]. Though the center finding method is indeed inspired by [207, 216], it is different in its intention and various technical aspects. First, the sets of (three) pixels for finding an ellipse are chosen from an edge and not from the entire image. As opposed to randomized Hough transform [207, 216], in our case the center finding is being performed on one edge at a time and not on the entire image. Due to this, the center finding is more robust. Second, each edge can be processed independently in this step. The computationally intensive step of finding the pixels around the ellipse and removing them from further consideration [207, 216] is not required for our method. Third, the histogram is generated for the centers only (as opposed to the regular Hough transform, which is typically five-dimensional), and no definitive decision (and identification of other parameters) is made at this step. This method serves as a starting step and guide for grouping the edges and further processing, and not as a standalone method for ellipse detection. Effectively, the proposed center finding method makes the overall method more time efficient and robust as compared to various Hough transform based methods.

2. The proposed method generates the histogram for the centers only, which are equivalent to two-dimensions in the context of Hough transform. However, the binning method proposed by us converts the two-dimensional parameter space into a one-dimensional parameter space using a one-to-one transformation and inexpensive computation. Further, dynamic linked list scheme [207, 216], which saves the information of a bin only if required, makes the overall scheme memory efficient.

3. Due to the use of center finding method and the advanced ellipse detection technique, the proposed method is different from conventional edge following methods [194-197] and does not suffer from various problems of edge following methods. First, since the grouping criterion is not restricted by proximity or continuity [194, 195], even the farthest or apparently unrelated edges are considered for grouping. Thus it is helpful in solving the problem of discontinuous edges of an ellipse. Second, even though every plausible grouping condition is taken into account, the computational complexity of the method is very low. The maximum time required (without parallelization) for the complete ellipse detection method is typically few tens of seconds for the Caltech 256 dataset [242], and crosses a minute for very few images. Third, large portions of the proposed method are independent for various edges, thus facilitating parallelization. Though, the presented results do not use parallelization, the possibility is indicated where applicable. Using advanced parallelization technology (on GPU) is expected to reduce the computation time below 1 second.

4. While the proposed center finding method provides an initial grouping cue, the grouping scheme is further improved using the proposed advanced ellipse detection method. The advantage is derived by the use of Boolean flags that are assigned to all the possible pair of edges indicating the plausibility of grouping, which is determined by the convex characteristics of the edges. This greatly enhances the efficiency of grouping and reduces the computational and memory burden.

5. While detecting elliptic shapes in real images, often multiple similar hypotheses are generated for a single elliptic object. This is especially true for Hough transform based methods. Thus, it is important to cluster the similar elliptic hypotheses and choose one representative elliptic hypothesis for a cluster. Though various distinctiveness measures have been proposed in this regard, these measures rather judge the elliptic hypotheses for reliability or precision (together called saliency) and do not present straight forward, computationally efficient methods of clustering. In this report, we propose a very simple, intuitive, and computationally inexpensive method for clustering similar ellipses. In this manner, after clustering, the comparatively more intensive process of generating the saliency or distinctiveness criteria needs to be performed on lesser number of elliptic hypotheses.

6. Another major contribution of the proposed method is the saliency criterion. As with every ellipse detection method for real images, including the current method, it is not guaranteed that every ellipse that gets detected indeed belongs to an actually present ellipse and that every ellipse actually present in an image gets detected. Further, the correctly detected ellipses may not closely follow the actual ellipses and it is difficult to discretely determine their correspondence. Every ellipse detection method is plagued by these problems. In this context, though the proposed method is highly likely to detect all the actually present ellipses, it also detects numerous false ellipses that do not correspond to any actually present ellipse. Though researchers have tried to use some saliency criteria or combination of saliency criteria, most of them are effective either for the selected images/dataset, level of noise, level of clutter, etc. None of them provides a general saliency scheme valid for a large range of images. The saliency scheme proposed in this report uses three mathematically determinable saliency parameters, which together indicate if a detected ellipse should be considered as a good candidate for actual ellipse. Besides defining these useful parameters, it is also important to combine them in such a manner that they are cumulatively effective in recognizing better ellipses. In this regard, various combination schemes are considered and their impact and effectiveness is studied. The scheme that is most effective is then presented. This scheme provides a fine balance of all the three parameters such that neither a single parameter has over bearing influence on the decision making, nor the significance of any parameter is negligible. Further, threshold for saliency is determined statistically from the image itself and does not need to be chosen empirically (like other methods). Due to this, the saliency scheme is more generic, widely applicable and performs well for a large set of images [242, 243].

## 3.5 Ellipse detection: stage 1 (generating elliptic hypotheses)

### *3.5.1 Finding the edges in the search region of an edge*

The convex envelope of an edge (as shown in Fig. 28) can be used to filter out the candidate edges that may be grouped with the edge. If we are interested in the elliptic contours, it is obvious that for a given edge, other edges belonging to the same ellipse as the given edge will have to lie in its convex region. Here, we assume that the edge is of smooth curvature and does not have any inflexion points, as ensured by our edge processing method discussed in chapter 2. This can be used to identify a search region in which we search for the edges that can be possibly grouped.

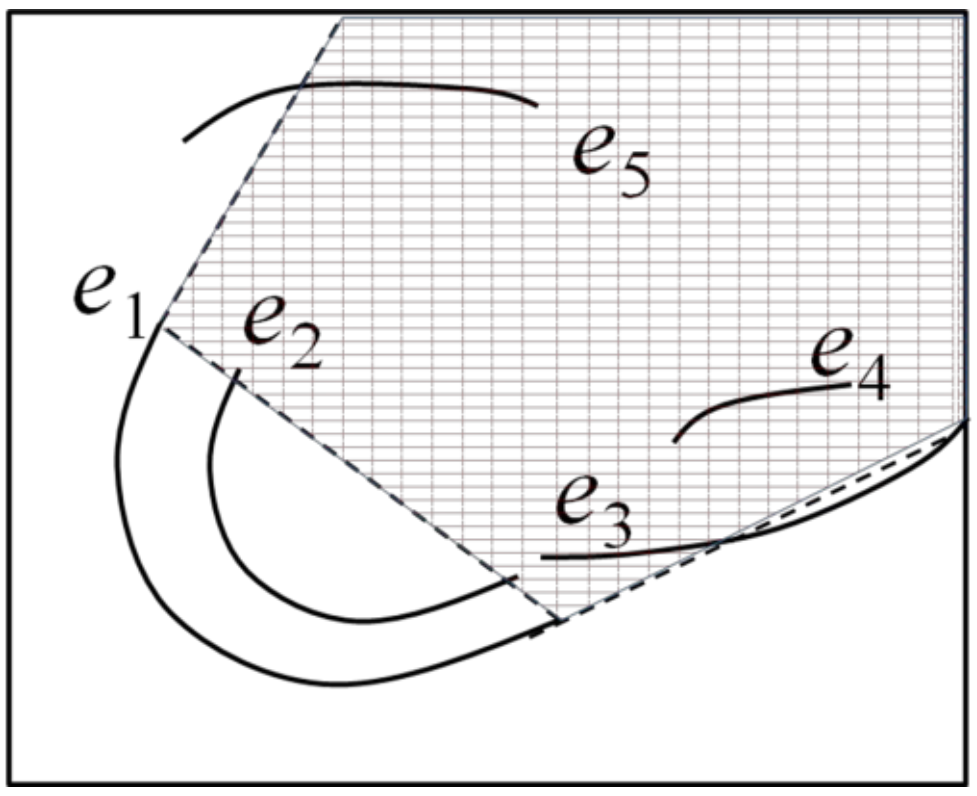

Fig. 28: Illustration of the convex envelope (search region) of the edge $e_1$.

#### 3.5.1.1 Finding the search region

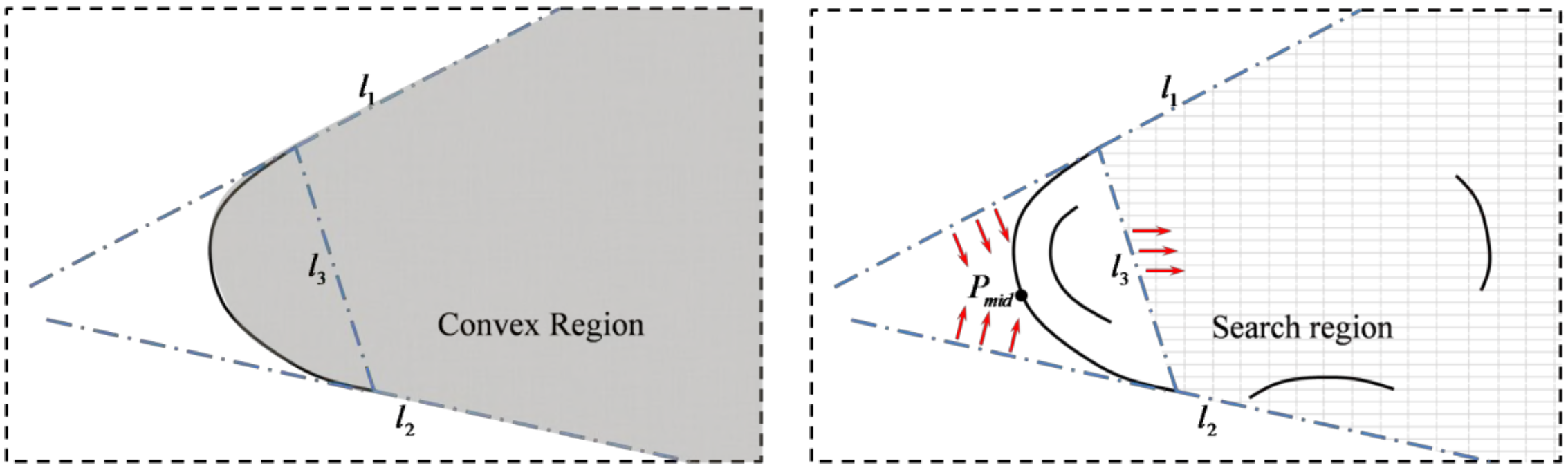

(a) Illustration of convex region (b) Illustration of search region

Fig. 29: Illustration of the convex region and the search region

With the term convex region of an edge, we mean the following. For a given edge, let the tangents to the edge at its end points be denoted by $l_1$ and $l_2$, and the line segment connecting its end points be denoted by $l_3$. The two tangents and the edge divide the space into two regions. The region which contains line segment $l_3$ is the convex region of the edge. See Fig. 29(a) for illustration.

The convex region itself can be used directly as a search region. However, there might be some edges within the closed region formed by the edge and $l_3$ (for example, see the edge between $P_{mid}$ and $l_3$ in Fig. 29(b)). These edges cannot be part of the ellipse formed by the edge under consideration. Thus, we may consider the shaded region in Fig. 29(b), which is formed by $l_1$, $l_2$, and $l_3$, and is a subset of the convex region, as the desired search region.

#### 3.5.1.2 Finding the edges within the search region of an edge

For a given edge $e'$, after finding its search region (characterized by $P_{mid}$, $l_1$, $l_2$, and $l_3$), the edges within the search region can be found as follows. Let $P_{mid}$ be the middle pixel of the considered edge (see Fig. 29(b)). $P_{mid}$ can be considered as a reference point to determine the edges within the search region. Thus, an edge $e_i$ lies in $S$ if all the three criteria below are satisfied:

- $e_i$ and $P_{mid}$ are on the same side of $l_1$,
- $e_i$ and $P_{mid}$ are on the same side of $l_2$, and

$e_i$ and $P_{mid}$ are on the opposite sides of $l_3$.

### 3.5.2 Finding the associated convexity of a pair of edges

The associated convexity of a pair of edges can be studied in order to further exclude the grouping of the edges that are unsuitable for grouping. Fig. 30 shows five scenarios of the associated convexity between two edges. It is evident that the scenario presented in Fig. 30(e) is the only scenario that should be considered for optimal grouping. We present a simple method below that can identify if the two edges have their associated convexity as shown in Fig. 30(e).

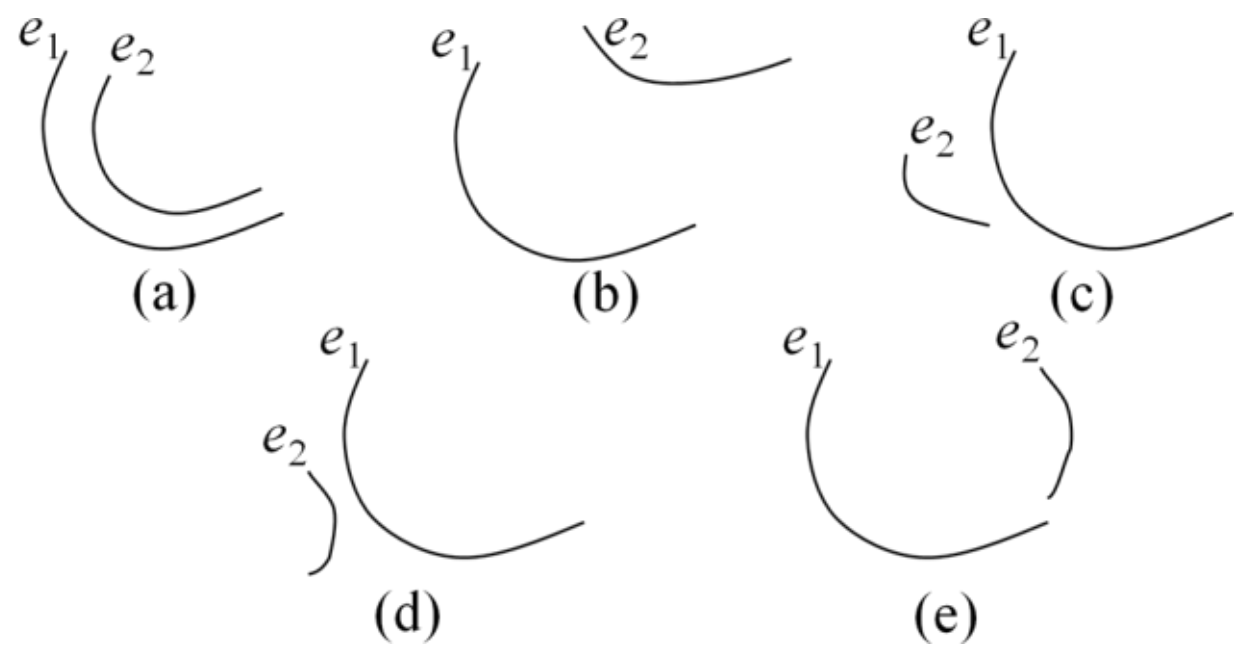

Fig. 30: Possible associated convexities between two edges. Only the pair in (e) should be a candidate for grouping.

Let us consider the line segments $l_1$ and $l_2$ formed by joining the end points of $e_1$ and $e_2$, respectively. Let $P_1$ and $P_2$ be the midpoints of the line segments $l_1$ and $l_2$. Let $l_3$ be a line passing through $P_1$ and $P_2$, such that it intersects the edges $e_1$ and $e_2$ at $P_1'$ and $P_2'$ respectively. This is illustrated in Fig. 31. The pair of edges $e_1$ and $e_2$ are suitable for grouping if and only if:

$$P_1'P_2' \approx P_1P_1' + P_1P_2 + P_2P_2' . \quad (8)$$

The approximation is attributed to the fact that $P_1'$ and $P_2'$ have to be the edge pixels nearest to the line $l_3$, and may not be exactly on $l_3$ due to digitization of the edge pixels.

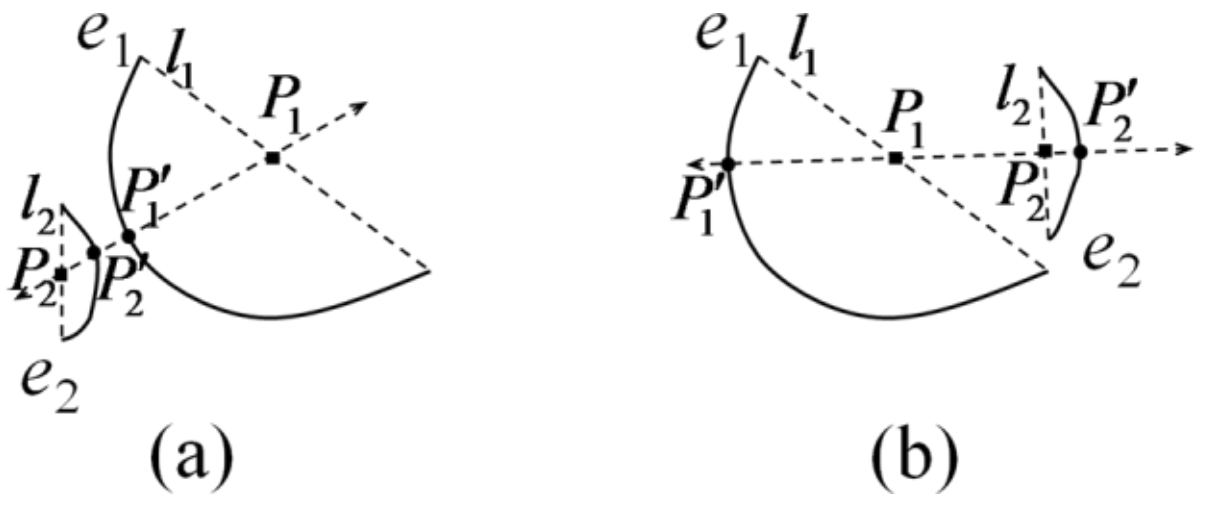

Fig. 31: Illustration of the concept of associated convexity

### 3.5.3 *Finding center of the ellipse*

The geometrical concept used for retrieving the centers [207] is presented here. The proof of this geometrical concept is provided in [209]. Let us consider a set of three distinct pixels, $P_1(x_1, y_1)$, $P_2(x_2, y_2)$, and $P_3(x_3, y_3)$, on an edge and represent the lines tangential to the edge at these three points as $t_1$, $t_2$, and $t_3$ respectively. We denote the intersection point of lines $t_1$ and $t_2$ as $P_{\tan,12}$ and that of $t_2$ and $t_3$ as $P_{\tan,23}$. Further, we denote the midpoint of the line segment joining $P_1$ and $P_2$ as $P_{\mathrm{mid},12}$ and the midpoint of the line segment joining $P_2$ and $P_3$ as $P_{\mathrm{mid},23}$. Now, we construct a line $l_{12}$ that passes through $P_{\mathrm{mid},12}$ and $P_{\tan,12}$, and a line $l_{23}$ that passes through $P_{\mathrm{mid},23}$ and $P_{\tan,23}$. Then the centre of the ellipse is given by the intersection point of the lines $l_{12}$ and $l_{23}$. The concept is illustrated in Fig. 32.

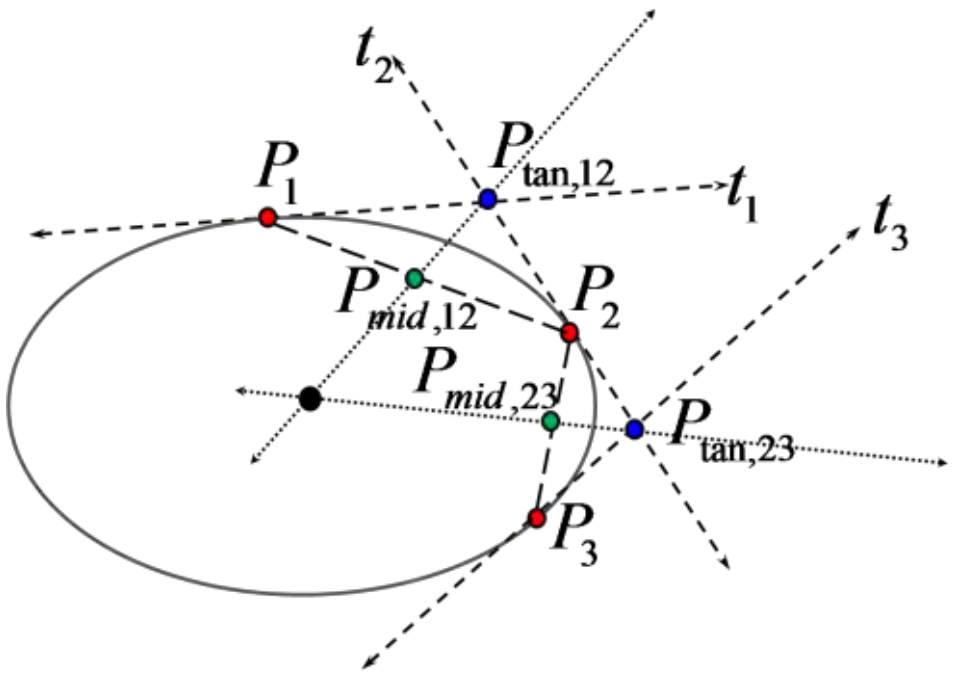

Fig. 32: Illustration of the geometric concept of finding the centre of an ellipse from three points $P_1$, $P_2$, and $P_3$.

There are two geometric exceptions to the above mentioned method. First case is that $t_1$ and $t_2$, or $t_2$ and $t_3$ are parallel to each other. Second case is that $l_{12}$ and $l_{23}$ are parallel to each other (i.e., the pixels in the set are collinear, though they may belong to a curved edge). The second case is unlikely to appear if the edge contours are of smooth curvature. In both these cases, the center of the ellipses cannot be found. Thus, in order to generate reliable estimate of the ellipse's center, many sets of points have to be generated for each edge contour. The selection of the sets is discussed in Appendix A. Here it suffices to state that a set of the points are generated by splitting the edge contour into three sub edges and selecting one point from each sub-edge.

One of the important steps in the above method is the calculation of the tangents at the chosen points. Due to the digitization of image, the tangents cannot be calculated directly. Further, changing the derivative $dy/dx$ into differences may result into inaccurate estimation of the tangent. Error bounded tangent estimator will give good performance [244, 245]. A reliable method of calculation of tangent is discussed in Appendix B.

In order to apply HT, the parametric space of the centers is quantized into bins. The two-dimensional parametric space can be actually considered as one-dimensional space. The binning scheme and calculation of bin numbers for computed centers are discussed in Appendix C.

### *3.5.4 Relationship score for a bin-edge pair*

Given an edge curve $e$, $S$ sets of three pixels are generated. For each set, a centre can be computed using the geometric concept in 3.5.3. As discussed in Appendix A, all the sets may not generate valid centers due to various reasons. Let the number of sets that generated a valid center be $S_e$. Ideally, all $S_e$ centers should fall in the same bin (which should coincide with the bin containing the center of the actual ellipse). In practice, all the computed centers will not fall in the same bin.

We propose to assign a score to each bin-edge pair, which is an indicator of the trust that can be put upon their relationship. A simple relationship can be $r_e^b = S_e^b$, where $S_e^b$ is the fraction of $S_e$ that voted for the bin $b$. We enhance $r_e^b$ as follows:

$$r_e^b = S_e^b \, r_1 \, r_2 , \tag{9}$$

where, $r_1$ is a function of $S_e^b / S_e$ and $r_2$ is a function of $S_e / S$.

The ratio $S_e^b / S_e \in [0,1]$ is an indicator of the relative weight of the bin $b$ as compared to other bins that were computed for the same edge. If $S_e^b / S_e$ is high, the bin is better ranked than the rest of the bins, indicating that the relation between the edge and the bin $b$ is stronger, and thus should be given more priority. On the other hand if $S_e^b / S_e$ is less, the bin might have been computed by a chance combination of the randomly selected pixels, and should not be given significant importance. A non-linear $r_1$ as shown in Fig. 33 is preferred to dampen $r_e^b$ for low $S_e^b / S_e$. We compute $r_1$ as follows:

$$r_1 = \left( \frac{S_e^b}{S_e} \right) \exp\left( \frac{S_e^b}{S_e} - 1 \right), \tag{10}$$

Though other functions might be chosen to achieve similar effect, this is not the scope of the current work to compare with other types of functions. Here, it suffices to say that the above function emulates well the desired effect.

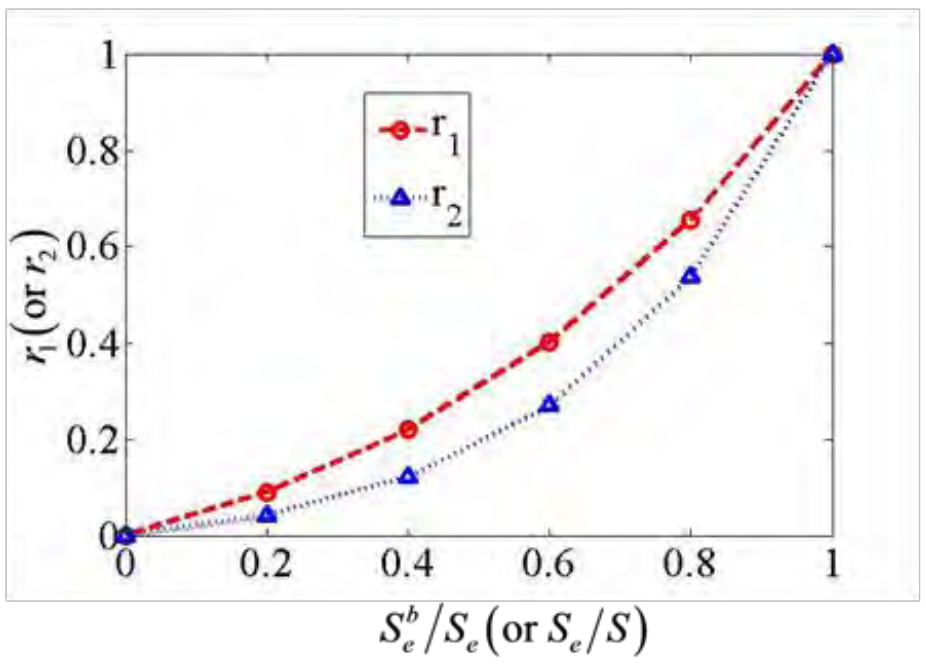

Fig. 33: Illustration of variation or $r_1(\text{or } r_2)$ with $S_e^b/S_e$ (or $S_e/S$ ).

As discussed before, out of the total $S$ sets generated for an edge, all may not result into valid bins. If there are only a few valid sets $S_e$ in comparison to $S$ , it may mean that the edge is a poor elliptic candidate (and thus an outlier). On the other hand, if the ratio $S_e/S \in [0,1]$ is high, then it is indicative of the edge being a good elliptic arc. Similar to $r_1$ , we set $r_2$ as follows:

$$r_2 = \left(\frac{S_e}{S}\right)\exp\left(2\left(\frac{S_e}{S}-1\right)\right), \tag{11}$$

where $r_2$ has stronger dampening effect than $r_1$ . It should also be noted that while $r_1$ is indicative of relative importance of a bin (among various bins computed for an edge), $r_2$ is indicative of the relative trust of an edge (in comparison to other edges).

In order to illustrate the effect of $r_e^b$ (equation (9)), one example is shown in Fig. 34 and Fig. 35. Fig. 34 shows three edge curves, for whom centers have to be computed. The computations are collected in Fig. 35, where the space is divided into 10 bins on each side and 200 sets of points are formed for each edge. We show the original histograms and the $r_e^b$ histograms three edges from left to right in Fig. 35. It is evident that $r_e^b$ is a more distinctive histogram.

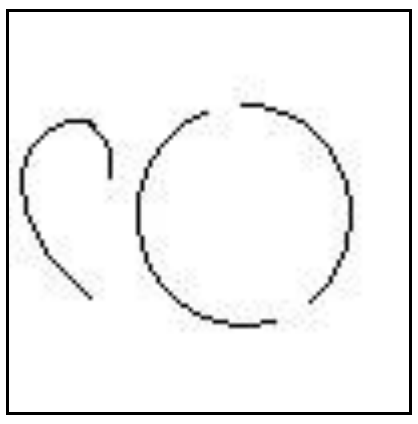

Fig. 34: A simple image for illustrating the effect of the $r_e^b$ histogram. The edges are numbered left to right as $e_1$, $e_2$, and $e_3$.

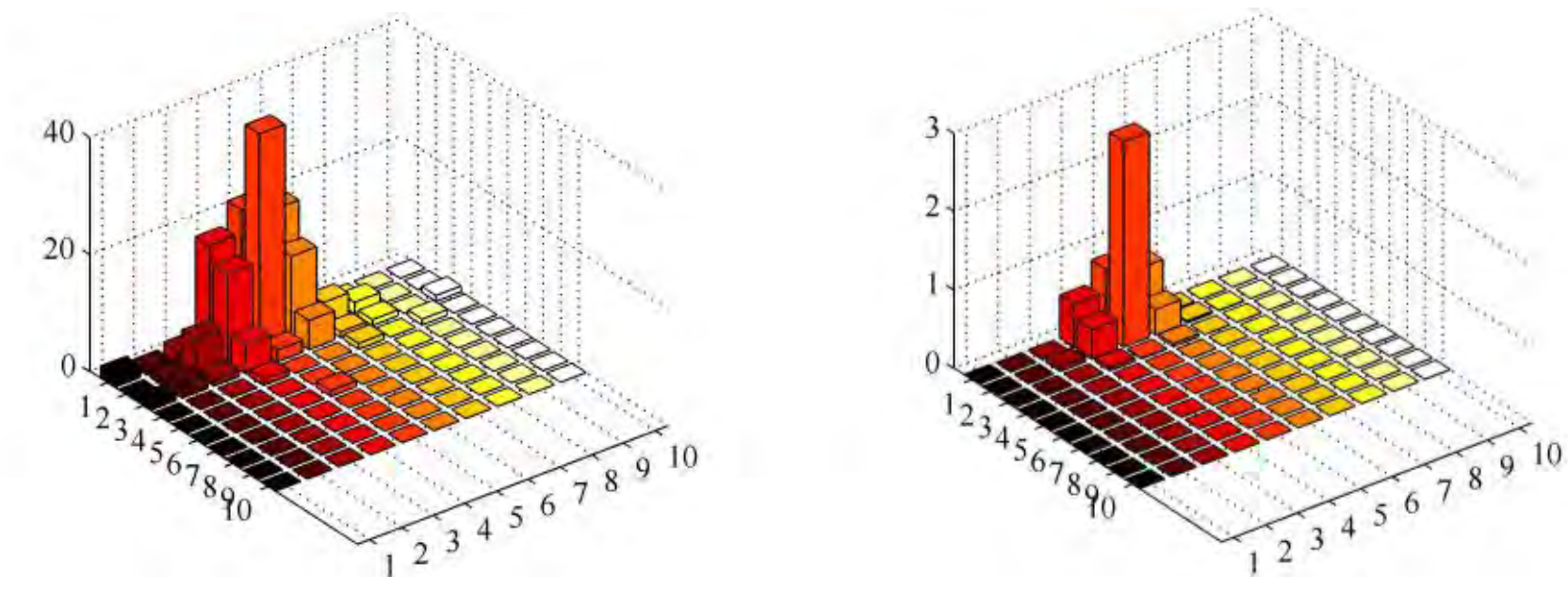

(a) The original and $r_e^b$ histograms for the edge $e_1$.

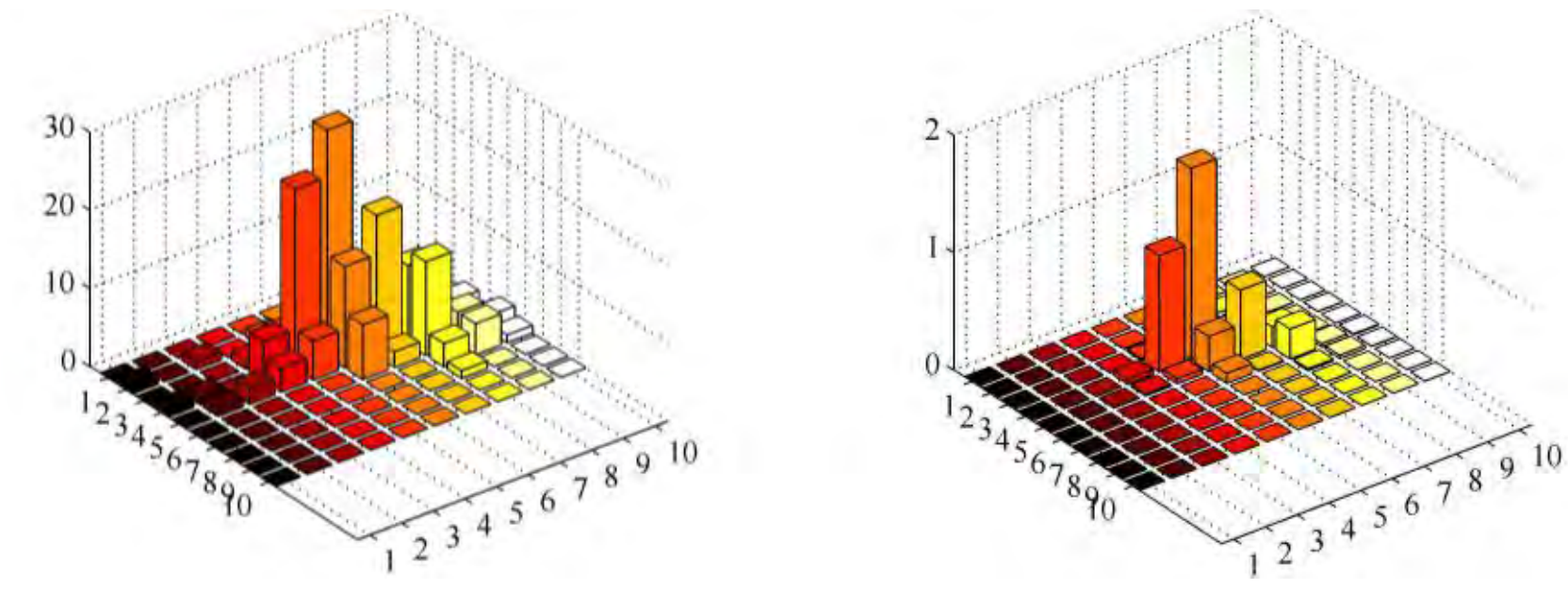

(b) The original and $r_e^b$ histograms for the edge $e_2$.

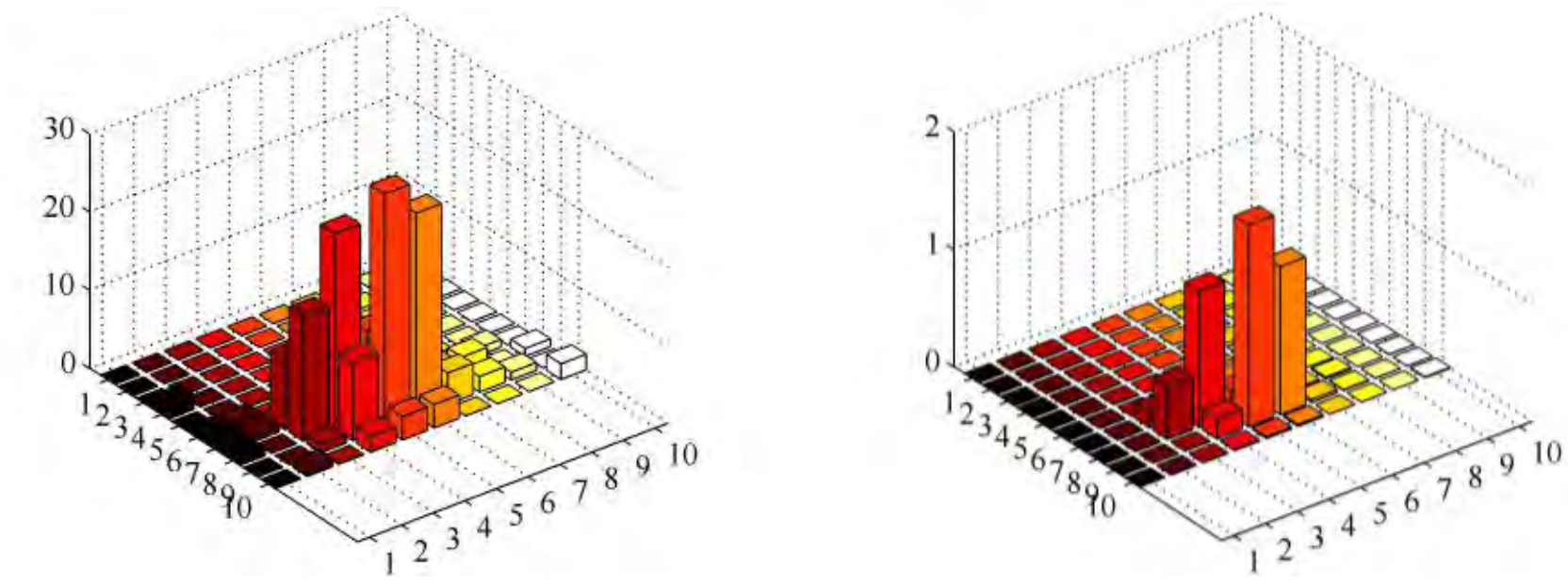

(c) The original and $r_e^b$ histograms for the edge $e_3$.

Fig. 35: Original and $r_e^b$ histogram for the three edges of the image in Fig. 34. The images on the left show the original histogram and the images on the right show the $r_e^b$ histogram.

### 3.5.5 Grouping the edges belonging to the same ellipse

In the proposed method, grouping does not mean physical merging/connecting of the edges. In the current context, grouping means collecting the edges that may possibly belong to same ellipse as one set.

After assigning the score, all the edges having a common bin $b$ may initially be considered as a group. In a group, it is reasonable to rank the various edges in the group based on their scores $r_e^b$. Thus various groups are formed for various bins and the edges in each group are ranked based on their scores. The edge pixels of the edges in a group are appended in the descending order of their scores and least squares fitting technique [189] is used to find all the parameters of the ellipse. Other precise ellipse fitting method can also be used [246-249]. Now, we judge the quality of this group based on two criteria listed below:

- Criterion 1 (C1): Error of least squares fitting $\leq \varepsilon_{ls}$, a chosen threshold error value.
- Criterion 2 (C2): The centre retrieved from least squares fitting is in the neighborhood of the reference center bin $b$.

If both C1 and C2 are satisfied, then the parameters of the ellipses computed using the least squares fitting are given as output. Otherwise the grouping is considered invalid.

The motivation of the first criterion is clear and well-understood. However, criterion 2 needs some discussion. Since the centers are generated using random selection of pixels for HT and not all pixels might have contributed for HT, for most groups, the center computed using least squares technique may not fall in the same bin as the group's bin. Thus, a margin should be allowed between the group's bin and the bin computed using least squares. This margin appears in the form of neighborhood. The neighborhood of the reference bin is specified by choosing a small window of $d \times d$ bins with the reference bin at the centre. The value of $d$ depends upon the bin size. If the bin size is large in comparison to the size of image (say 10 bins in along one direction), $d = 1$ typically suffices. If the bin size is small (say more than 20 bins in along one direction), $d = 5$ works satisfactorily. We have used $d = 5$ for all our numerical results.

### 3.5.6 The algorithm for stage 1 of the ellipse detection method

Step 1: The edges are arranged in the order of decreasing order of their edge lengths[1] $\{e_i\}$. Set $i = 1$.

Step 2: Set $G = e_i$. ($G$ signifies a set of edges, which may contain one or more edges).

Step 3: Find the edges lying in its search region (section 3.5.1) of $G$ and satisfying the convexity criterion (section 3.5.2) with $G$. Let the sequence of such edges be denoted by $\{e_j\}$.

Step 4: If $\{e_j\}$ is not empty, arrange $\{e_j\}$ in the increasing order of their distance[2] from $G$. Set $j = 1$. Go to Step 5.

If it is empty, apply the geometric center finding method (section 3.5.3), relationship score finding (section 3.5.4), and grouping (section 3.5.5) for the edge $G$. Extract the parameters of the ellipse (if it is elliptic, verified using the criteria in 3.5.5). Do $i = i + 1$ and go to step 2.

Step 5:Perform the geometric center finding method (section 3.5.3), relationship score finding (section 3.5.4), and grouping (section 3.5.5) for the edge $e_i$ and the first edge in $\{e_j\}$.

Step 6: If both criteria mentioned in section 3.5.5 are not satisfied, Do $j = j + 1$ and go to step 5.

If both criteria mentioned in section 3.5.5 are satisfied, then $G = G \cup e_j$. Go to step 3.

The flowchart of the above algorithm is shown in Fig. 36. Ideally, for an edge $e_i$, the recursion cycle of $G$ (step 3 to step 6 and back to step 3) should continue until $\{e_j\}$ is empty. However, practically, it may be computation intensive and time consuming. It may suffice in general to perform this operation for any edge till a predetermined number of recursion cycles (which we call the recursion depth $D$).

[1] In real images, the objects of interest often appear in the cluttered environment. Thus, it is reasonable to consider longer edges as foreground and very short edges as background. Thus, longer edges should be considered first for ellipse detection process.

[2] According to human perception, elements that are closer to each other will be perceived as a coherent object. Accordingly, among all the edges in the search region, the edges closer to the edge under consideration should be given higher priority for grouping process.

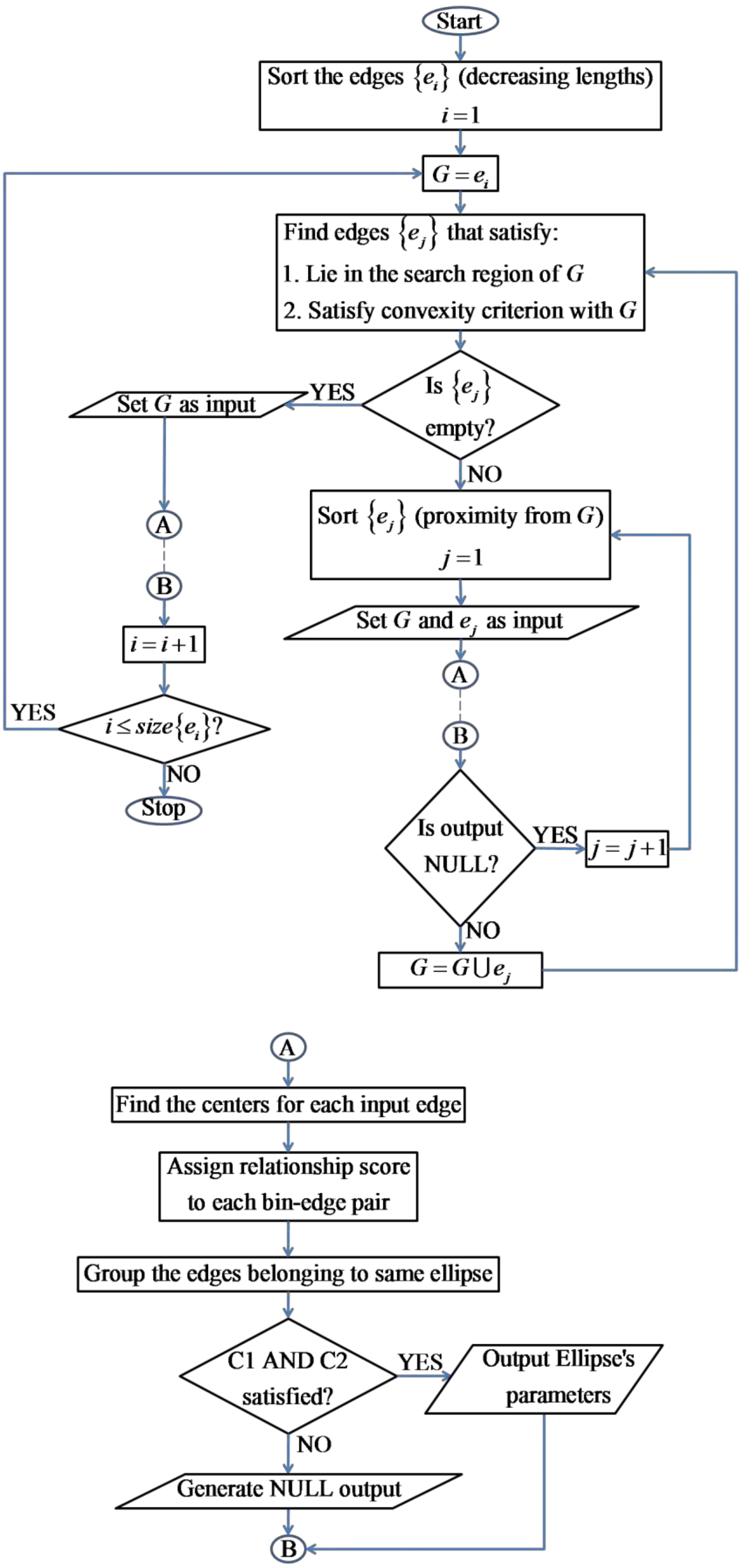

Fig. 36: The flowchart of the stage 1 for the proposed ellipse detection method.

## 3.6 Ellipse detection : stage 2 (elliptic hypotheses selection using saliency)

We perform the hypotheses selection in two steps. First, we cluster the elliptic hypotheses that are very similar to each other and choose the best representative among them [246, 250-253]. This increases the chances of one elliptic object being represented by a single elliptic hypothesis. The elliptic hypotheses that remain after the clustering are then evaluated for their saliency. We present three kinds of saliency criteria and combine them to select the more salient elliptic hypotheses.

### *3.6.1 Identification of similar ellipses*

Since every possible optimal grouping combination is considered independently, the method may result into more than one elliptic hypothesis for a single elliptic object. However, such elliptic hypothesis are characterized by small variation in their parameters (or close locality in the parameters space), since they belong to the same object. Thus, it is reasonable to cluster them and choose only one representative ellipse among them.

Here, we consider a simple Euclidean distance based metric. We represent each ellipse as a point in a five-dimensional parameter space as follows:

$$\bar{V}(x, y, a, b, \theta), \tag{12}$$

where $(x, y)$ is the center of the ellipse, $a$ and $b$ are the lengths of semi-major and semi-minor axes, and $\theta$ is the angle of orientation.

We propose the following method for defining the similarity between two ellipses with parameters $\{x_i, y_i, a_i, b_i, \theta_i\}$, $i = 1, 2$ . Let the differences between two ellipses be represented by:

$$D_x = \frac{|x_1 - x_2|}{X}, D_y = \frac{|y_1 - y_2|}{Y}, D_a = \frac{|a_1 - a_2|}{\max(a_1, a_2)}, D_b = \frac{|b_1 - b_2|}{\min(b_1, b_2)},. \tag{13}$$

$$D_\theta = \begin{cases} 0 & b_1/a_1 \geq 0.9 \text{ AND } b_2/a_2 \geq 0.9 \\ 1 & b_1/a_1 \geq 0.9 \text{ AND } b_2/a_2 < 0.9 \\ 1 & b_1/a_1 < 0.9 \text{ AND } b_2/a_2 \geq 0.9 \\ \dfrac{\angle(\theta_1, \theta_2)}{\pi} & b_1/a_1 < 0.9 \text{ AND } b_2/a_2 < 0.9 \end{cases} \tag{14}$$

where, $X$ and $Y$ are the number of pixels in the $x$ and $y$ directions respectively and $\angle(\theta_1, \theta_2)$ represents the smallest difference in the angles of orientation. In addition to the usual ellipses, (14) is valid for elliptic hypotheses that are close to circles as well.

It is evident that the above parameters represent the Euclidean distance normalized with respect to certain parameters. The detail regarding the choice of normalization parameters is discussed in Appendix D.

Let $\tilde{D}_x$, $\tilde{D}_y$, $\tilde{D}_a$, $\tilde{D}_b$, and $\tilde{D}_\theta$, all in the range $[0,1]$, be the maximal difference values tolerable for the parameters defined in (13). We have chosen their values to be 0.1, which corresponds to a maximum 10% difference in the individual parameters.

We can define a Boolean variable $D$ as:

$$D = \text{AND}\left\{\left(D_x < \tilde{D}_x\right),\left(D_y < \tilde{D}_y\right),\left(D_a < \tilde{D}_a\right),\left(D_b < \tilde{D}_b\right),\left(D_\theta < \tilde{D}_\theta\right)\right\}. \tag{15}$$

Finally, the choice of representative candidate should depend upon the reliability of the ellipses in the cluster. One way of determining the reliability is to choose the ellipse that was formed by maximum amount of data. Thus, we have used percentage circumference of an ellipse (introduced below) and its edge(s).

### *3.6.2 Criteria for saliency*

#### *3.6.2.1 Percentage circumference for an ellipse and its edge(s)*

This method has been used by researchers for a long time to choose the best elliptic hypotheses. Typically, the number of pixels of the edges that are used to generate an elliptic hypothesis is divided by the perimeter of the ellipse. This has been called as the pixel count feature in [236].

Suppose an ellipse $E$ was fitted to a group $G$, then we define a function $c(E,G)$ as below:

$$c(E,G) = \frac{\sum_{\forall e \in G} \alpha(E,e)}{2\pi}, \tag{16}$$

where $\alpha(E,e)$ is the angle subtended by the ends of the edge $e$ at the centre of the ellipse $E$. A higher value of $c(E,G)$ implies a larger support of $E$ on $G$.

#### *3.6.2.2 Percentage alignment of an ellipse with its edge(s)*

Another saliency criterion considers the distribution of the pixels around the elliptic hypothesis. This idea was first proposed by [238], though in the context of straight lines primarily. In the context of elliptic hypotheses, we present the following method.

We consider the pixels $\{P\}$ in the edge that generated an elliptic hypothesis, and compute their Euclidean distance, $d$, from the elliptic hypothesis. The lesser the Euclidean distance, the more reliable is an edge pixel for generating the elliptic hypothesis. After applying a threshold ($d_0 = 2$) on the Euclidean distance, we count the number of pixels that are reliable for the current hypothesis and normalize it with respect to the total number of pixels, $N_G$, in the edges that generated the hypothesis as shown below:

$$s(E,P) = \begin{cases} 1 & if\ d < d_0 \\ 0 & \text{otherwise} \end{cases}, \tag{17}$$

$$a(E,G) = \frac{\sum_{i=1}^{N_G} s(E,P_i)}{N_G}, \tag{18}$$

The higher the value of $a(E,G)$, the better is the fit between ellipse $E$ and group $G$.

*3.6.2.3Angular continuity ratio*

Another criterion for choosing salient hypotheses is based on the angular continuity of the edges that generated a hypothesis. Let us consider two edges as shown in Fig. 37. The angle between the two intersecting tangents made at the two nearest end points of $e_1$ and $e_2$, $\theta_{\text{diff}}$, is the angle that determines the continuity between the two edges. It can have a maximum value $\pi$. Thus, the ratio of the angle, $\theta_{\text{diff}}$, and $\pi$ is an indicator of the continuity between the two edges. Similar idea was proposed in [197].

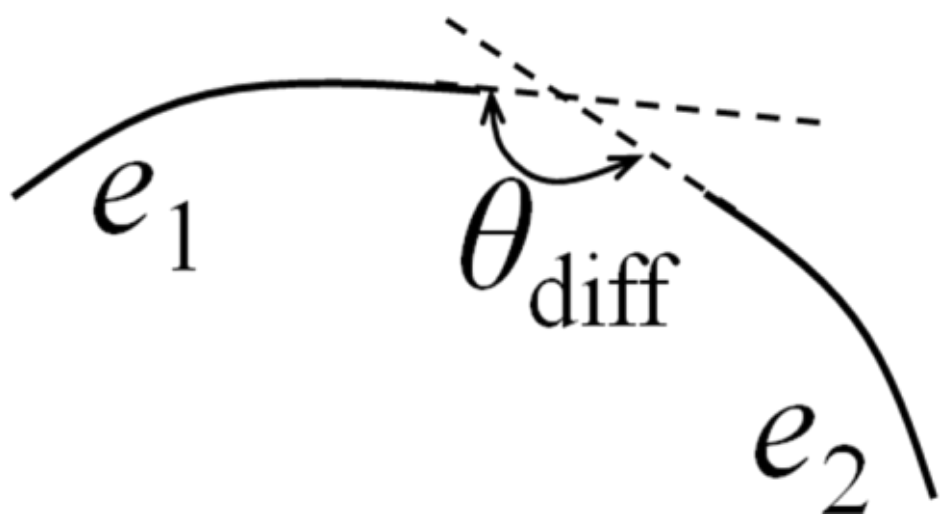

Fig. 37: Illustration of the angle used for generating the angle continuity ratio.

The angular continuity ratio is defined as:

$$\phi(E,G) = \begin{cases} 1 & \text{if}\, N = 1 \\ \frac{1}{N-1}\sum_{i=1}^{N-1} \theta_{\text{diff}}\left(e_i, e_{i+1}\right) & \text{if}\, N > 1 \end{cases}, \tag{19}$$

where $N$ is the number of edge curves in the group $G$. It is worth noting that if an elliptic object has large $\phi(E,G)$, the reliability of such elliptic hypotheses is better than an elliptic object that appears in the form of far apart contours.

### *3.6.3 Combining the three criteria and making the final decision*

There are various ways of combining the saliency criteria. We propose to use the additive combination, $\sigma_{\text{add}}(E,G)$ as:

$$\sigma_{\text{add}}(E,G) = \frac{a(E,G) + c(E,G) + \phi(E,G)}{3}. \tag{20}$$

Now, the decision of selecting the elliptic hypothesis $E$ is made using the Boolean outcome of the expression below:

$$\text{AND}\left(a(E,G) \ge avg\{a(E,G)\},\ c(E,G) \ge avg\{c(E,G)\},\ \phi(E,G) \ge avg\{\phi(E,G)\},\ \sigma_{add}(E,G) \ge avg\{\sigma_{add}(E,G)\}\right). \tag{21}$$

Here, $avg\{a(E,G)\}$ is the average value of the alignment percentages calculated for all the elliptic hypotheses remaining after the similar ellipses identification. The same applies for the other expressions in (21). This method assures that the selected hypotheses perform better than average in every criterion and have overall good saliency.

## 3.7 Experimental results

### *3.7.1 Performance evaluation*

We present results for the proposed algorithm on a set of challenging synthetic and real images. The following measures are used for evaluating the performance of the proposed ellipse detection method:

$$\text{Precision}=\frac{\text{number of true postive elliptic hypotheses}}{\text{total number of elliptic hypotheses}}, \tag{22}$$

$$\text{Recall}=\frac{\text{number of true postive elliptic hypotheses}}{\text{number of actual ellipses}}, \tag{23}$$

$$\text{F-measure}=\frac{2\times\text{Precision}\times\text{Recall}}{\text{Precision}+\text{Recall}}, \tag{24}$$

where the true positive elliptic hypotheses are the hypotheses that have the relative root mean square error of less than 0.05 (with respect to the actual ellipses).

### *3.7.2 Synthetic dataset*

We test the proposed method under various scenarios such as occluded ellipses and overlapping ellipses using synthetic images. To generate the synthetic images, we consider an image size of $300\times300$ and generate $\alpha\in\{4,8,12,16,20,24\}$ ellipses randomly within the region of image. The parameters of the ellipses are generated randomly: center points of the ellipses are arbitrarily located within the image, lengths of semi-major and semi-minor axes are assigned values randomly from the range $\left[10,300/\sqrt{2}\right]$, and the orientations of the ellipses are also chosen randomly. The only constraint applied is that each ellipse must be completely contained in the image and overlap with at least one ellipse.

**Occluded ellipses** – For each value of $\alpha$, 100 synthetic images with occluded ellipses are generated and the proposed ellipse detection method is applied to each of the synthetic images. The average values of the measures in (22)-(24) for each $\alpha$ are calculated and plotted in Fig. 38. We show the comparison of the proposed method with the Simplified Hough Transform (SHT [205]) Randomized Hough Transform (RHT [217]), the method proposed in [190] (Kim), the method proposed in [194] (Mai), the method proposed in [195] (Chia). The results for a few images are shown in Fig. 39 - Fig. 44.

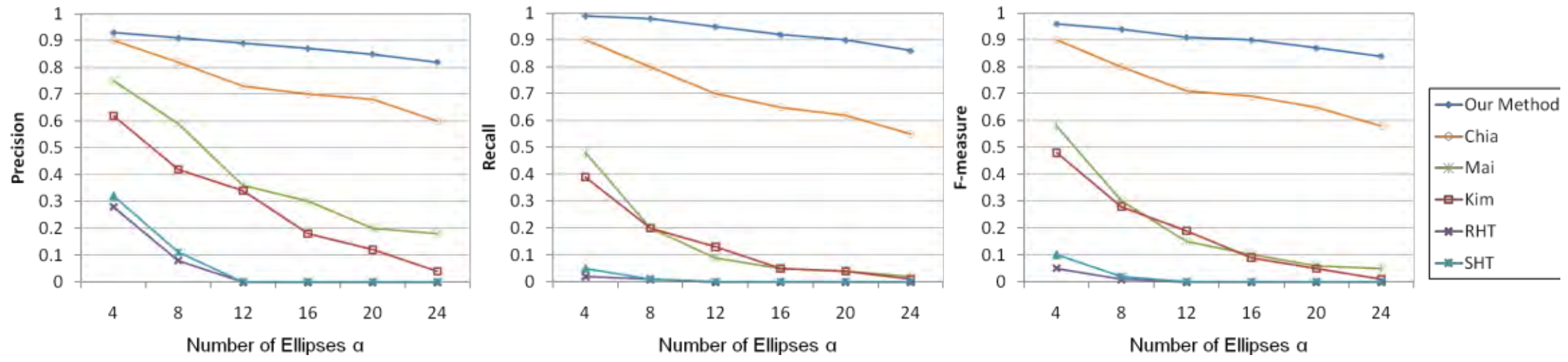

Fig. 38: Comparison of the proposed methods with various existing methods for images with occluded ellipses. For each value of $\alpha$, 100 synthetic images were generated. The result for each value of $\alpha$ is the mean of the measures for those 100 images.

(a) Original Image (b) Edge curves (c) Detected ellipses

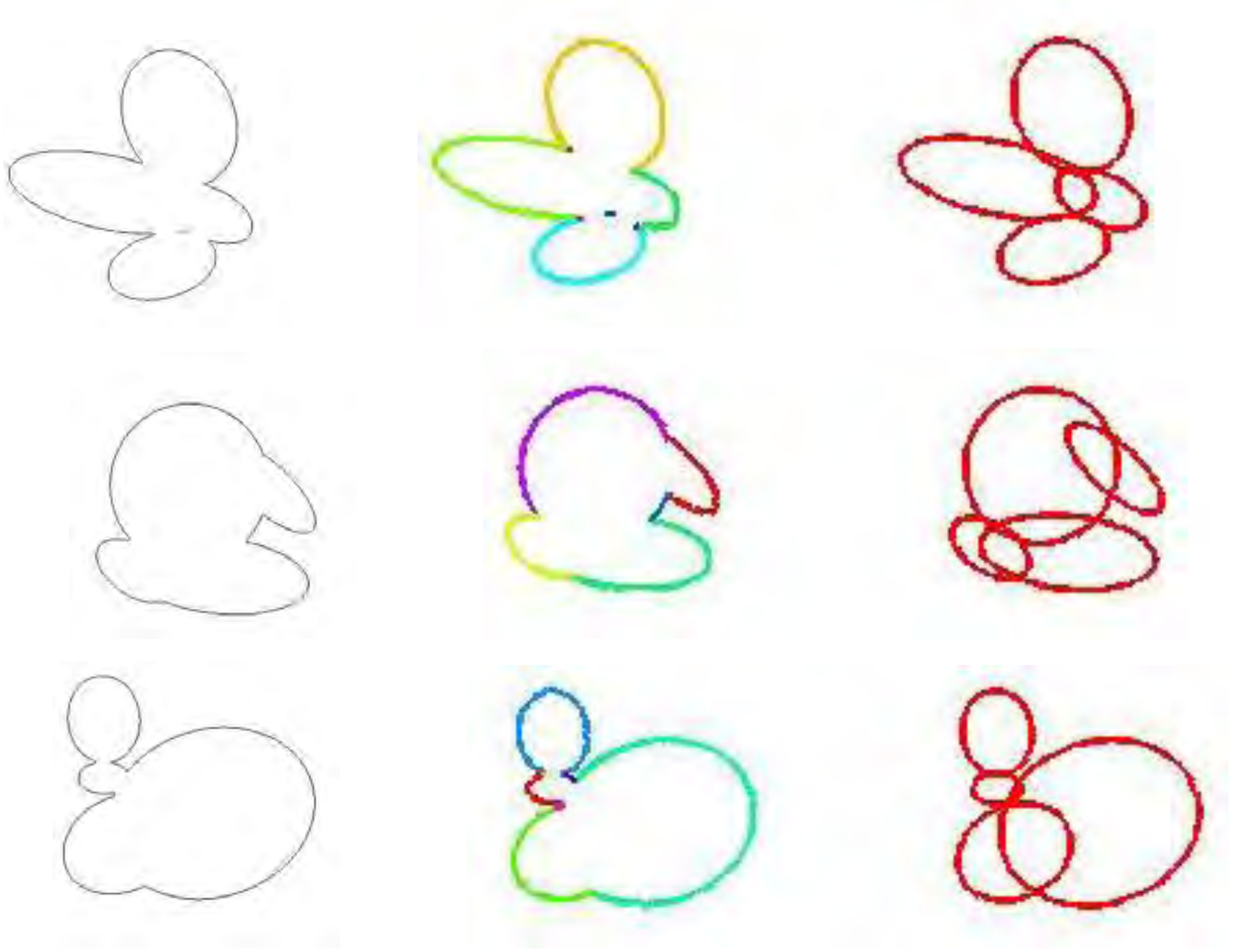

Fig. 39: Synthetic images with 4 occluded ellipses - detected ellipses

(a) Original Image (b) Edge curves (c) Detected ellipses

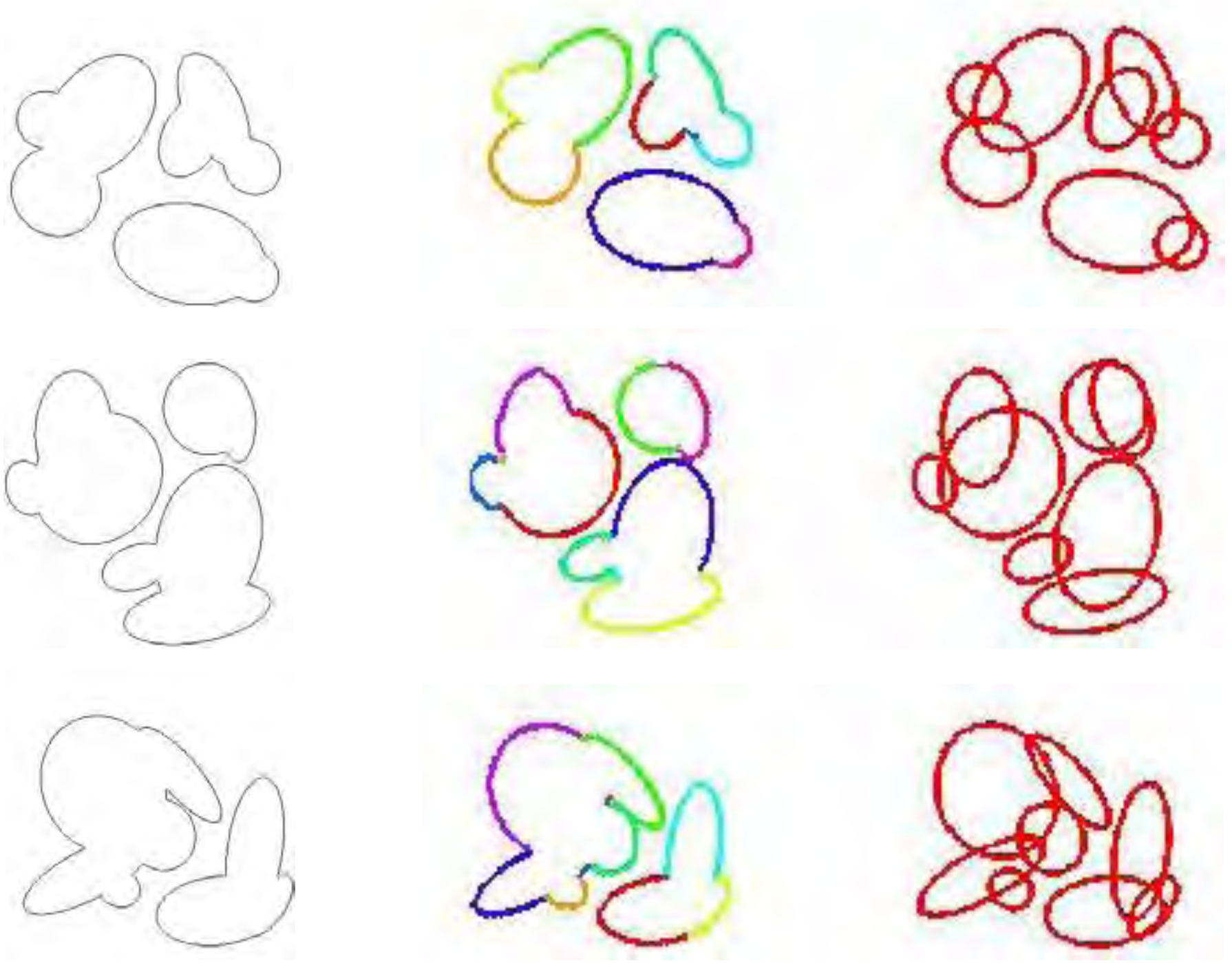

Fig. 40: Synthetic images with 8 occluded ellipses - detected ellipses

(a) Original Image (b) Edge curves (c) Detected ellipses

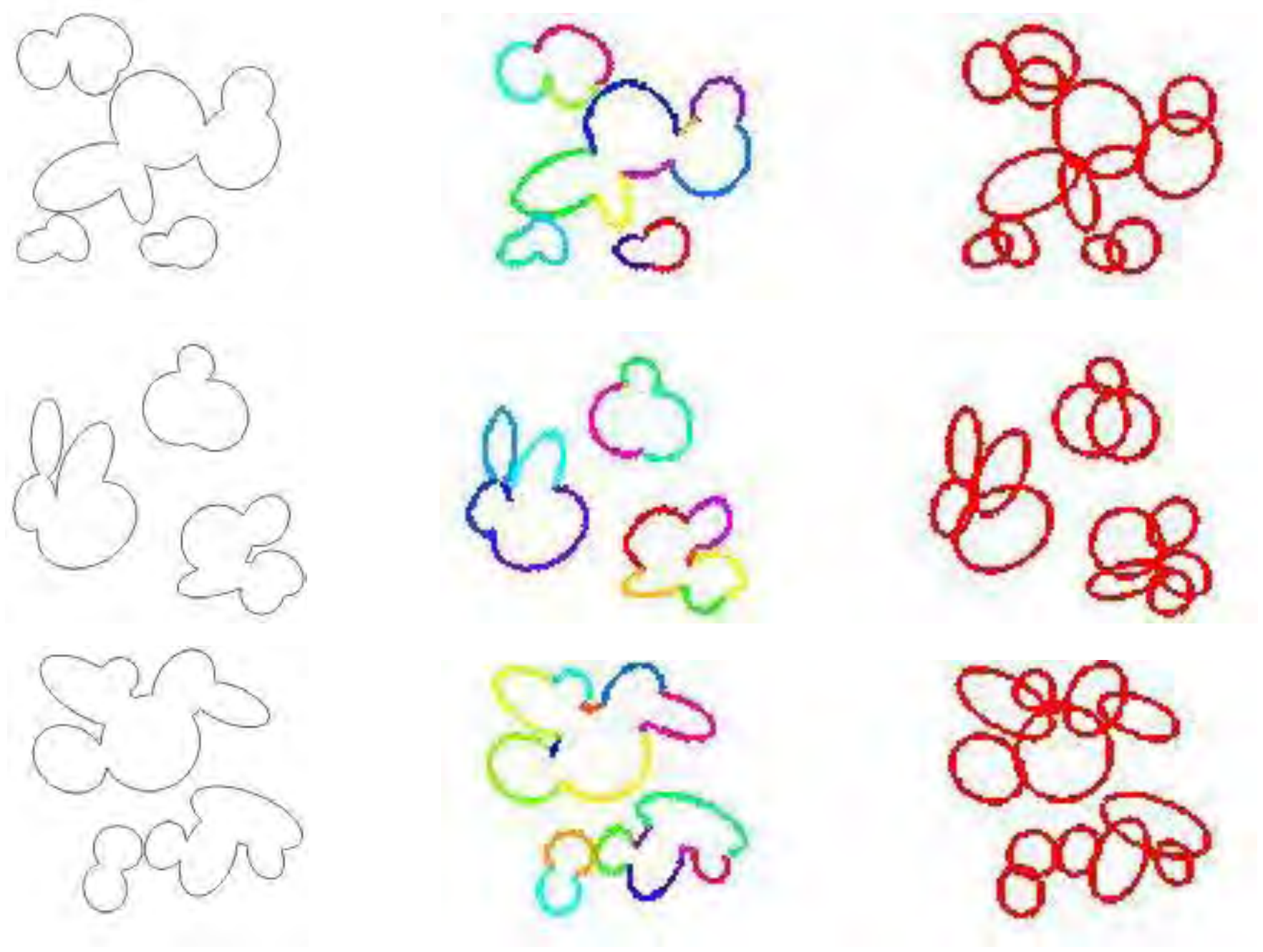

Fig. 41: Synthetic images with 12 occluded ellipses - detected ellipses

(a) Original Image (b) Edge curves (c) Detected ellipses

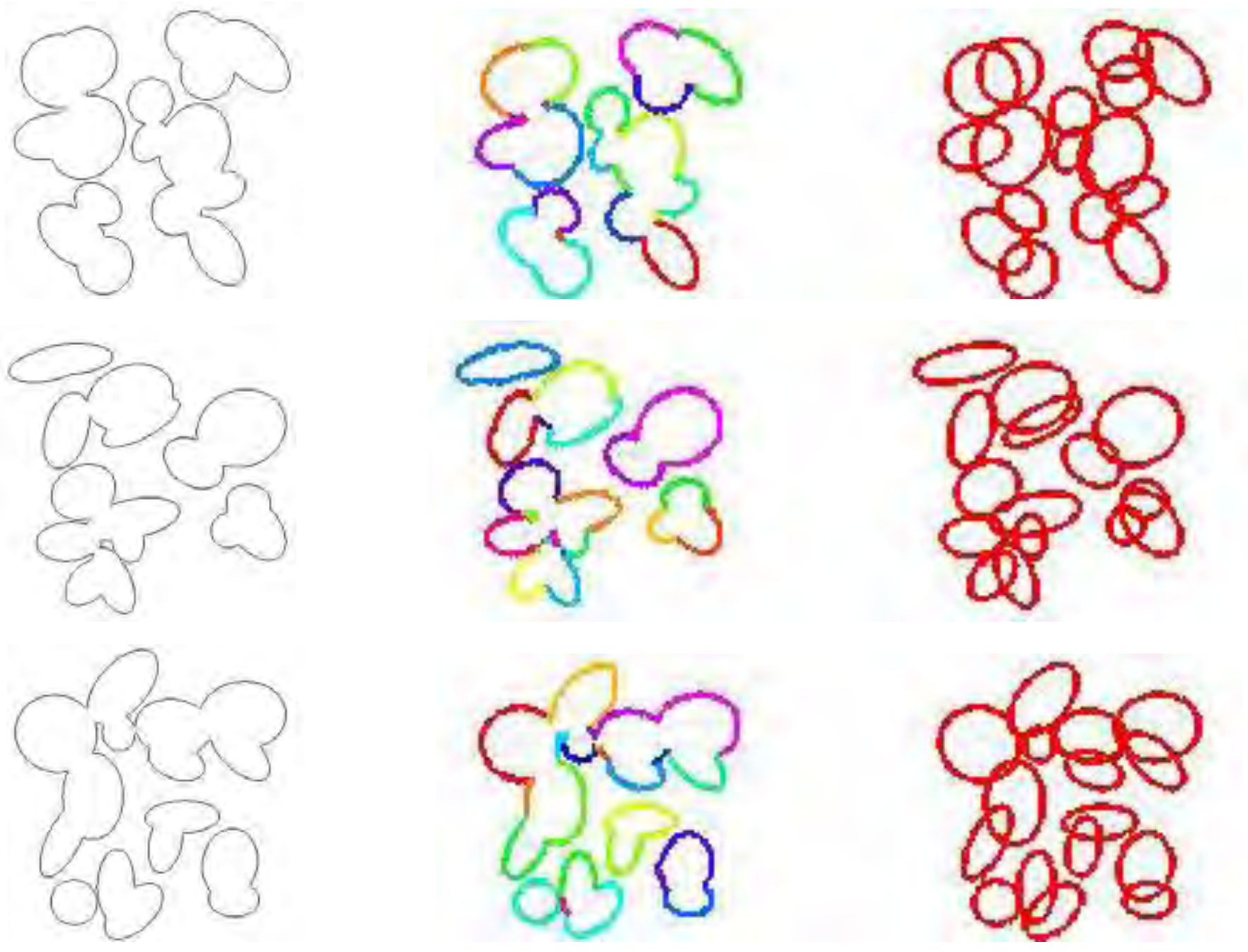

Fig. 42: Synthetic images with 16 occluded ellipses - detected ellipses

(a) Original Image (b) Edge curves (c) Detected ellipses

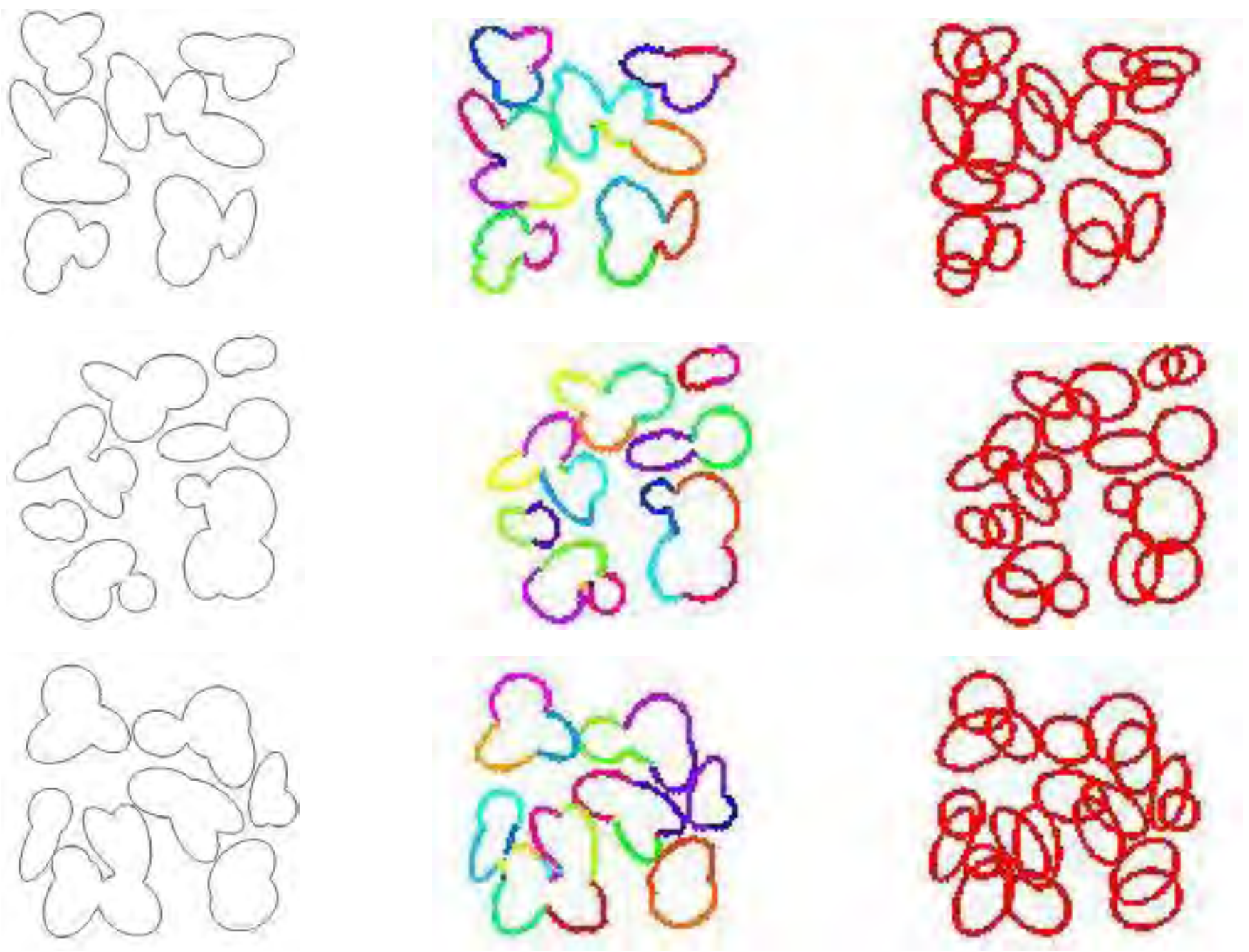

Fig. 43: Synthetic images with 20 occluded ellipses - detected ellipses

(a) Original Image (b) Edge curves (c) Detected ellipses

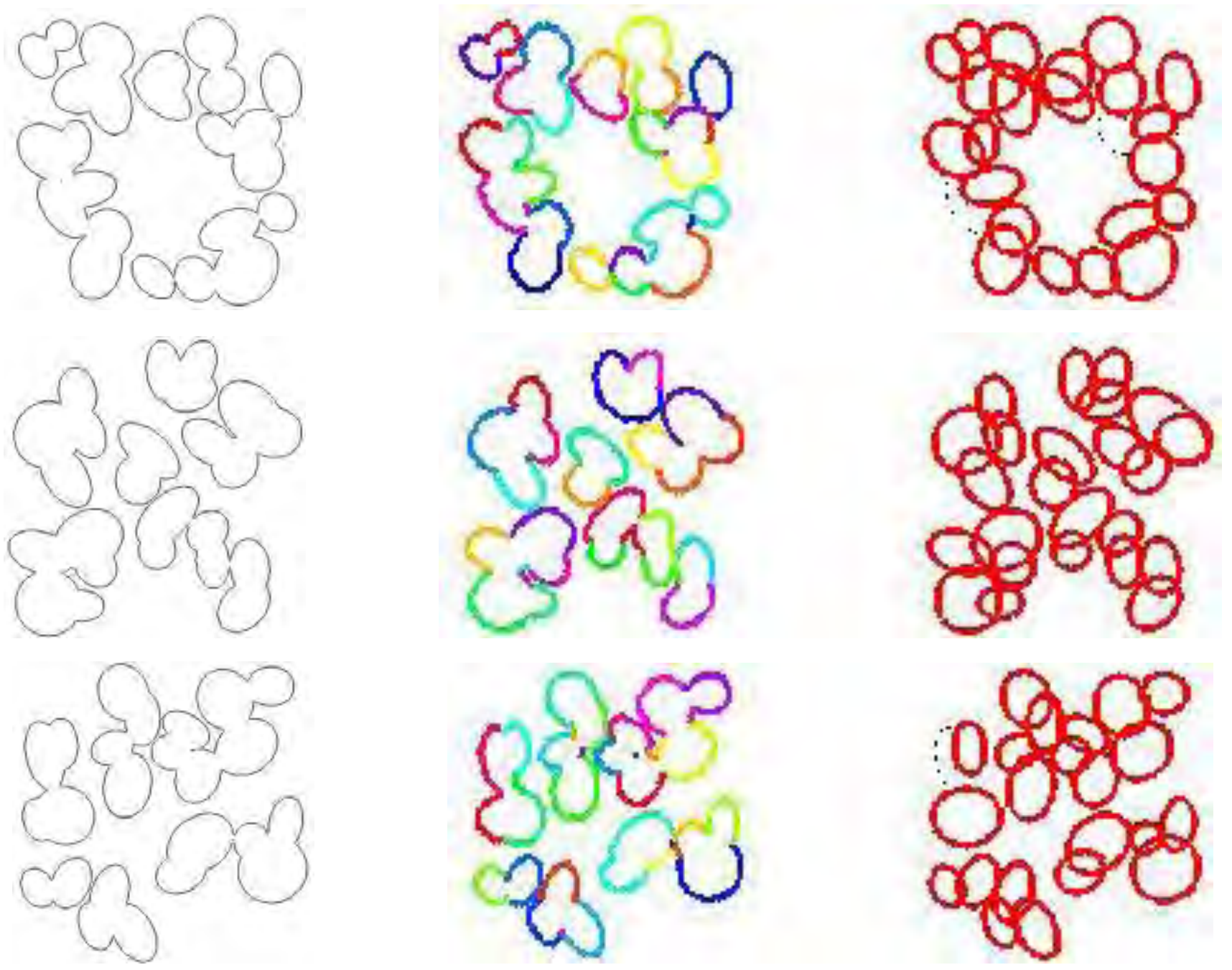

Fig. 44: Synthetic images with 24 occluded ellipses - detected ellipses

**Overlapping ellipses** – For each value of $\alpha$, 100 synthetic images with overlapping ellipses are generated and the proposed ellipse detection method is applied to each of the synthetic images. The average values of the measures in (22)-(24) for each $\alpha$ are calculated and plotted in Fig. 45. We show the comparison of the proposed method with the Simplified Hough Transform (SHT [205]) Randomized Hough Transform (RHT [217]), the method proposed in [190] (Kim), the method proposed in [194] (Mai), the method proposed in [195] (Chia). The results for a few images are shown in Fig. 46 - Fig. 51.

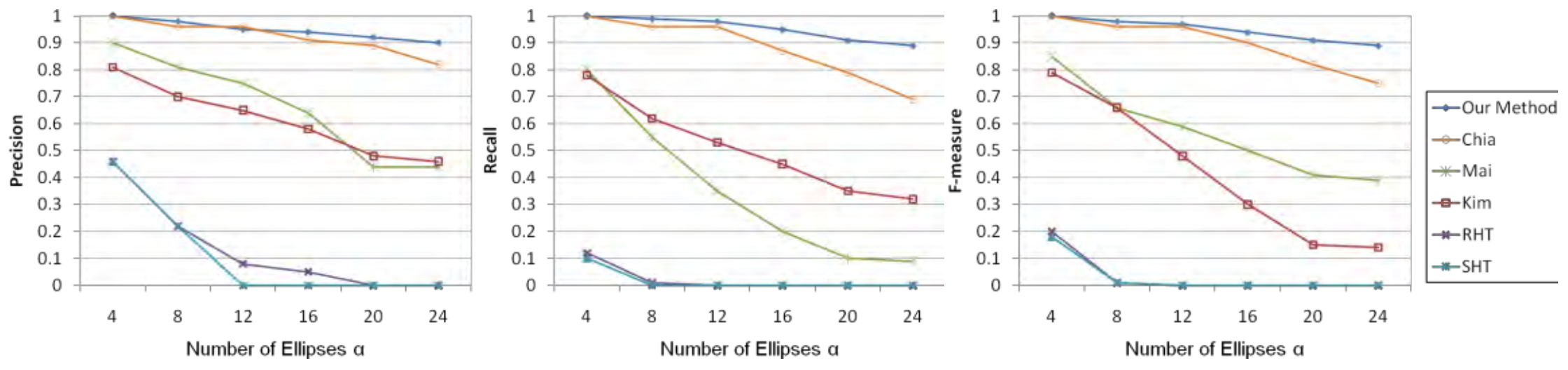

Fig. 45: Comparison of the proposed methods with various existing methods for images with overlapping ellipses. For each value of $\alpha$ , 100 synthetic images were generated. The result for each value of $\alpha$ is the mean of the measures for those 100 images.

(a) Original Image (b) Edge curves (c) Detected ellipses

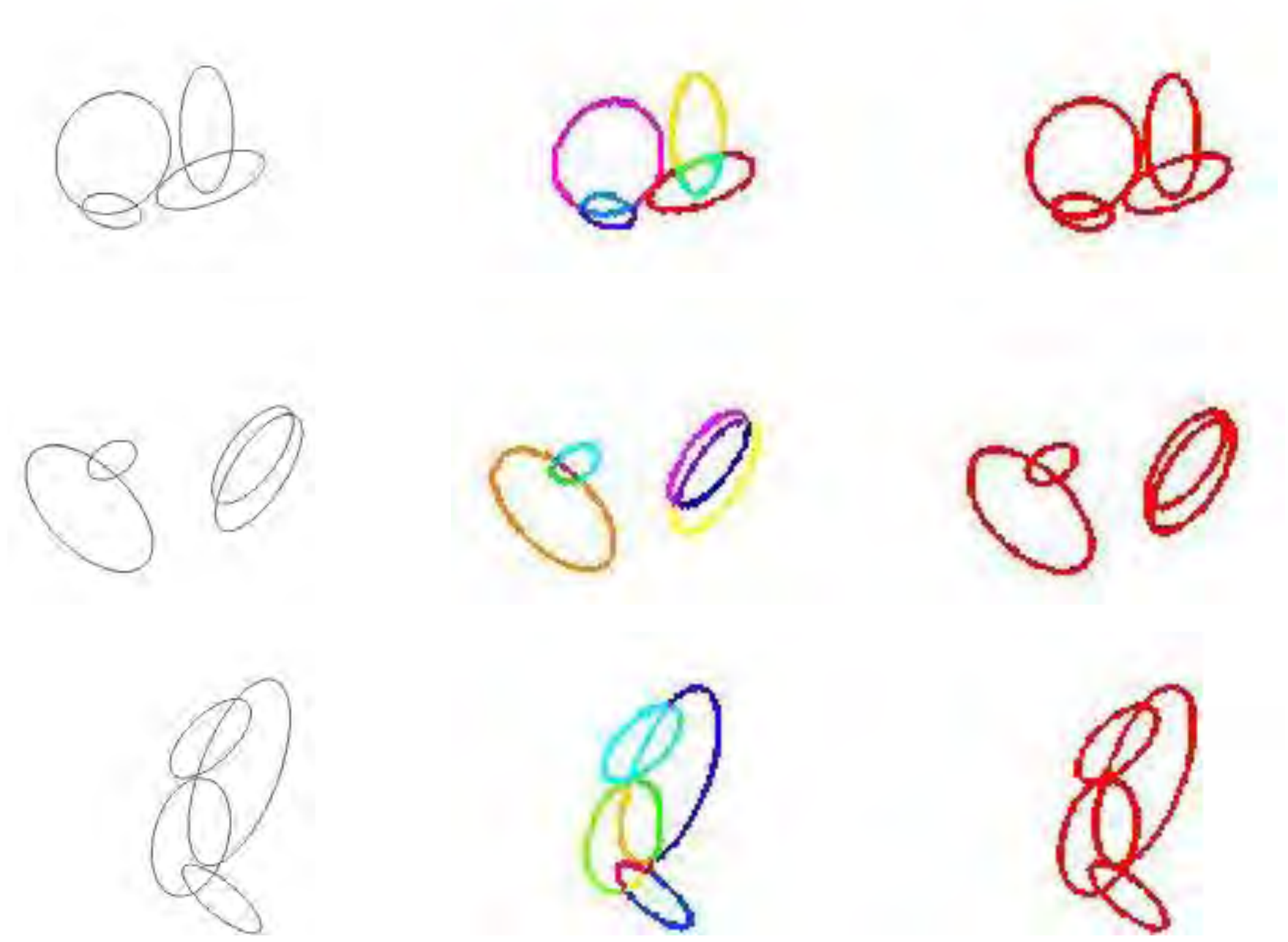

Fig. 46: Synthetic images with 4 overlapping ellipses - detected ellipses

(a) Original Image (b) Edge curves (c) Detected ellipses

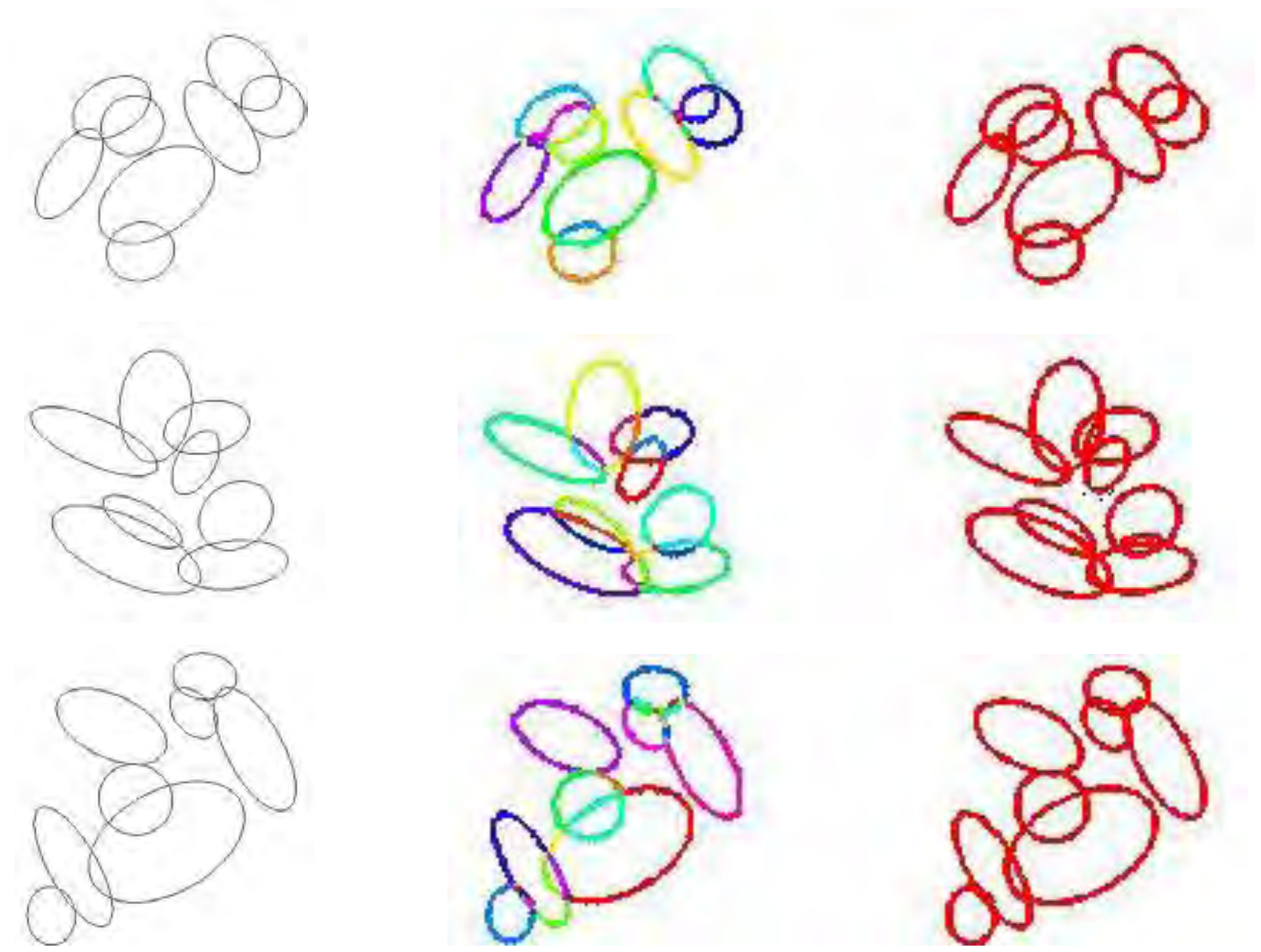

Fig. 47: Synthetic images with 8 overlapping ellipses - detected ellipses

(a) Original Image (b) Edge curves (c) Detected ellipses

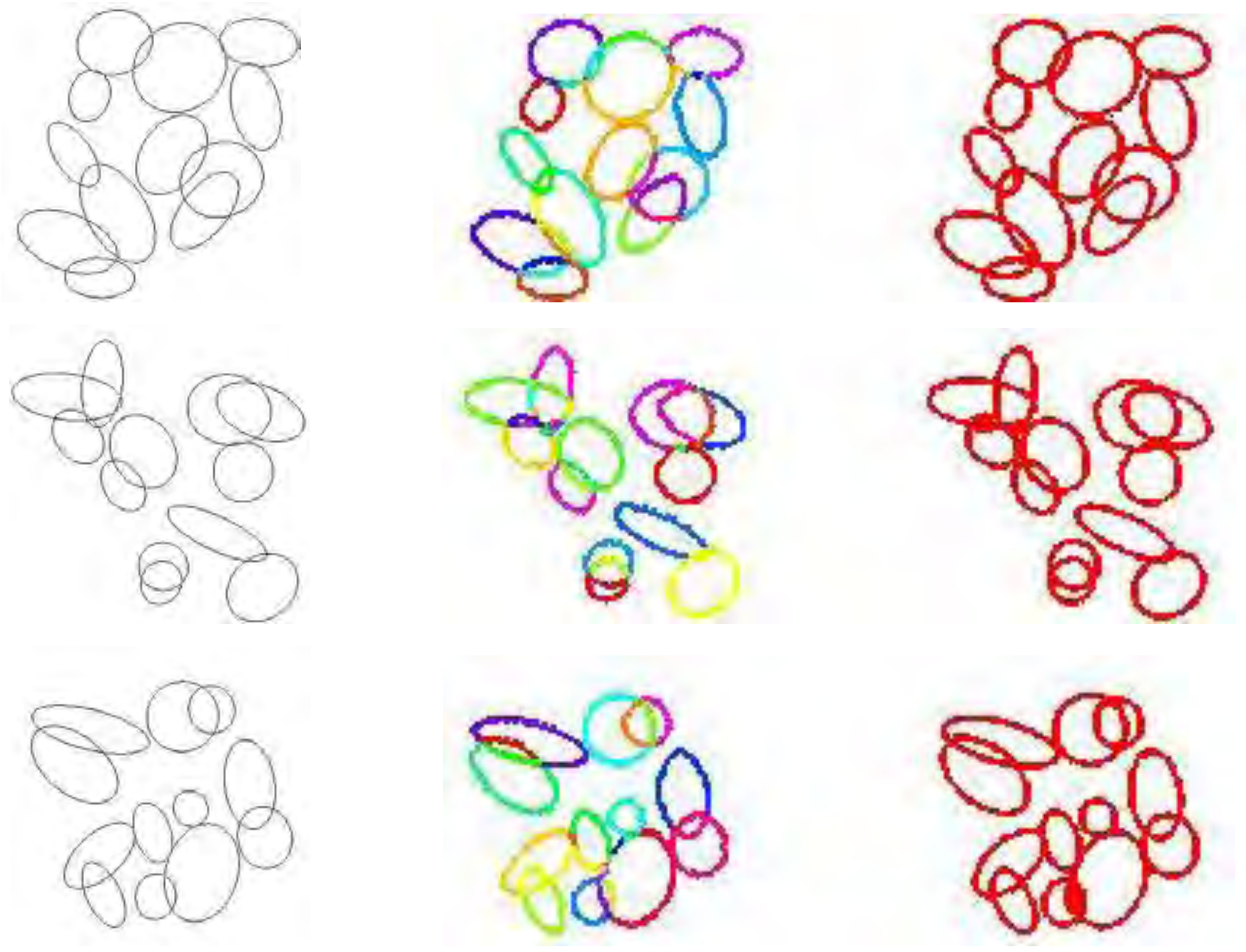

Fig. 48: Synthetic images with 12 overlapping ellipses - detected ellipses

(a) Original Image (b) Edge curves (c) Detected ellipses

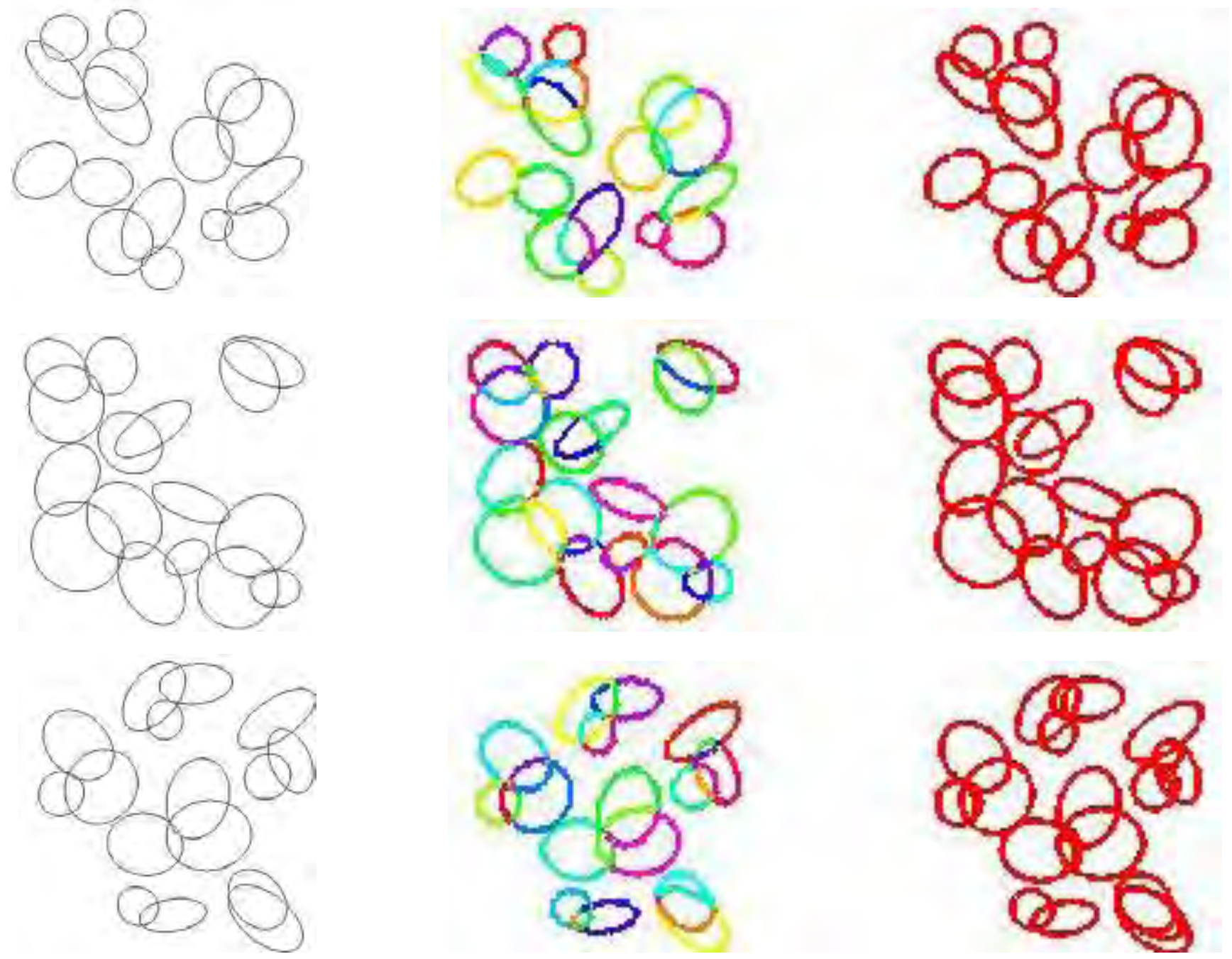

Fig. 49: Synthetic images with 16 overlapping ellipses - detected ellipses

(a) Original Image (b) Edge curves (c) Detected ellipses

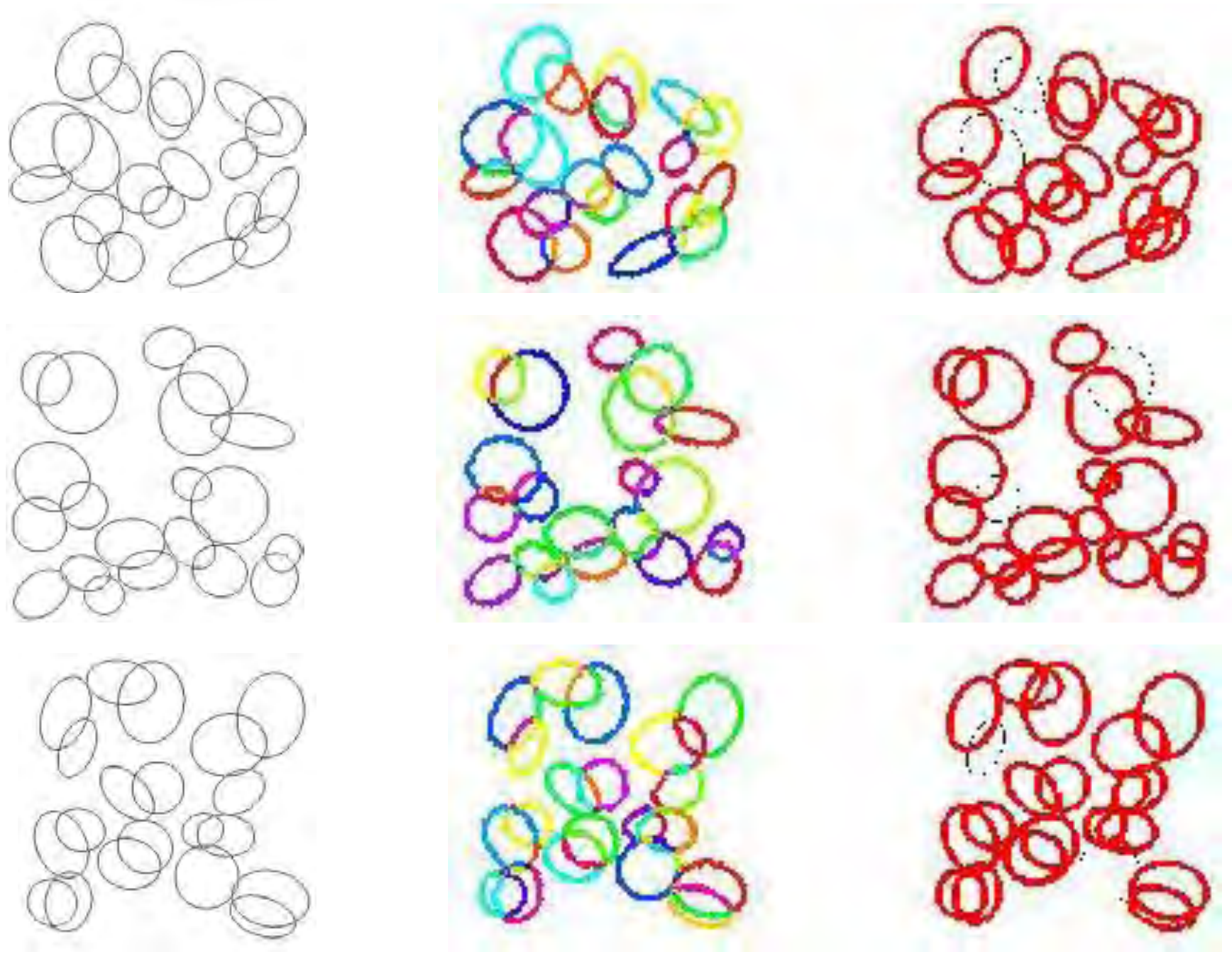

Fig. 50: Synthetic images with 20 overlapping ellipses - detected ellipses

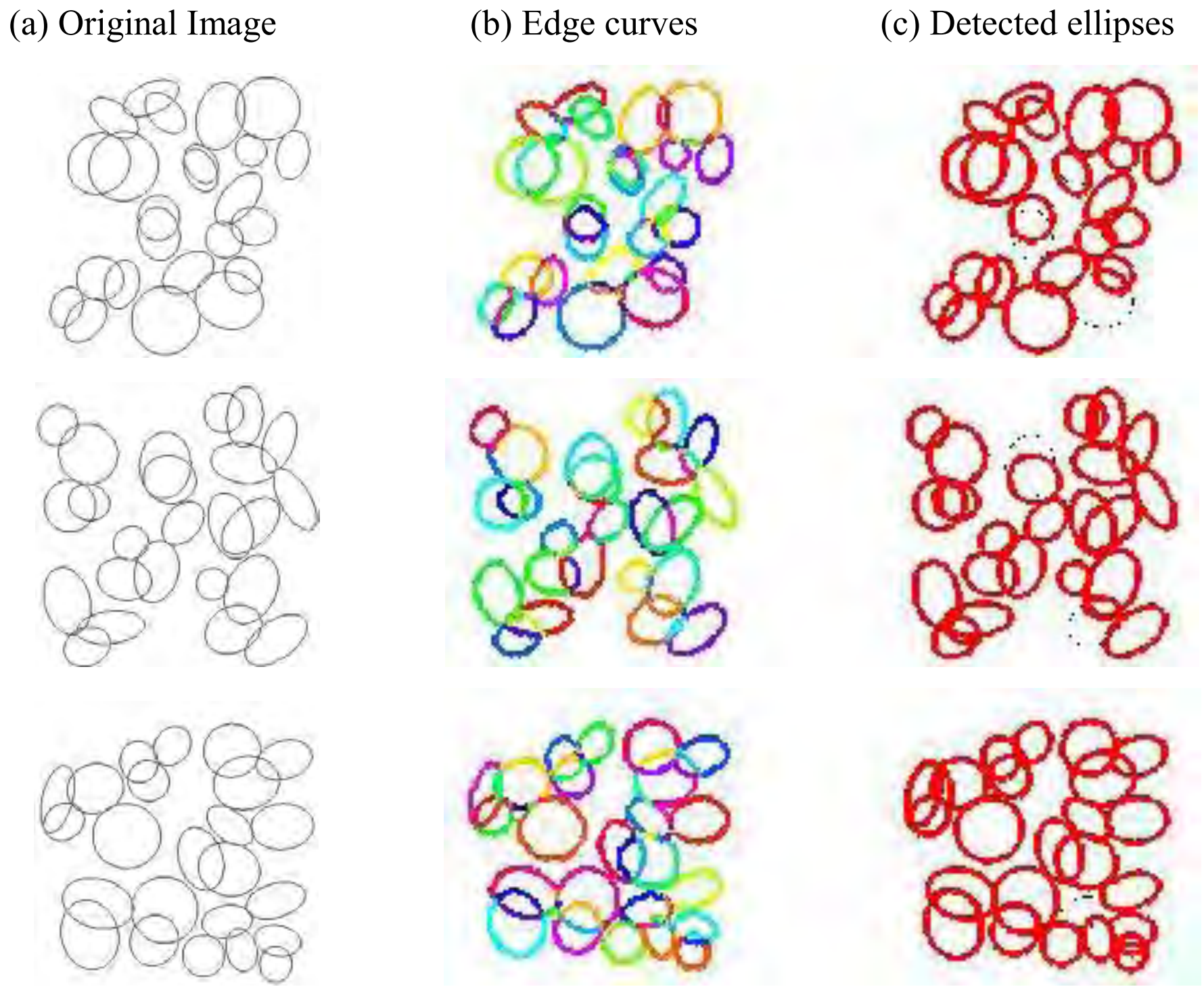

Fig. 51: Synthetic images with 24 overlapping ellipses - detected ellipses

It is evident that the proposed method shows good performance in either cases, even when the number of ellipses is significantly large.

### 3.7.3 *Real dataset*

Next, we show results of few test images from the Caltech-256 database [242]. These images present greater challenges than those tested above due to corruption of the contours of elliptical shaped objects by complex and varied backgrounds, illumination variations, partial occlusions, image noise, shadows and spectral reflections. For these images, we have used the depth of recursion $D = 2$. We present the results for 37 categories in Fig. 52-Fig. 57.

(a) Original Image (b) Canny edge map (c) Extracted edge contours (d) Detected ellipses

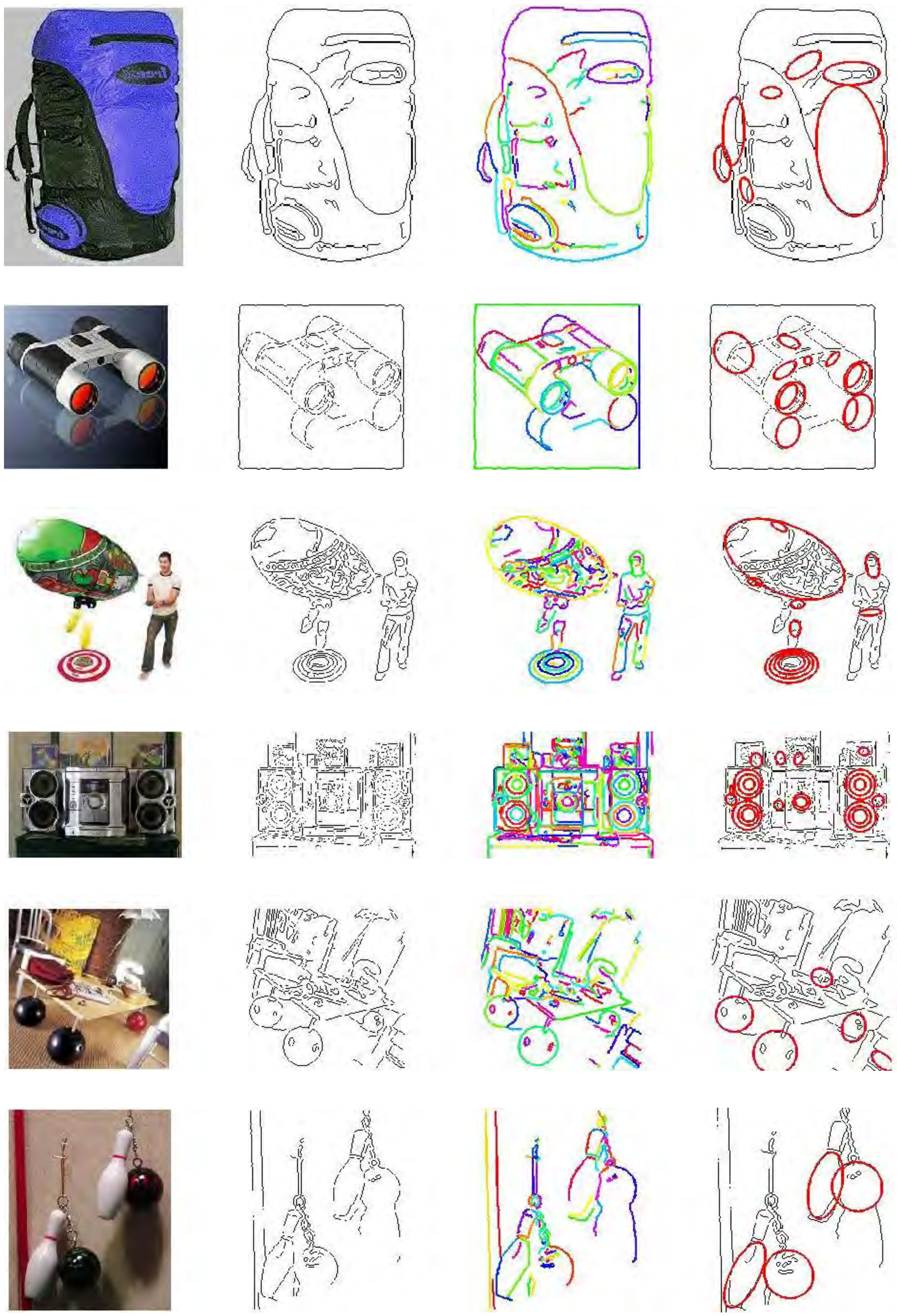

Fig. 52: Examples of real images and ellipse detection using the proposed method

(a) Original Image (b) Canny edge map (c) Extracted edge contours (d) Detected ellipses

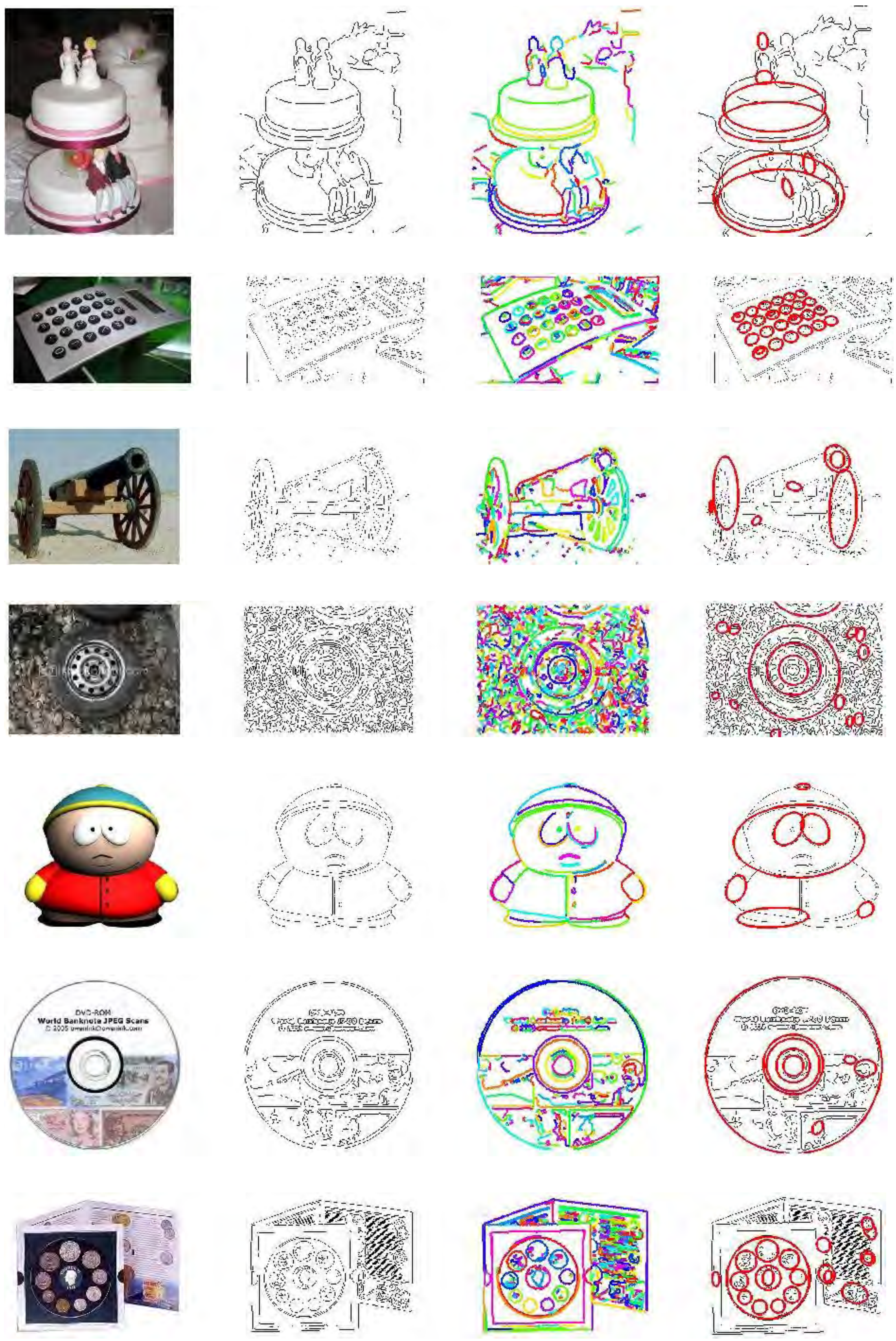

Fig. 53: Examples of real images and ellipse detection using the proposed method (continued).

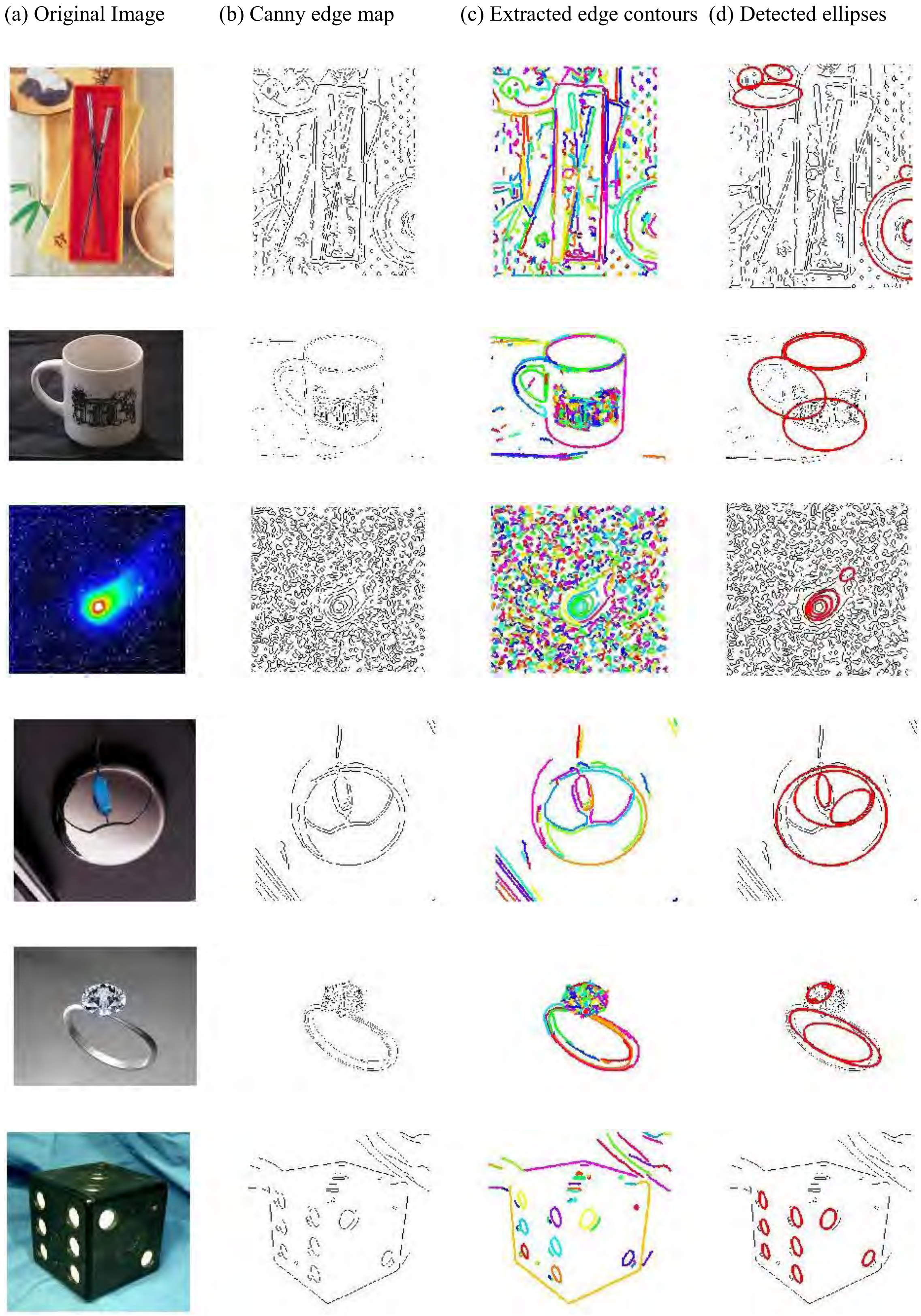

Fig. 54: Examples of real images and ellipse detection using the proposed method (continued).

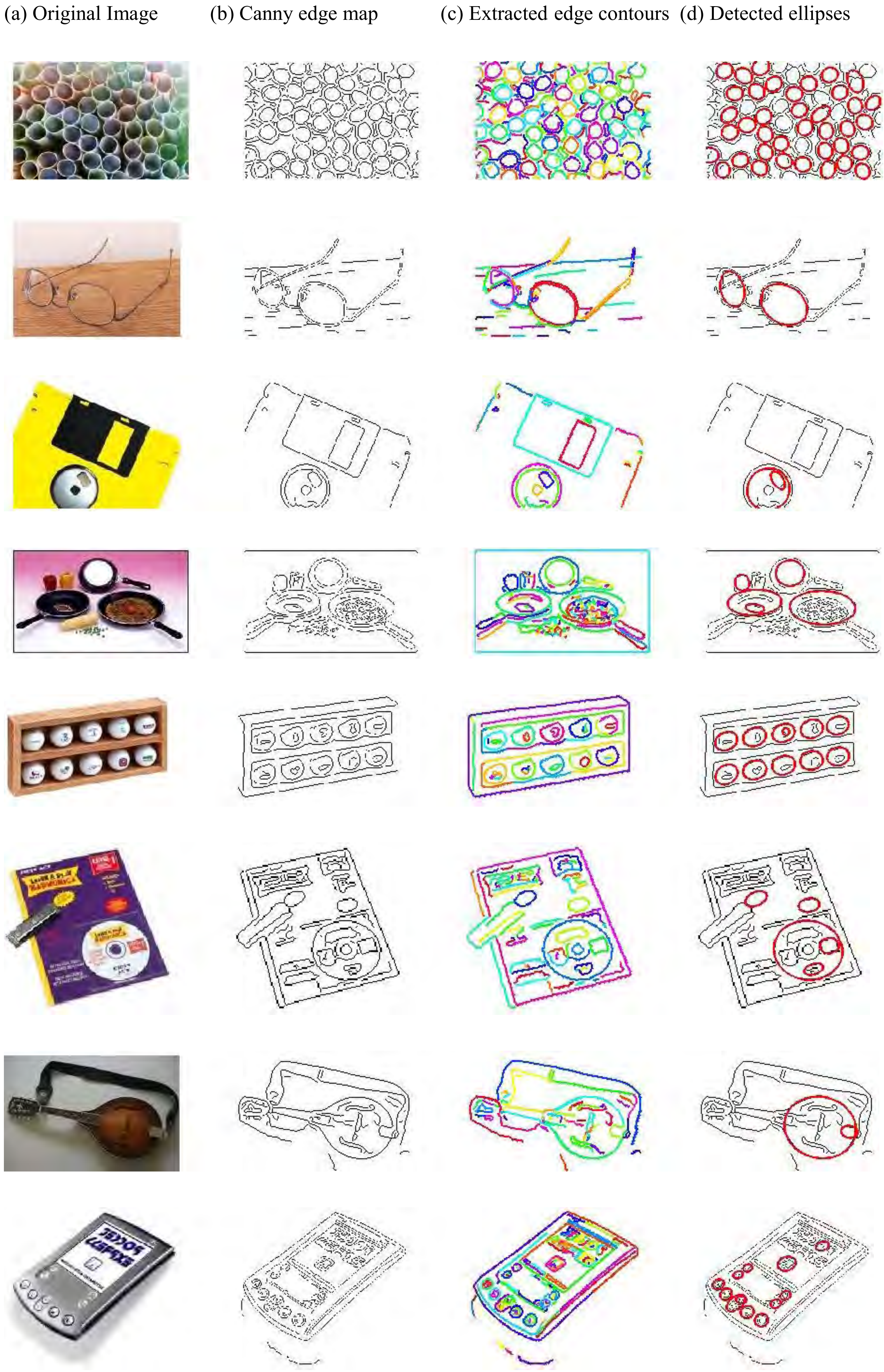

Fig. 55: Examples of real images and ellipse detection using the proposed method (continued).

(a) Original Image (b) Canny edge map (c) Extracted edge contours (d) Detected ellipses

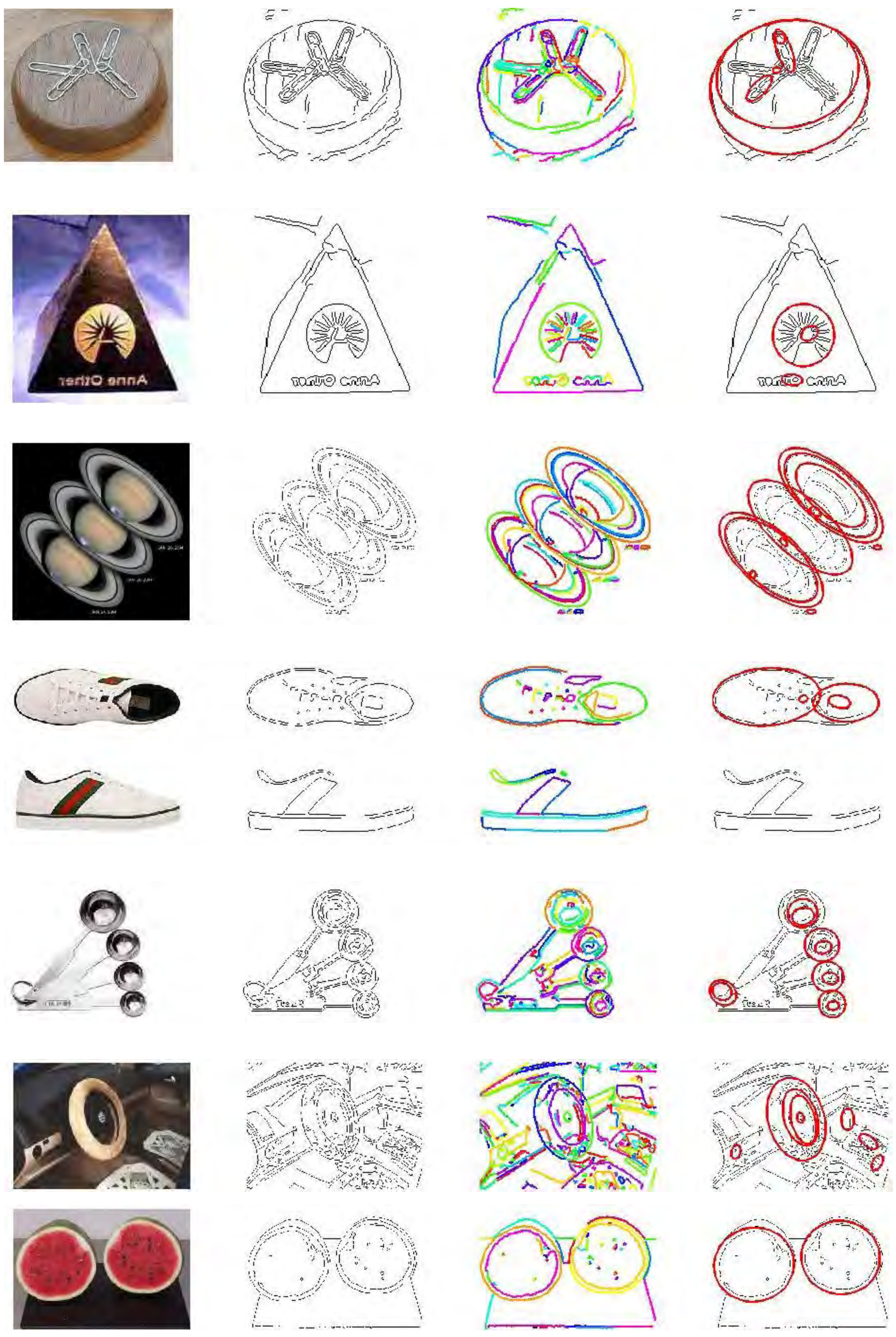

Fig. 56: Examples of real images and ellipse detection using the proposed method (continued).

(a) Original Image (b) Canny edge map (c) Extracted edge contours (d) Detected ellipses

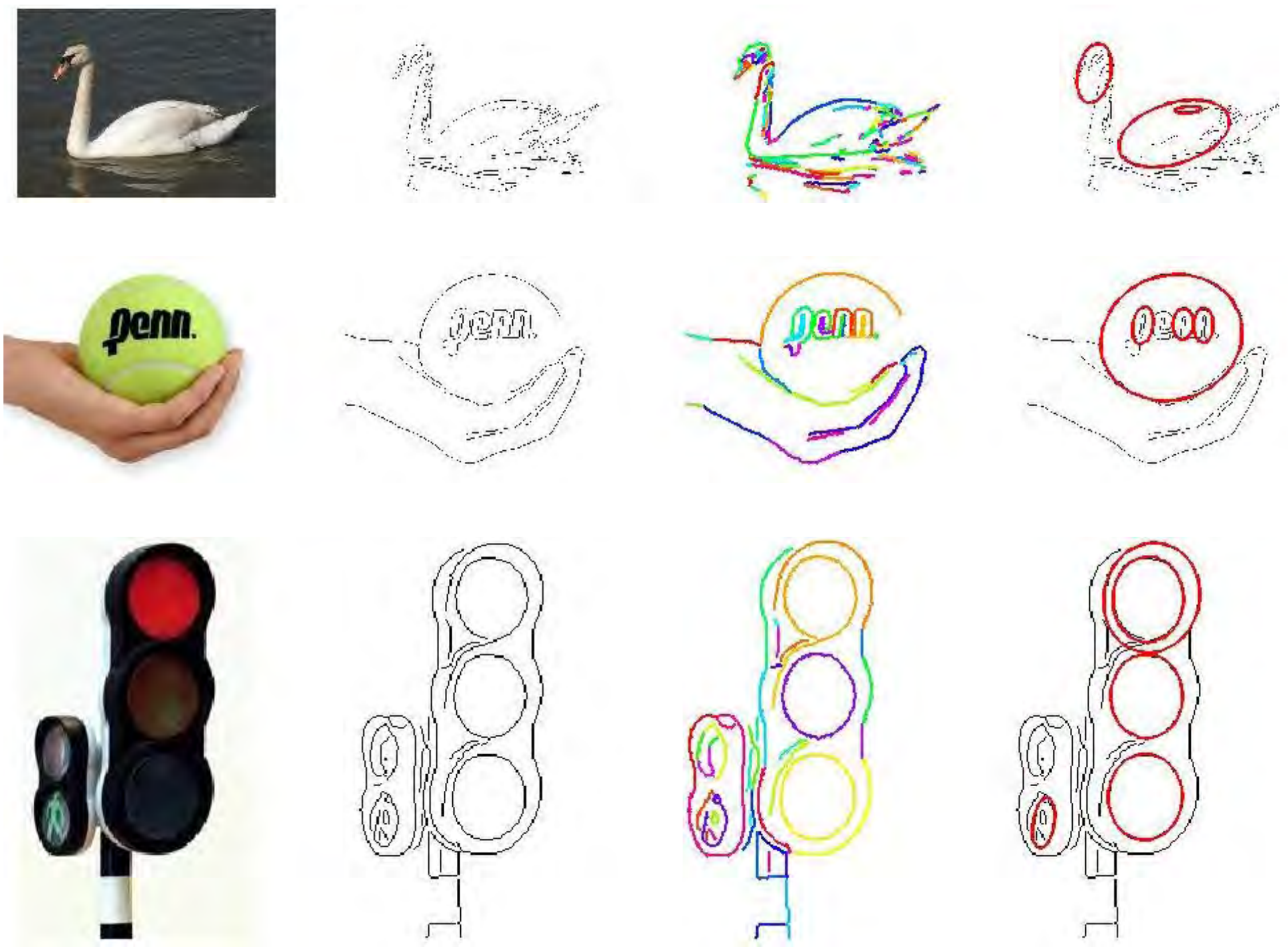

Fig. 57: Examples of real images and ellipse detection using the proposed method (continued).

## 3.8 Conclusion

It has been demonstrated using 37 object categories that elliptic shapes can serve as cues for object detection. The proposed method is able to find the significant elliptic shapes in most cases with good reliability. Thus, the preliminary work done gives sufficient background and foundation for the use of elliptic cues in object detection problems.

# 4 Future work and conclusion

## 4.1 Future work

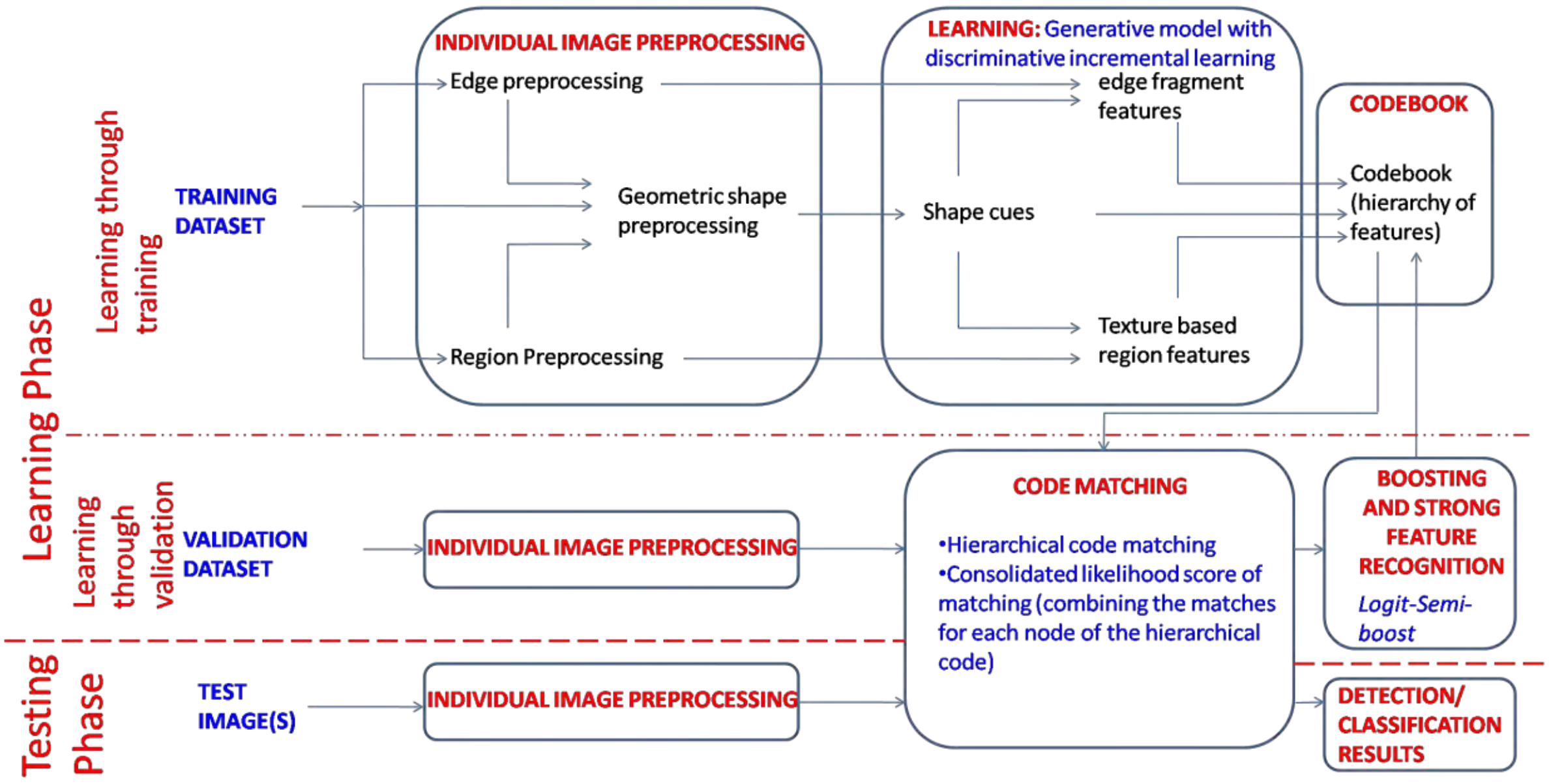

Fig. 58: Block diagram of the proposed object detection/recognition approach.

In this section, we present the details about the future work. The proposed object detection/recognition method and its novel features were introduced in section 1.3. Fig. 58 above is the same as Fig. 7 in chapter 1, and has been presented again here in order to retain the context. The proposed method follows the basic structure of Fig. 1, but has some very important differences from the contemporary methods. These are listed below:

### *4.1.1 Feature types*

The contemporary methods typically use a single feature type. The methods typically use multiple feature types usually use edge fragments and patches as features. However, as discussed in the second point in section 1.2.2, in order to ensure a good representation of the image as well as the object class, we intend to use edge fragments (for example, Fig. 2(c)) and region features (for examples, Fig. 3(e)). Further, region features are more correlated to the edge fragments, and good methods like triangulation and contour extension are available for this purpose. Due to this, such feature combination shall provide good object detection, recognition, and segmentation capabilities.

The potential of edge fragments as features has been demonstrated in [2, 7]. The edge fragments obtained using the edge processing method, discussed in chapter 2, can be used for finding the edge features. We intend to use textons as the region features, since texture based features provide good performance for segmentation, recognition, as well as detection. See section 1.2 for more details. Basic framework for integrating the textons and edge fragments has been developed and tested by [6, 75, 254].

However, the main contribution of the proposed method in terms of features is the use of geometrical shapes for finding more reliable features [53]. It has been shown in chapters 2 and 3, that linear and elliptic shapes can be retrieved with good reliability (better than most existing techniques) in various kinds of real images. However, we do not use the geometric/structural model of objects as the primary cues, as in [16, 52]. We use them in addition to the edge and region features as potential features, which may add to the robustness of object detection and recognition in various difficult scenarios. In addition, geometrical shapes (like lines, quadrangles, and ellipses) can be used to find reliable edge and region features in a more deterministic manner than the currently used methods. For example, the edges that should be used as potential features are selected randomly and in large quantities, with the hope that by selecting large number of potential features, one may not miss the reliable features [2, 7]. However, by using the geometrical shape contexts, reliable features can be extracted from the initial stage itself. The geometric shapes can also be used to identify (and cluster if necessary) the regions of maximum importance to the object class.

Consider the examples in Fig. 59. The thick red lines in the examples can be used as features. They can also be used to identify the edge features in their proximity as strong and reliable features. Further, a more deterministic method can be used to find the edge fragments. See the images in the third column of Fig. 59. Each of these images is the edge pixel density diagram of the original image. To compute edge pixel density, the image is divided into bins, each of size $10 \times 10$ pixels, and the number of edge pixels in each bin are computed and normalized by the total number of pixels. The bins around the detected shape features with high pixel densities can be used to identify the reliable edge fragments. However, this is a preliminary study and crude at present. In future, refinement over this idea could be possible.

We also suggest to incorporate agility of classes like animals by using a hierarchical scheme, in which the agile features are also trained and tested for agility (like rotation, displacement, translation, etc). For example, the thick blue line in Fig. 59(a,b) represent the tail in each case. Incorporating such orientation change as in these examples is not possible in the existing methods. However, in our representation, agility can be considered by using tail as just a node in the hierarchical code book and specifying a larger rotation range for it.

Original Image | Linear and elliptic shapes | Edge pixel density diagram

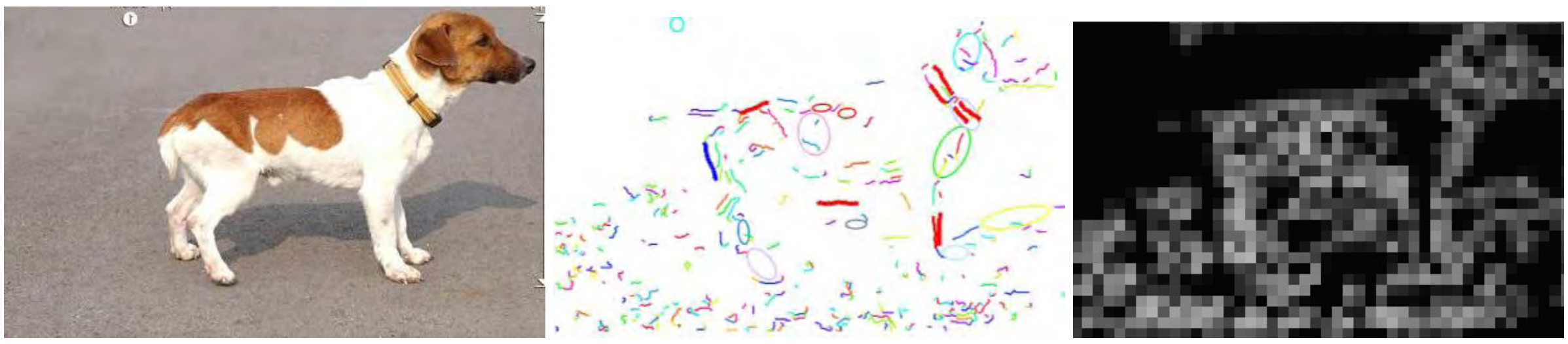

(a) Dog: example 1

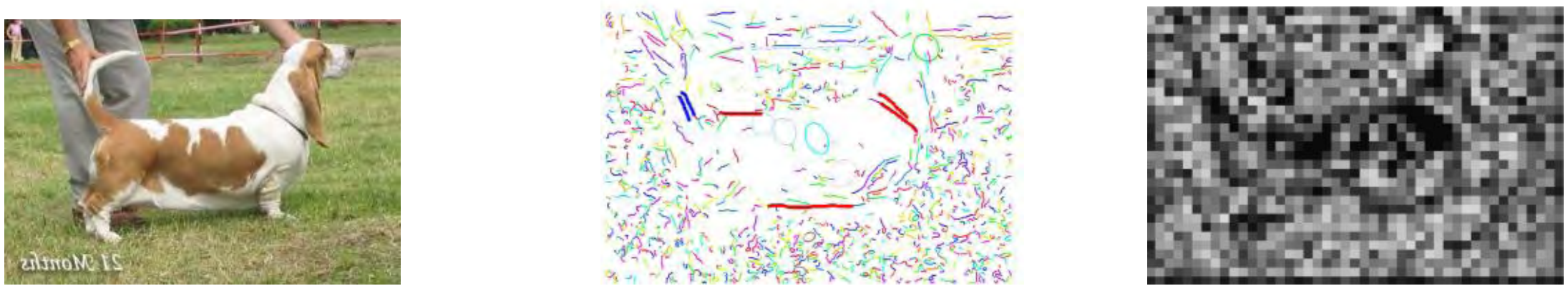

(b) Dog: example 2

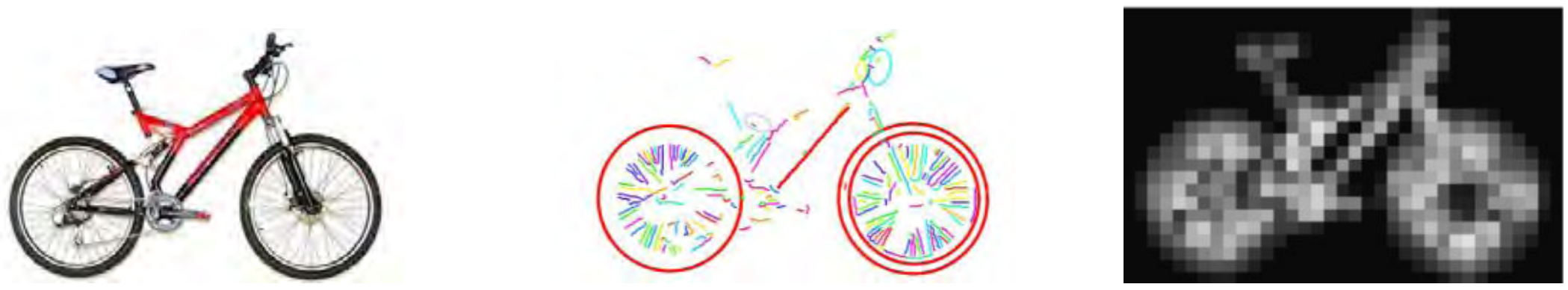

(c) Bicycle: Example 1

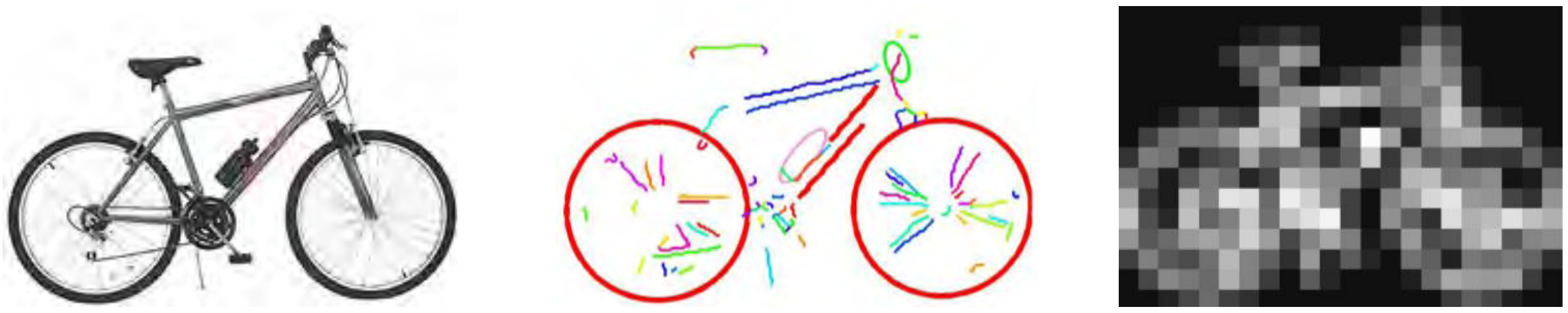

(d) Bicycle: Example 2

Fig. 59: Examples of the use of geometric shape cues. The thick red lines/arcs show the cues that may be helpful as features as well as identifying strong edge features. Note the thick blue lines in (a) and (b). The agility of tail can be learnt by supervision as a feature (a node in the hierarchical code) with a larger orientation margin.

The identification of the prominent shape cues and integrating the shape cues with the edge and region features shall be the first steps in the future work.

### *4.1.2 Codebook and matching schemes*

We shall create a hierarchical code for each object class. The hierarchical code shall contain most generic features at the top. The generic features have more likelihood ratios than the other features. The next level of features will be less generic and more class-specific than the ones at the top. Each node will contain the feature, the type of feature, the type of matching technique, the type and amount of agility to be considered, and the likelihood ratio. As the number of classes grows, look up tables for features may be made for efficient storage of the features and the nodes may contain only index of the feature instead of the actual feature, along with other data. This hierarchical template has been introduced in [255] in the context of realizing consciousness in machine [256-258].

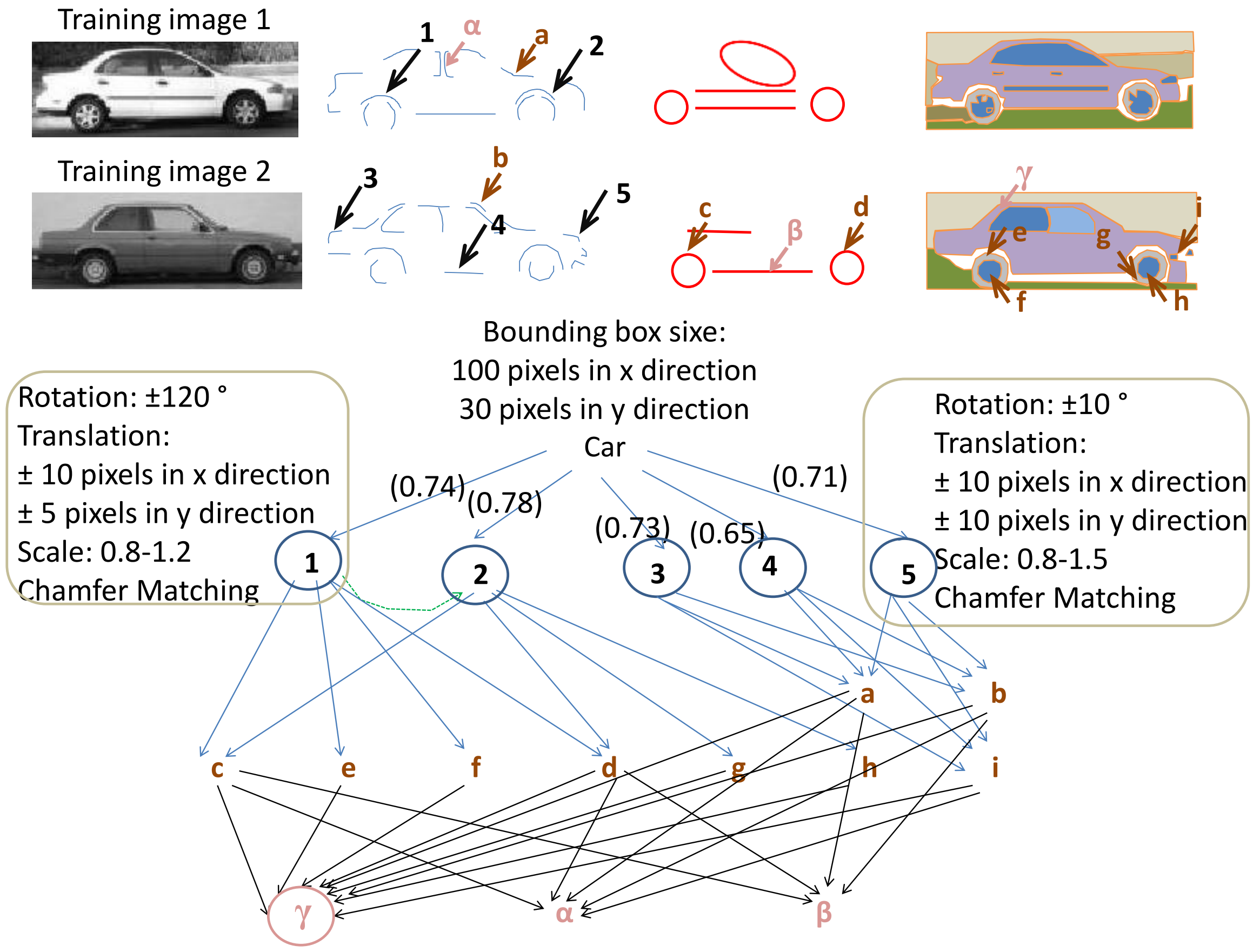

Fig. 60: An example of the proposed hierarchical code. The numbers 1-5 marked using large black arrows show the highest level in the hierarchical (most generic features). The alphabets a-i marked using medium brown arrows show next level of hierarchical code (which are more discriminative than upper level features). The Greek alphabets α-γ marked using small arrows show the lowest level in the hierarchical code. Each connection is given a weight that is equal to the likelihood of presence of a feature given the upper node in the connection is present. All the features are their levels are just examples selected heuristically for the demonstration of the concept, actual training may result in another set of features and code.

An example of such a code is shown in Fig. 60. Two training images of cars are considered on which the edge features, geometric features and texture based region features are shown. All of them have been generated heuristically only for the sake of illustration of the concept. The features that represent the object class car have been marked heuristically using three kinds of arrows and three different notations. The numbers 1-5 marked using large black arrows show the highest level in the hierarchical (most generic features). The alphabets a-i marked using medium brown arrows show next level of hierarchical code (which are more discriminative than upper level features). The Greek alphabets α-γ marked using small arrows show the lowest level in the hierarchical code. The text boxes near features 1 and 5 show examples of the details of using the features 1 and 5. Such info about the expected variability and type of matching should be stored with each node.

Each connection is given a weight that is equal to the likelihood of presence of a feature given the upper node in the connection is present. It is notable that the model looks similar to a perceptron network and the descriptive framework typical of Markov fields. If we compare it to perceptron network, we need to remember that the presented framework is essentially feed forward and the network is not fully connected. For example, the node 2 is not connected to node ‘i’. In a perceptron network, it is inferred as the weight of connection is zero. However, in the present case, it simply means that the likelihood of feature ‘i’ is not increased if feature 2 is known to be present. This is not to say that the concepts of perceptron network should not be applied here. This is just to say that though the concepts of perceptron network may be useful, they should be carefully applied here. Similarly, random Markov fields typically (though not always) assume a fully connected graph. Here, the hierarchical code is acyclic and not fully connected. In a more refined and accurate model, we may consider connection between the nodes of same level (shown in green dashed arrows), but this is not the scope of present work.

Another important point to note is that the sum of likelihoods of the nodes belonging to the same parent is not one. This is expected because the feature types chosen are not linearly independent. They are rather correlated, and there are intersections between various features’ likelihood ratios. This correlation of features has been retained deliberately in order to increase the recall ratio. Even if one of the paths among all possible paths is chosen, the object stands a chance to recognize as the object model is sufficiently represented by any path to enable recognition. Further, this model is expected to provide good discriminative performance as well, because the lower level features may have their likelihoods very low, but conditional to the presence of generic features, their likelihood increases, which implies that the simultaneous presence of generic and discriminative features is required to infer the class.

### 4.1.3 *Generative learning with semi-boost (discriminative incremental learning)*

Instead of using the discriminative model, we propose to use the generative model (with discriminative/incremental learning). As discussed in section 1.2.2, generative models aim at learning the object model, which should prove more versatile, scalable to unknown categories, generalizable, and thus robust than the discriminative model. However, as elucidated in the same section, generative models suffer from convergence problems, accuracy, and requirement of large datasets. Thus, it is advisable to use discriminative techniques within the framework of generative models, such that desirable aspects of both can be combined.

Some amount of work has already been done in this regard. Specifically, [92] used a modified version of Bayesian classifier, such that the likelihood ratios, learning path and convergence are controlled by some adaptive parameters, which are in turn optimized using discriminative techniques. However, more inspirational is the work in [38, 47], which combines boosting with the Bayesian classifier. It is inspiring for two main reasons. First, a well optimized boosting scheme can be used in this stage as well as in the validation stage. Further, with the proposal of semi-boost [177], it is possible to use semi-supervised dataset, and to adaptively scale the dataset for number of images as well as classes as and when needed. Thus, the advantages of generative algorithms can be availed by using small labeled dataset in the beginning and scale it up in the later stages. We also propose to develop semi-boost in the framework of Logit-boost, which is gentler than Ada-boost, and more robust to noise and clutter. In order to learn the geometric features, we may include ideas of Olson [53, 259].

Though the above scheme is expected to be semi-supervised, we will need to provide supervised input (for example a rotation range for the tails) for incorporating agility as mentioned in section 4.1.1. However, given the strengths of the above scheme, this may just be an initial requirement, just for the first few training images.

It is also important to discuss the learning of the hierarchical code too. After first execution of semi-boost, the features with high likelihood (the determination of thresholds is a part of future work) are identified as the highest level features. One high level feature is considered at a time and the likelihood of the remaining features conditional to the selected high level feature is computed. The features with the conditional likelihood greater than a certain threshold (determinable in future) are then assigned as the children of the selected high level feature. After completing the second level, the process is repeated by using the likelihood conditional to the presence of features in the second level (irrespective of the first level features). A chain terminates if no more features have the conditional likelihood above the required threshold.

Since the proposed scheme is expected to be scalable for larger dataset and newer classes, when a new training/validation dataset arrives, the existing hierarchical codes need to be checked and improved if necessary. In such situation, we evaluate the likelihoods in the hierarchical code and either update the likelihood or terminate a portion of chain or append it only if the change in likelihood is significant. This means that we trust the existing code which has been in use as long as we find sufficient possibility of improvement. At present this idea is in nascent state and possibilities of further improvements exist.

### *4.1.4 Matching schemes*

The matching will begin at the top and progress down the hierarchy. Thus, effectively, we begin with a high optimism to find an object, and become more and more decisive as we progress down the code. If we cannot go further down the tree, we need not look for multiple other traversal paths beginning again from the top. This provides multiple paths in the object detection/recognition chain and improves the robustness of overall algorithm.

Another important advantage of such hierarchical codes with generic features at the top is that it enables the clustering of object classes based on the generic features [94]. The idea is that super-classes of the object classes may be identified based on the generic features. Accordingly, a newly learnt object class can also be identified as potentially belonging to a super-class, and already learnt features may be reused. For example, the learnt codes may have the potential of grouping horses, cows, dogs, etc as animals; cups, vases, kettles, etc as pots; bicycles, cars, carts, etc as vehicles; and so on.

Substantial amount of literature is available on making the features and feature matching affine invariant, scale invariant, rotation invariant, and so on. Thus, we may use good matching techniques like chamfer distance (for edge features) and correlation/entropy based methods (for region features) for matching. We also suggest that the edge features can be stored in a framework similar to SIFT, where the orientation is used as a powerful tool for improving the feature quality. Such technique shall make the features less orientation/rotation sensitive, and simplify the requirement on the chamfer distance in terms of rotation and pose variance.

While testing, the amount of matching at a node and the likelihood ratio assigned to a node shall be combined together to form a trust score (match value) at each node and the trust scores for all the nodes can be combined according to the topology of the code in order to generate a trust score which can be then used to make the decision.

Every node will have a match value calculated using the matching scheme allotted to it, say $\tilde{m}_{node}$, denoted by a tilde above it. If a node has children, the node will have an additional match value due to the presence of children. The additional match value of the parent due to children, denoted by a hat above it, is computed as follows:

$$\hat{m}_{parent} = \sum_{\forall children} l_{parent,child} m_{child}, \tag{25}$$

where, $l_{parent,child}$ is the likelihood of the child feature conditional to the parent feature and $m_{child}$ is the match value of the child. The net match value for a parent is:

$$m_{parent} = \frac{\tilde{m}_{parent} + \hat{m}_{parent}}{2}. \tag{26}$$

In this way, the match value for the complete code can be computed in a bottom-up manner (beginning at the lowest level feature nodes), though the matching is actually done in a top-down manner. The match value of the complete code is compared against a threshold (to be determined later) in order to make an inference.

## 4.2 Conclusion

We have proposed a versatile method for object detection and recognition which uses a hierarchical code that contains various possible object models of each object such that the hierarchical code provides both generalization within class and good inter-class separation. Further, the method shall require a small completely supervised training set and can deal with large unsupervised datasets that may be added to the system at any stage. The scheme is also expected to be capable of learning new classes and optimizing the current class models online. The foundation of the proposed approach has been laid in the current report and the discussion of existing methods and future path adequately demonstrate the feasibility of the proposed approach. Thus, the presented approach is promising for highly adaptive multipurpose computer vision applications [255, 260, 261].

# Appendix

## A. Selection of points for the geometric method for center finding

As discussed in section 3.5.3, one set consisting of three points is required to generate one center point. Numerous such sets are required to generate reliable results. There are two ways in which we can choose the sets of three points. First, we may choose the sets of three points randomly. Second, we can split the edge into three sub-edges and choose one pixel from each sub-edge to form a set. Again, from each sub-edge, we may choose the pixels randomly or sequentially. Here, we present the effect of adopting each method.

First of all, it should be noted that some sets of points $P_1$, $P_2$, and $P_3$ in Fig. 32, may not give feasible solutions. One case is that the retrieved centre for a set falls outside the region of interest. For example, we might be interested in ellipses that are within the region covered by the image. Or we may be interested in ellipses that are located in a region that includes the image and a certain portion around the image. Another case is that $t_1$ and $t_2$, or $t_2$ and $t_3$ are parallel to each other. Third case can be that $l_{12}$ and $l_{23}$ are parallel to each other (i.e., the pixels in the set are collinear, though they may belong to a curved edge). All such cases will generate an invalid center. Second, since we perform center voting in the next step, we need sufficiently high number of sets so that the voting is reliable and robust. Thus, it is important to select large number of sets for finding the centers so that the detection of the centers is reliable and robust.

Here, we refer to [232] for studying the effect of selection of points. Their work suggests that every pair of point in a set (of three points) should satisfy these conditions such that chances of occurrence of the above mentioned problems are reduced: (1) Proper convexity, (2) proper distance between the points, and (3) reasonable angle between the tangents

From this perspective, it is most reasonable to split an edge into three sub-edges and choose points sequentially from each sub-edge. However, if the edge is not long enough, using this method will not ensure sufficiently high number of sets. Thus, we decide to split an edge into three sub-edges and choose points randomly from each sub-edge to form the desired number of sets.

## B. Computation of tangents: effect of quantization

Computation of tangents is required for the geometric method for center finding as well as for checking the continuity between two edges (section 3.5.1, 3.5.3 and 3.6.2.3). The tangent to any two-dimensional curve is mathematically given as $\partial y/\partial x$. In the discretized space, the tangent is usually calculated using the method of differences as $\Delta y/\Delta x$. Typically, $\Delta x$ should be very small, such that $\Delta y/\Delta x$ is a good approximation of $\partial y/\partial x$. However, for the edges in an image, this will definitely lead to a poor result. This can be easily explained from the fact that if we take smallest possible values $\Delta x$, i.e. 0 or $\pm 1$, $\Delta y/\Delta x$ is either 0, or ± 1, or $\infty$. This is because the next edge pixel shall lie in $3\times 3$ neighborhood of the pixel. Thus, the calculation of tangent cannot be done as above. This problem is depicted graphically in Fig. 61 (a). This effect in our opinion is directly related to the reliability/precision uncertainty discussed in [203, 204].

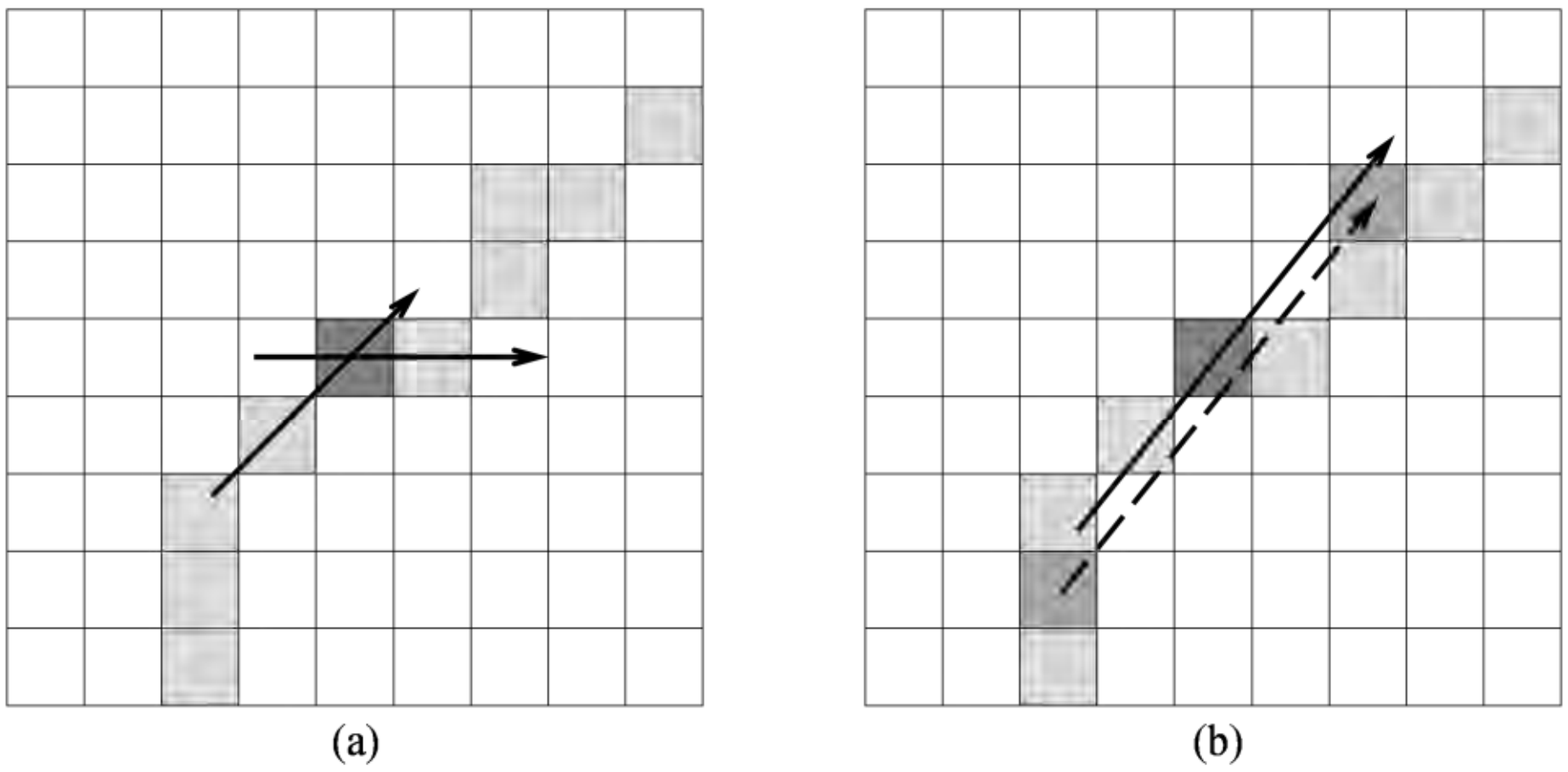

Fig. 61: Finding the tangents. The darkest pixel is $P_i$, where the tangent has to be calculated. The non-white pixels represent an edge. (a) The tangents if calculated with $\Delta x$, i.e. 0 or $\pm 1$ is not able to represent the tangent well. (b) Proposed approach for calculating the slope of the required tangent: we find the slope of the line passing through the points $P_{i-p}$ and $P_{i+p}$, $p=3$, which is approximately close to the slope of the required tangent. Using the calculated slope, the line passing through $P_i$ is calculated and used as tangent.

Thus, for calculating the tangent, we have used the following. Let the points in an edge list $e$ be denoted as $e=\{P_1(x_1,y_1) \quad P_2(x_2,y_2) \quad \dots \quad P_N(x_N,y_N)\}$. Suppose we want to find the tangent at pixel $P_i$. We first find the slope of the line passing through the points $P_{i-p}$ and $P_{i+p}$, where $p$ is a small integer. We have used $p=3$ in our implementation. The slope of this line is approximately parallel to the tangent we seek. Then, we can find a line with this slope, however, passing through the point $P_i$. Geometrically, the equation of the tangent is being calculated as below:

$$y = mx + c, \tag{27}$$

where, $m = \dfrac{y_{i+p} - y_{i-p}}{x_{i+p} - x_{i-p}}$ and $c = y_i - mx_i$. This is also illustrated graphically in Fig. 61 (b).

## C. Finding bin numbers of the centers

The digitization in the images causes the computation of centers (section 3.5.3) to be inaccurate. Simplistically speaking, if we try to focus on a small region, though the calculations will be locally precise, they are unreliable on a larger scale. This brings us to the reliability/precision uncertainty principle [204]. In effect, the centers computed above cannot be used directly as the centers calculated from various sets may be close but not exactly the same. To obtain a reliable pattern, we need to quantize the parametric space of centers. This is done by forming bins in the space where centers may lie. By doing so, the computed centers can be clustered and a representative center for each significant cluster can be used.

Let the input image be of size $M \times N$, where $M$ and $N$ are the number of pixels along the rows and columns respectively. We divide this region into $B_m$ equal bins along the rows and $B_n$ equal bins along the columns. Thus, each bin is of size $m \times n$ pixels, where $m = M/B_m$, $n = N/B_n$, and the total number of bins is $B = B_m B_n$. Each bin is assigned a unique bin number in the order of their occurrence while performing a raster scan of the region of interest. Fig. 62 gives an illustration of the concept. Thus any pixel $(x, y)$ shall belong to a bin $b$ where,

$$b = \left( \mathrm{ceil}\left( \frac{y-1}{n} \right) - 1 \right) \frac{M}{m} + \mathrm{ceil}\left( \frac{x-1}{m} \right). \tag{28}$$

where the function $\mathrm{ceil}(x)$ rounds the value of $x$ to the next larger integer. The above formula considers the image as the region of interest. However, a larger region around the image may be considered by suitably scaling $M$ and $N$, and modifying (28). It shall enable the detection of the ellipses whose centers do not lie within the image. It should be noted that considering larger region of interest shall increase the computation time.

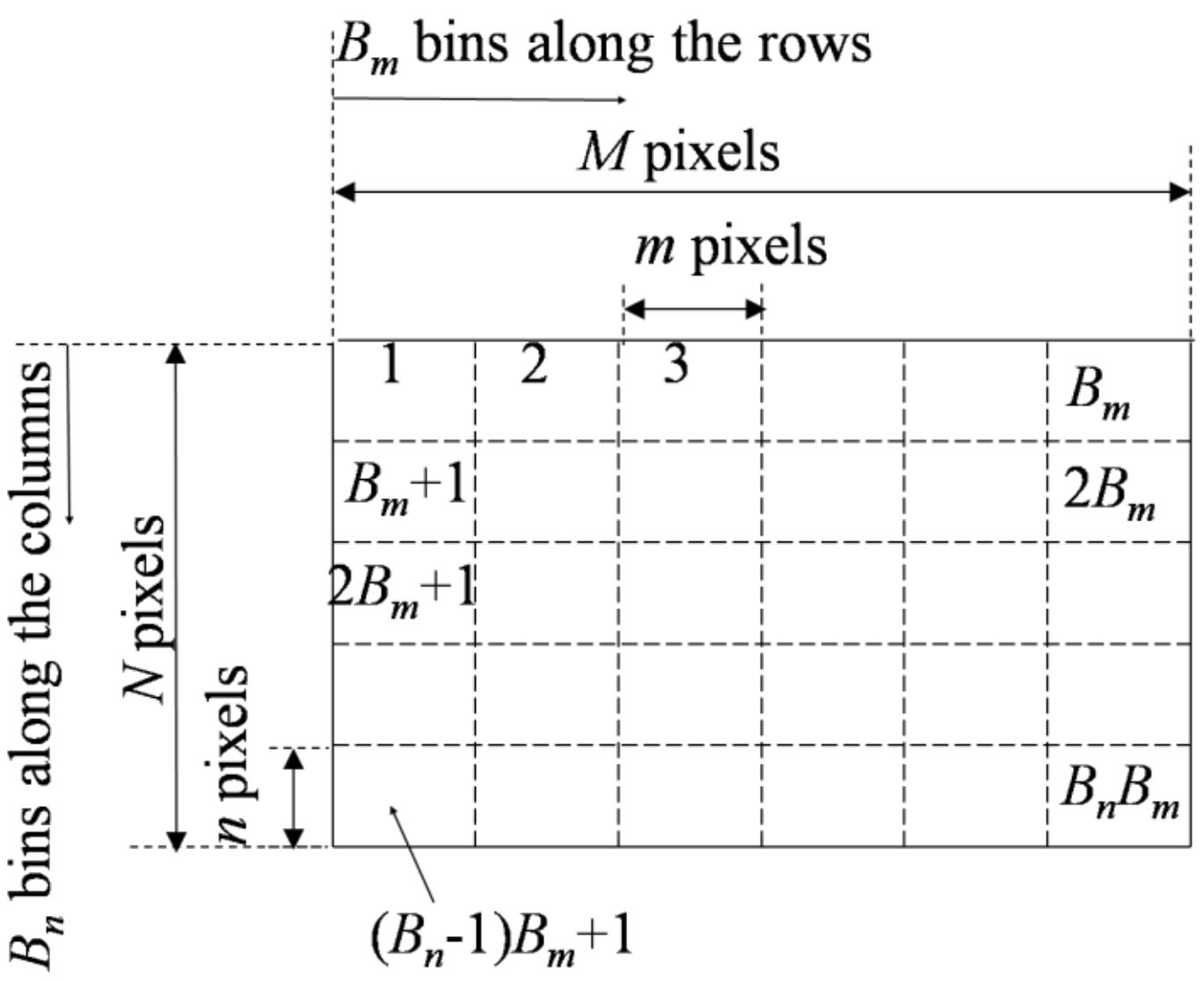

Fig. 62: Illustration of the binning concept for determining the bins of the centers found using this method.

## D. Normalization of the parameters

The method proposed for determining the similarity between two elliptic hypotheses used the following measures (rewritten from equation (13)):

$$D_x = \frac{|x_1 - x_2|}{X}, \tag{29}$$

$$D_y = \frac{|y_1 - y_2|}{Y}, \tag{30}$$

$$D_a = \frac{|a_1 - a_2|}{\max(a_1, a_2)}, \tag{31}$$

$$D_b = \frac{|b_1 - b_2|}{\min(b_1, b_2)}, \tag{32}$$

$$D_\theta = \frac{\angle(\theta_1, \theta_2)}{\pi}. \tag{33}$$

The details of other notations can be found in section 3.6.1. First, we evaluate the bounds on the five parameters and discuss their effect on the similarity measure:

1) Typically, the centers are expected to be in the image region, or at most in an additional small region around the image. For simplicity and without the loss of generality, at present we assume that the centers are in the image region only: $x_0 \in [0, X]$ and $y_0 \in [0, Y]$, where $X$ and $Y$ are the total number of pixels in the $x$ and $y$ directions respectively.

2) The sizes of the ellipses detected cannot be larger than the length of the diagonal across the image region. This is because, if the size of an ellipse would be larger than this, either a very small fragment can appear in the image region, or the center of the ellipse would far from the image region. In either case, such an ellipse will not only be hard to detect, such detection will also be not reliable enough. On the other hand, the smallest size of the ellipse is more than zero (or else it would be a point). Thus the bounds on $a$ and $b$ are: $b \in \left(0, \sqrt{X^2 + Y^2}\right]$ and $a \in \left[b, \sqrt{X^2 + Y^2}\right]$. It should be noted that the lower bound on $a$ is $b$, so that a is indeed the semi-major axis.

3) The orientation angle may take values in the range $[0, \pi)$.

In order to identify a suitable normalization parameter for each of the five variables, we consider one variable at a time and assume that the remaining variables have the same value for two elliptic hypotheses that are being compared. In the following, we discuss the normalization of the parameters using the above framework.

## *Normalization for* $x_0$ *and* $y_0$

Here, we consider that two ellipses have all the parameters same except the $x$ coordinates of their centers. This can be extended to the $y$ coordinate in an analogous manner. Let the $x$ coordinates of the two ellipses be $x_1$ and $x_2$ respectively. Let us define $D = |x_1 - x_2|$. The displacement $D$ between the two ellipses is small or large in comparison to the dimension of the image in the $x$ direction. Thus $X$ should be used to normalize $D = |x_1 - x_2|$.

## *Normalization for* $a$*,* $b$*, and* $\theta$

Say, all the parameters of two ellipses except $b$ are same. Then, the difference in $b$ is manifest as the difference in the eccentricity. Let the semi-minor axes of the two ellipses be $b_1 = b$ and $b_2 = b + \Delta b$, where $\Delta b$ is positive. Then, the difference in their eccentricities, $e_1$ and $e_2$, is given as follows:

$$e_1^2 - e_2^2 = \frac{b_2^2 - b_1^2}{a^2} = \frac{\Delta b(2b + \Delta b)}{a^2} = \left(\frac{b}{a}\right)^2 \left(\frac{\Delta b}{b}\left(2 + \frac{\Delta b}{b}\right)\right). \tag{34}$$

Since $a$ is considered a constant and $b$ is the reference value for calculating the deviation, $(b/a)$ can be considered a constant and the difference in eccentricities is proportional to the second term in the right hand side of (34). To further understand the impact of $\Delta b/b$ on the difference in eccentricities, we plot the graph of expression $(a/b)^2\left(e_2^2-e_1^2\right)$ vs. $\Delta b/b$ in Fig. 63. It is noticeable that the expression increases rapidly with $\Delta b/b$, the function is non linear and even when $\Delta b/b$ =0.5, the value of expression has reached above 1. This trait is desirable to evaluate similarity.

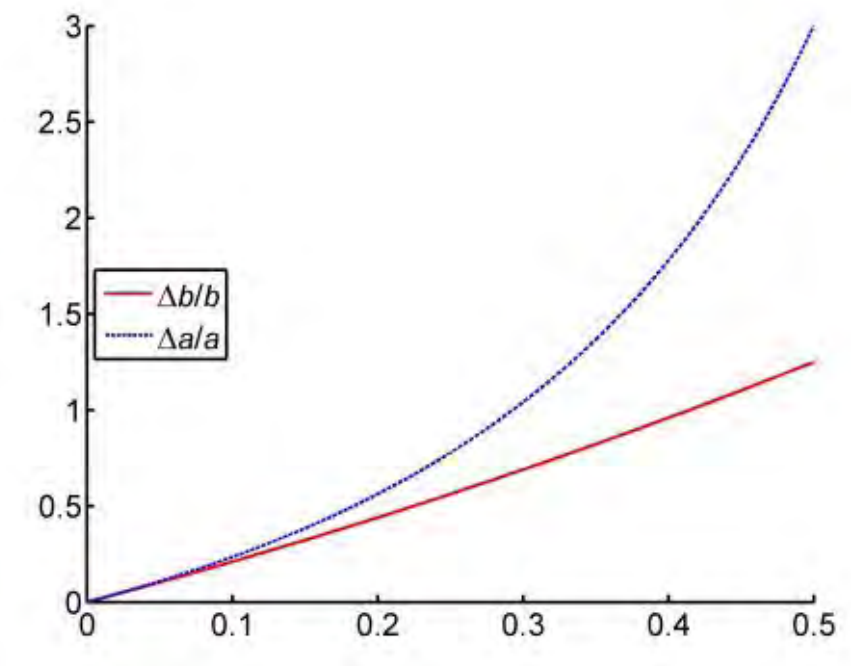

Fig. 63: The difference in the eccentricities of two ellipses.

Now, let us study the impact of difference in $a$ on the difference in eccentricities. As before, we assume that all parameters except $a$ are same for both the ellipses. Let the semi-major axes of the two ellipses be $a_1 = a$ and $a_2 = a - \Delta a$ (note that $a_1 > a_2$), where $\Delta a$ is positive. Then, the difference in their eccentricities, $e_1$ and $e_2$, is given as follows:

$$e_1^2 - e_2^2 = b^2\frac{a_1^2 - a_2^2}{a_2^2 a_1^2} = b^2\frac{\Delta a\left(2a-\Delta a\right)}{a^2\left(a-\Delta a\right)^2} = \left(\frac{b}{a}\right)^2\left(\frac{\Delta a}{a}\left(2-\frac{\Delta a}{a}\right)\right)\Big/\left(1-\frac{\Delta a}{a}\right)^2. \tag{35}$$

Since $b$ is considered a constant and $a$ is the reference value for calculating the deviation, $(b/a)$ can be considered a constant and the difference in eccentricities is proportional to the remaining term in the right hand side of (35).

To further understand the impact of $\Delta a/a$ on the difference in eccentricities, we plot the graph of $(a/b)^2\left(e_2^2-e_1^2\right)$ vs. $\Delta a/a$ in Fig. 63. Though the expression in (35) is more complicated than in (34), the general observations are same as before.

From the above discussion regarding the parameters $a$ and $b$, we conclude that while discussing the similarity between two ellipses with respect to these parameters, the best way of normalization is not with respect to the maximum bounds. It is rather useful to consider $|b_1 - b_2| / \min(b_1, b_2)$ while comparing two ellipses on the basis of their semi-minor axes, and $|a_1 - a_2| / \max(a_1, a_2)$ while comparing two ellipses on the basis of their semi-major axes.

If we choose the expressions $|b_1 - b_2| / \max(b_1, b_2)$ and $|a_1 - a_2| / \min(a_1, a_2)$ instead of the above, we do not see a rising trend as seen in Fig. 63. Instead, the curves become insensitive to the rise in $\Delta b/b$ and $\Delta a/a$. The details of this have been skipped for brevity.

The range of the orientation angle suggests that $\angle(\theta_1, \theta_2)$ (the smallest difference in the angles of orientation) be normalized by $\pi$.